\pdfoutput=1

\documentclass[dvipsnames, 11pt]{article}

\usepackage[final]{acl}

\usepackage{times}
\usepackage{latexsym}
\usepackage[T1]{fontenc}
\usepackage[utf8]{inputenc}
\usepackage{microtype}
\usepackage{inconsolata}
\usepackage{colortbl}
\usepackage{adjustbox}
\usepackage{diagbox}
\usepackage{amsmath,amssymb, amsthm}
\usepackage[inline]{enumitem}
\usepackage{array,multirow,graphicx,makecell}
\usepackage{booktabs}
\usepackage{tabularx}
\usepackage{xfrac}
\usepackage[skins]{tcolorbox}
\tcbuselibrary{breakable}
\usepackage{anyfontsize}
\usepackage{framed}
\usepackage{tabto}
\usepackage{tabulary}
\usepackage{anyfontsize}
\usepackage[toc,page,header]{appendix}
\usepackage{minitoc}

\newcolumntype{C}[1]{>{\centering\let\newline\\\arraybackslash}m{#1}}
\newcounter{tcbcounter}

\newtcolorbox[auto counter, number within=section]{prompt}[3][]{%
  enhanced,
  breakable,
  colback=#2!5!white,
  colframe=#2!75!black,
  title=\textbf{Box \thetcbcounter: #3},
  fontupper=\normalsize\fontfamily{cmss}\selectfont,
  #1
}

\newtcolorbox{interpretabilitybox}{
enhanced,
boxrule=0pt,frame hidden,
borderline west={4pt}{0pt}{Emerald!75!black},
colback=Emerald!9!white,
sharp corners
}

\newtcolorbox{explainingbox}{
enhanced,
boxrule=0pt,frame hidden,
borderline west={4pt}{0pt}{RoyalBlue!55!black},
colback=RoyalBlue!9!white,
sharp corners
}

\newtcolorbox{mechanismbox}{
enhanced,
boxrule=0pt,frame hidden,
borderline west={4pt}{0pt}{OliveGreen!75!black},
colback=OliveGreen!9!white,
sharp corners
}

\newtcolorbox{termsbox}{
enhanced,
boxrule=0pt,frame hidden,
borderline west={4pt}{0pt}{Peach!65!black},
colback=Peach!9!white,
sharp corners
}

\newtcolorbox{communicatingbox}{
enhanced,
boxrule=0pt,frame hidden,
borderline west={4pt}{0pt}{DarkOrchid!75!black},
colback=DarkOrchid!9!white,
sharp corners
}

\newtcolorbox{extractingbox}{
enhanced,
boxrule=0pt,frame hidden,
borderline west={4pt}{0pt}{Red!75!black},
colback=Red!9!white,
sharp corners
}

\newtcbox{\inlinebox}[1]{enhanced,
  on line, 
  box align=base,
  nobeforeafter,
  colback=#1!25!white,
  colframe=#1!85!black,
  size=fbox,
  left=0pt,
  right=0pt,
  boxsep=2pt,
  boxrule=1pt,
  tcbox raise base}

\newcommand{\inlinelabel}[2]{\inlinebox{#1}{{\fontfamily{qag}{\selectfont\scriptsize{\textcolor{#1!70!black}{#2}}}}}}

\newcommand{\mechinputoutput}{\inlinelabel{OliveGreen}{input-output}}
\newcommand{\mechconceptoutput}{\inlinelabel{LimeGreen}{concept-output}}
\newcommand{\mechinputconceptoutput}{\inlinelabel{YellowGreen}{input-concept-output}}
\newcommand{\mechinputinternal}{\inlinelabel{PineGreen}{input-internal}}
\newcommand{\mechinternalinternal}{\inlinelabel{Green}{internal-internal}}

\newcommand{\localexp}{\inlinelabel{Goldenrod}{local}}
\newcommand{\globalexp}{\inlinelabel{Dandelion}{global}}

\newcommand{\posthoc}{\inlinelabel{CadetBlue}{post-hoc}}
\newcommand{\intrinsic}{\inlinelabel{Gray}{intrinsic}}

\newcommand{\modelagnostic}{\inlinelabel{Cerulean}{agnostic}}
\newcommand{\modelspecific}{\inlinelabel{NavyBlue}{specific}}

\newcommand{\prescores}{\inlinelabel{Thistle}{scores}}
\newcommand{\previs}{\inlinelabel{Orchid}{visualiztion}}
\newcommand{\preexamples}{\inlinelabel{Purple}{examples}}
\newcommand{\pretext}{\inlinelabel{RoyalPurple}{text}}

\newcommand{\causalbased}{\inlinelabel{Red}{causal}}
\newcommand{\notcausal}{\inlinelabel{Maroon}{not}}

\newcommand{\cmechinputoutput}[1]{\textit{\textbf{\textcolor{OliveGreen!70!black}{#1}}}}
\newcommand{\cmechconceptoutput}[1]{\textit{\textbf{\textcolor{LimeGreen!70!black}{#1}}}}
\newcommand{\cmechinputconceptoutput}[1]{\textit{\textbf{\textcolor{YellowGreen!70!black}{#1}}}}
\newcommand{\cmechinputinternal}[1]{\textit{\textbf{\textcolor{PineGreen!70!black}{#1}}}}
\newcommand{\cmechinternalinternal}[1]{\textit{\textbf{\textcolor{Green!70!black}{#1}}}}

\newcommand{\clocalexp}[1]{\textit{\textbf{\textcolor{Goldenrod!70!black}{#1}}}}
\newcommand{\cglobalexp}[1]{\textit{\textbf{\textcolor{Dandelion!70!black}{#1}}}}

\newcommand{\cposthoc}[1]{\textit{\textbf{\textcolor{CadetBlue!70!black}{#1}}}}
\newcommand{\cintrinsic}[1]{\textit{\textbf{\textcolor{Gray!70!black}{#1}}}}

\newcommand{\cmodelagnostic}[1]{\textit{\textbf{\textcolor{Cerulean!70!black}{#1}}}}
\newcommand{\cmodelspecific}[1]{\textit{\textbf{\textcolor{NavyBlue!70!black}{#1}}}}

\newcommand{\cprescores}[1]{\textit{\textbf{\textcolor{Thistle!70!black}{#1}}}}
\newcommand{\cprevis}[1]{\textit{\textbf{\textcolor{Orchid!70!black}{#1}}}}
\newcommand{\cpreexamples}[1]{\textit{\textbf{\textcolor{Purple!70!black}{#1}}}}
\newcommand{\cpretext}[1]{\textit{\textbf{\textcolor{RoyalPurple!70!black}{#1}}}}

\newcommand{\ccausalbased}[1]{\textit{\textbf{\textcolor{Red!70!black}{#1}}}}
\newcommand{\cnotcausal}[1]{\textit{\textbf{\textcolor{Maroon!70!black}{#1}}}}

\title{On Behalf of the Stakeholders: \\ Trends in NLP Model Interpretability in the Era of LLMs}

\author{Nitay Calderon \and Roi Reichart \\
        Faculty of Data and Decision Sciences, Technion\\ \texttt{nitay@campus.technion.ac.il} and \texttt{roiri@technion.ac.il}}

\begin{document}
\maketitle

\doparttoc 
\faketableofcontents 

\begin{abstract}
Recent advancements in NLP systems, particularly with the introduction of LLMs, have led to widespread adoption of these systems by a broad spectrum of users across various domains, impacting decision-making, the job market, society, and scientific research. 
This surge in usage has led to an explosion in NLP model interpretability and analysis research, accompanied by numerous technical surveys.
Yet, these surveys often overlook the needs and perspectives of explanation stakeholders.
In this paper, we address three fundamental questions: Why do we need interpretability, what are we interpreting, and how?
By exploring these questions, we examine existing interpretability paradigms, their properties, and their relevance to different stakeholders.
We further explore the practical implications of these paradigms by analyzing trends from the past decade across multiple research fields.
To this end, we retrieved thousands of papers and employed an LLM to characterize them.
Our analysis reveals significant disparities between NLP developers and non-developer users, as well as between research fields, underscoring the diverse needs of stakeholders.
For example, explanations of internal model components are rarely used outside the NLP field.
We hope this paper informs the future design, development, and application of methods that align with the objectives and requirements of various stakeholders.

\end{abstract}
\section{Introduction}
\label{sec:intro}

Recent advancements in Natural Language Processing (NLP), particularly with the introduction of Large Language Models (LLMs), have dramatically enhanced model performance. These models are now capable of executing a wide array of tasks and have been adopted across various domains and research fields \citep{Aletras2016PredictingJD, calderon2023measuring, YangJTHFJZYH24}. 
Their applications extend beyond the NLP community, and they are widely used by the general public \citep{Choudhury2023InvestigatingTI, Kasneci2023ChatGPTFG, vonGarrel2023ArtificialII}. However, these black-box models are complex and opaque \citep{WallaceTWSGS19, CalderonMRK23, LuoIHP24}.
While performance has advanced, this comes at the cost of understanding their underlying mechanisms \citep{Lyu2022, MadsenRC23, Singh2024}. 

The ability to explain decisions is particularly crucial, given that NLP models, especially LLMs, significantly influence individual decision-making \citep{Tu2024, Yu2024MedicalAI}, society \citep{Samuel2023ResponseTT, Taubenfeld2024}, the job market \citep{Eloundou2023}, and academic research \citep{2023ToolsSA, Liang2024}. Moreover, model interpretability and analysis are utilized for scientific insights and discoveries \citep{RoscherBDG20, Allen2023InterpretableML, Badian2023SocialMI, Birhane2023ScienceIT, Lissak2024}.

\begin{figure}[t]
    \centering
    \includegraphics[width=0.485\textwidth]{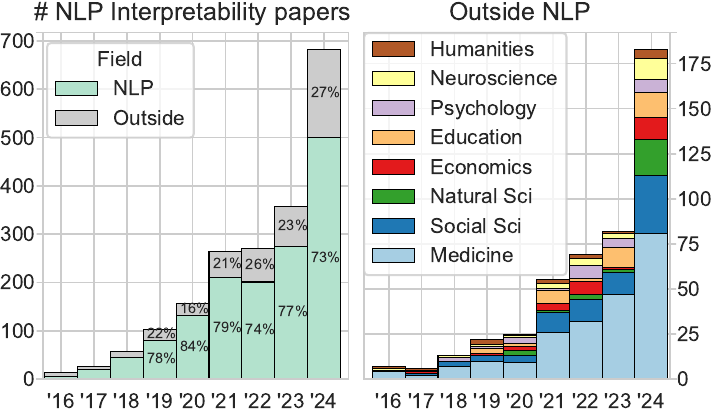}
    \caption{Number of \textit{NLP Interpretability} papers published over time. Each year spans from June of the previous year to the following June. The left plot shows the distribution of papers across NLP and the other fields (\textit{Outside}).
    The right plot shows trends in other fields besides NLP. Only papers that use, propose, or discuss interpretability methods applied to natural language are counted, following relevance filtering by an LLM.}
    \label{fig:intro_papers}
\vspace{-0.8em}
\end{figure}

Unsurprisingly, research on model interpretability and analysis has become one of the most prolific areas within the NLP community and beyond, yielding thousands of publications in recent years, as illustrated in Figure~\ref{fig:intro_papers}.
Consequently, many technical NLP model interpretability and analysis surveys have emerged, reviewing hundreds of methods \citep{BelinkovG19, Danilevsky2020, Balkir2022,sajjad-etal-2022-neuron, Bereska2024, LuoIHP24, ZhaoCYLDCWYD24, MosbachGBKG24}. In this paper, we aim to bridge a gap in the existing literature and discuss model interpretability from the stakeholders' perspective. Our goals are to broaden the NLP community’s point of view on the application of interpretability methods in various fields and to promote the design and development of methods that align with the objectives, expectations, and requirements of various stakeholders.

We will explore three key questions: why do we need interpretability (\S\ref{sec:why}), what are we interpreting (\S\ref{sec:definitions}), and how are we interpreting (\S\ref{sec:properties_discussion})? This approach allows us to examine common interpretability paradigms (Table~\ref{tab:categorization}), their properties and their applications by different stakeholders.

We start by presenting four perspectives on interpretability and their relevant stakeholders in \S\ref{sec:why}. Next, in \S\ref{sec:definitions}, we address a pressing issue in the literature: the lack of consensus on the definition of interpretability. We examine various definitions within and outside the NLP community and propose a broad definition: \textit{Extracting insights into a mechanism of the NLP system and communicating them to the stakeholders in understandable terms}.

In \S\ref{sec:properties_discussion}, we propose six properties of interpretability methods and discuss the relevance of each property to different stakeholders. For example, the \textit{scope} property distinguishes between local and global explanations. If the stakeholder is a physician, a local explanation that clarifies the prediction for an individual patient is preferred. Conversely, a global explanation is more suitable for a scientist, as it facilitates understanding broader phenomena.

We survey in \S\ref{sec:methods} seven prevalent interpretability paradigms, explain which properties characterize each (see Table~\ref{tab:categorization}), and discuss their applications by different stakeholders. Throughout the survey, we review over 200 works.

Following that, in \S\ref{sec:trends} we aim to understand how the paradigms and their properties are reflected in practice by analyzing trends over the years and across different research fields. To this end, we retrieved over 14,000 papers using the Semantic Scholar API and employed an LLM to select only relevant ones, resulting in 2,000 papers. Furthermore, we utilized the LLM to annotate papers with their interpretability paradigm and properties.\footnote{Data: \url{www.github.com/nitaytech/InterpreTrends}} Importantly, the LLM annotation is in strong agreement (over 90\%) with human expert annotation. To the best of our knowledge, this is the first successful application of an LLM for such a task.

Below, we summarise our main findings:
\begin{enumerate}[itemsep=1pt,parsep=1pt, topsep=1pt, align=left,leftmargin=*]
\item Within the NLP community, interpretability paradigm trends have remained stable over the decade. However, the introduction of LLMs in the past two years has prompted a notable shift.
\item Outside the NLP community, our main claim gains support: different stakeholders have varying needs, reflected in significant differences between research fields in terms of both the paradigms and their properties.
\item Comparing NLP developers to non-developers reveals that the latter group is less interested in understanding internal model components.
\item Non-developers opt for popular methods not originally developed within the NLP community, such as SHAP and LIME, likely due to their user-friendly and easy-to-apply software.
\item LLMs have shifted the trends in interpretability research: not only has the number of published papers doubled, but there has also been a substantial increase in the use of natural language explanations. These explanations are utilized in nearly half of the papers outside the NLP field.
\end{enumerate}
We hope this first-of-its-kind paper, which reviews NLP interpretability through the stakeholders' perspective and rigorously analyzes trends within and outside the NLP field, will pave the way for improved design, development, and application of these essential methods.
To further this aim, we outline in \S\ref{sec:discussion} practical steps that NLP researchers can undertake to promote the adoption of interpretability methods in other disciplines.
\begin{figure*}[t]
    \centering
    \includegraphics[width=0.85\textwidth]{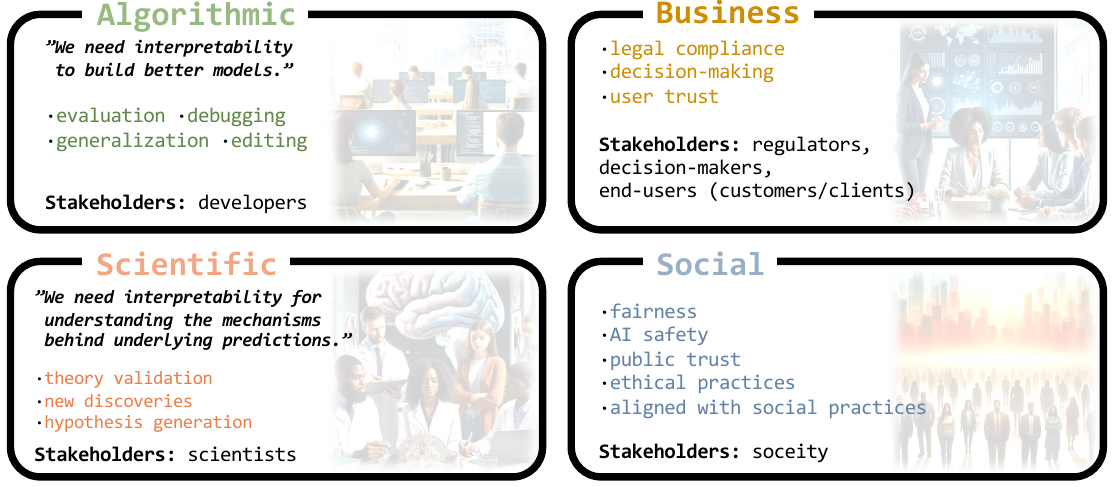}
    \caption{Overview of four perspectives on the need for interpretability proposed in this paper.}
    \label{fig:perspectives}
\vspace{-0.8em}
\end{figure*}

\section{Why Do We Need Interpretability?}
\label{sec:why}

Understanding why interpretability is necessary provides a solid framework for discussing, assessing, and enhancing interpretability methods, ensuring they meet practical objectives and expectations. 
So, when and why do we need interpretability? We gather ideas from other surveys \citep{GadeGKMT20, RaukerHCH23, Waddah2O23} and propose a decomposition of the need for interpretability into four perspectives (see Figure~\ref{fig:perspectives}): \textit{algorithmic, business, scientific, and social.} 

The four perspectives define the objective and use case of the interpretability method.
Clearly, there can be overlaps between the different perspectives, particularly with the algorithmic one. For example, using interpretability to build a better model (algorithmic perspective) might coincide with making it fairer (social perspective) or one that promotes more informed business decisions (business perspective). Similarly, promoting social values through interpretability (social perspective) can build customer trust (business perspective).

Besides the objectives of the interpretability method, another key consideration is the \textit{stakeholders}--the audience to whom the explanation is aimed and communicated. Accordingly, when designing the interpretability method, we should consider not only the objective (and the usage) of the explanation but also the stakeholders, their level of expertise, and their familiarity with NLP models. By identifying different stakeholders' specific requirements and concerns, we can foster practical interpretability methods that align with their expectations \citep{KaurNJCWV21}. We next discuss the four perspectives and the main stakeholders (\textbf{in bold}):

\medskip\noindent\textbf{\textcolor{LimeGreen!70!black}{1. The Algorithmic Perspective:}} emphasizes using interpretability for building better models. Thus, the stakeholders are  \textbf{developers}.
Interpretability allows for an open-ended, more rigorous evaluation beyond standard metrics \citep{Ribeiro0G18, Lertvittayakumjorn21, KabirL024}. 
It helps uncover why the model fails, offering insights into identifying and rectifying mistakes \citep{YaoCYJR21} and improving its generalization. 
For instance, \citet{GhaeiniFST19} use saliency maps, and \citet{JoshiCLNSF022} use counterfactual explanations for modifying the training objective and improving model robustness. \citet{Gekhman24} study the source of hallucinations in LLMs by curating a diagnostic set that utilizes the model's pre-existing knowledge.
Moreover, by understanding how the model works, we can intervene and modify it or design better models from the start (e.g., reverse engineering) \citep{MengBAB22, Arad2024}. For example, \citet{DaiDHSCW22} use attributions to locate knowledge neurons, modify them, and edit factual knowledge of the NLP model.
The algorithmic perspective underscores interpretability for debugging, refining model deployment, and forecasting progress.

\medskip\noindent\textbf{\textcolor{Dandelion!75!black}{2. The Business Perspective:}} focuses on leveraging interpretability across various sectors to enhance informed decision-making, legal compliance, and user trust. 
Models often aid decision-making at the \textbf{business} level (e.g., sentiment analysis for market research \citep{Hartmann2022MoreTA}) and at the \textbf{user} level (e.g., LLMs assisting \textbf{physicians} in \textbf{patient} diagnostics \citep{clusmann2023future}). In both cases, interpretability aids ensure well-grounded and trustworthy decisions \citep{LaiT19}.

Legal compliance includes cases where interpretability is explicitly the \textbf{regulator's} requirement, such as the \textit{GDPR's ``right to explanation''} \citep{right_to_explanation} and the \textit{Algorithmic Accountability Act} proposed in the US \citep{maccarthy2019examination}, or cases where interpretability is instrumental in ensuring the \textbf{business} adheres to legal standards, thereby reducing the risk of legal penalties. 
For instance, when NLP models are used to process credit applications \citep{ZhangWZW20, YangYL22, Sanz2024}, they must comply with the \textit{Equal Credit Opportunity Act (ECOA)}, which prohibits discrimination.

Finally, interpretability enhances transparency, fostering trust and goodwill. When \textbf{end-users} understand how decisions are made, they are more likely to trust AI systems, improving \textbf{business} reputation. For example, Facebook's 'Why am I seeing this ad' tool is designed specifically to provide more transparency and build trust \citep{Pavon_2024}. The link between interpretability and trust is well-documented 
\citep{Parasuraman1997, miller2016behavioral, BucincaLGG20}.

\medskip\noindent\textbf{\textcolor{RedOrange!65!white}{3. The Scientific Perspective:}} Language is tightly connected to human behavior, cognition, and communication. \textbf{Researchers} and \textbf{scientists} from various disciplines, such as social science \citep{lazer2020computational, ZiemsHSCZY24}, psychology \citep{psychology}, psychiatry \citep{psychiatry}, psycho-linguistics \citep{WilcoxLMF18}, health \citep{singhal2023large, thirunavukarasu2023large}, neuroscience \citep{goldstein2022shared, tikochinski2023perspective}, finance \citep{el2019search}, behavioral economics \citep{Shapira2023, Shapira2024}, political science \citep{gennaro2022emotion}, and beyond, are now turning to NLP to model scientific phenomena, decode complex patterns and derive meaningful insights about humanity. Science is all about gaining knowledge, and interpretability enables us to understand the underlying mechanisms and patterns the NLP model identifies, facilitating deeper comprehension and advancing scientific discoveries \citep{RoscherBDG20}. 
For example, by interpreting the representations of Facebook posts extracted by an NLP model, \citet{Lissak2024} identify a new risk factor for suicide ideation: boredom.


\medskip\noindent\textbf{\textcolor{CornflowerBlue!60!black}{4. The Social Perspective:}} addresses the broader impact of NLP systems on society, fairness, the ethical implications of its use and AI safety.
Since NLP models are optimized using data that may contain human biases and prejudices \citep{BlodgettBDW20, DevMOSPC21}, interpretability is crucial for understanding the rationale behind the models, ensuring they serve what they are designed for rather than reflecting their training data \citep{RuderVS22}.
Interpretability can confirm the model predictions are just and equitable \citep{OrgadGB22, AttanasioANL23, SantoshBSGN24}, foster public trust, promote ethical practices, and prevent misuse or other harmful consequences \citep{Bereska2024, Lissak2024_b}. Accordingly, interpretability helps \textbf{society} embrace the model or reject it, depending on how well it aligns with expected social values.

\section{Definitions} 
\label{sec:definitions}

\begin{table*}[!t]
\centering
\normalsize
\begin{adjustbox}{width=0.975\textwidth}
\begin{tabular}{c m{0.15\linewidth} m{0.375\linewidth} | C{0.175\linewidth} | C{0.065\linewidth} | C{0.075\linewidth} | C{0.075\linewidth} | C{0.14\linewidth}}
\toprule\toprule

& \textbf{Paradigm}  & \textbf{Examples} & \textbf{Mechanism} & \textbf{Scope} & \textbf{Time} & \textbf{Access} & \textbf{Presentation} \\\midrule
\midrule

\S\ref{sub:features} & Feature \qquad Attributions & Perturbations, Gradients, Propogations, Surrogate (\textbf{LIME/SHAP}), Attentions & \mechinputoutput{} \mechconceptoutput & \localexp & \posthoc & \modelspecific{} \modelagnostic & \prescores{} \previs \\\midrule

\S\ref{sub:probing} & Probing & Probing and \textbf{Clustering} & \mechinputinternal & \globalexp &  \posthoc & \modelspecific & \prescores \\\midrule

\S\ref{sub:mechanistic} & Mechanistic \qquad Interpretability & Stimuli, Sparse Autoencoders, Patching, Scrubbing, Logits lens & \mechinternalinternal &  \globalexp & \posthoc & \modelspecific & \previs{} \quad \pretext \\\midrule

\S\ref{sub:diagnostic} & Diagnostic Sets & Challenge/Probing sets, Test suites  & \mechinputoutput & \globalexp & \posthoc & \modelagnostic & \prescores \\\midrule

\S\ref{sub:counterfactuals} & Counterfactuals & Contrastive examples, \textbf{Adversarial attacks}, Concept counterfactuals & \mechinputoutput{} \mechconceptoutput & \localexp{} \globalexp & \posthoc & \modelspecific{} \modelagnostic & \prescores{} \preexamples \\\midrule

\S\ref{sub:language_exp} & Natural Lang. Explanations & Extractive, Abstractive, explain-then-predict, predict-and-explain, CoT & \mechinputoutput &  \localexp & \intrinsic & \modelspecific & \pretext \\\midrule

\S\ref{sub:self_explain} & Self-explaining Models & \textbf{Classic ML}, Concept bottleneck, KNN-based, Neural module nets & \mechinputoutput{} \mechinputconceptoutput & \localexp{} \globalexp & \intrinsic & \modelspecific & \prescores{} \preexamples{} \pretext \\
\bottomrule\bottomrule
\end{tabular}
\end{adjustbox}
\caption{Overview of the interpretability paradigms discussed in this paper, categorised by their \textit{what} and \textit{how} properties (\S\ref{sec:properties_discussion}). A detailed survey of these paradigms is provided in \S\ref{sec:methods}. \textbf{In bold}, methods (SHAP, LIME, Clustering, Adversarial Attacks, Classic ML) that were analyzed separately of their paradigm in our trend analysis in \S\ref{sec:trends}.}
\label{tab:categorization}
\vspace{-0.8em}
\end{table*}

\subsection{What is an Interpretability Method?}
\label{sub:interpretability_method}

In the AI literature, the terms \textit{interpretability} and \textit{explainability} are often subjects of debate, and there is no clear consensus on their definitions \citep{doshi2017towards, Lipton18, Krishnan2019AgainstIA}. While these terms are used interchangeably in much of the NLP literature \citep{Jacovi2020, Lyu2022, ZhaoCYLDCWYD24}, many papers in the XAI literature distinguish between the two \citep{Rudin2018, ArrietaRSBTBGGM20}, see our note in \S\ref{sub:posthoc}. Moreover, within this broad umbrella of model interpretability, the NLP literature also discusses model analysis \citep{BelinkovG19, MosbachGBKG24}. 

For the purposes of this paper, we embrace a broad perspective and define both \textit{interpretability and explainability methods} as:
\begin{interpretabilitybox}
\textbf{\textcolor{Emerald!55!black}{Interpretability Method}} \\
\textit{Any approach that extracts insights into a mechanism of the NLP system.}
\end{interpretabilitybox}
We justify this broad definition, which explicitly encompasses model analysis, because our paper focuses on the perspective of stakeholders for whom, to some extent, analysis alone may suffice to achieve their objectives. For instance, a regulator might only need to ensure that model performance does not significantly differ between two subpopulations. This does not necessarily demand that the interpretation elucidate the precise cause of each decision. Moreover, our broad definition does not restrict the interpretability method to explain the full system, but rather, only a mechanism within it. For example, developers might want to gain insights about specific components of the system to improve or modify their functionality.

\subsection{What is an Explanation?}
\label{sub:explanation}

\citet{Miller2017ExplanationIA} and \citet{Lipton18} rightfully emphasize that interpretability should not be confused with an explanation. \citet{Miller2017ExplanationIA} distinguishes between (causal) attributions and (causal) explanations. Attribution involves extracting relationships and causes, but it is not necessarily an explanation, even if a person could use attributions to derive their own explanation. Explanation also involves selecting, contextualizing, and presenting causes and relationships to the stakeholders. Thus, explanations are about communicating insights in a way that aligns with human cognitive biases and social expectations. In some sense, the output of interpretability methods is an attribution. 

Most existing work in the NLP literature is on \textit{how we extract} insights and not about \textit{communicating} them. Since this paper is directed at this community rather than the HCI or XAI communities, we mostly focus on interpretability methods. However, to begin the discussion about the \textit{what} and \textit{how} parts (see the paragraph below the following definition), we must first define an explanation.
This is because the \textit{what} and \textit{how} are derived from the \textit{why} -- the stakeholders, and clearly, they are part of an explanation.
To this end, we have gathered common (though not formal) definitions from seminal works in the literature \citep{doshi2017towards, Lipton18, murdoch2019definitions, ArrietaRSBTBGGM20, Lyu2022, RaukerHCH23}, and propose the following definition:
\begin{explainingbox}
\textbf{\textcolor{RoyalBlue!55!black}{Explanation (explaining):}} \\
\textit{Extracting insights into a mechanism of the NLP system and communicating them to the stakeholders in understandable terms.}
\end{explainingbox}
We define and elaborate on the \textbf{\textcolor{OliveGreen!85!black}{mechanism}} and \textbf{\textcolor{Peach!75!black}{understandable terms}} aspects of the above definition in Appendix \S\ref{sec:mechanism_terms_app}. 
These two aspects are related to the \textit{what} part: \textit{what mechanism are we interpreting, what terms are we using to describe its states, and what is the scope of the explanation?} 

Conversely, the \textbf{\textcolor{Emerald!70!black}{extracting}} and \textbf{\textcolor{DarkOrchid!85!black}{communicating}} aspects are related to the \textit{how} part: \textit{how are we interpreting and extracting insights and how are we presenting and communicating insights?}
Note that \textbf{\textcolor{Emerald!70!black}{extracting}} is essentially the interpretability method defined in \S\ref{sub:interpretability_method}. 

To summarize, an interpretability (or explainability or analysis) method extracts insights from a model, whereas an explanation involves communicating these insights to stakeholders. This process includes filtering and selecting relevant insights, processing them, and presenting them in an understandable terms. For example, computing SHAP values is an interpretability method, while visualizing these values using the SHAP Python package\footnote{\url{https://shap.readthedocs.io}} and providing guidance on interpreting these visualizations constitute an explanation.

\section{Properties and Categorization}
\label{sec:properties_discussion}

In this section, we propose and describe properties that answer the \textit{what} and \textit{how} questions derived from our interpretability definitions. We aim to provide the stakeholders' perspective, deepening our understanding of how these properties align with their objectives and requirements. We begin by discussing the \textit{what} aspect properties in \S\ref{sub:what_properties}, followed by the \textit{how} aspect properties in \S\ref{sub:how_properties}.

In Table~\ref{tab:categorization}, we present a categorization of interpretability paradigms based on the properties. For the reader's convenience, we briefly describe each property in Appendix \S\ref{sec:properties}.

\subsection{\textit{What} Properties}
\label{sub:what_properties}

\subsubsection{The Explained Mechanism}
\label{sub:mechanism}

In Appendix \S\ref{sub:def_mechanism}, we formally define what a \textit{mechanism} is. Broadly, a mechanism can refer to the entire NLP system or a specific process or component within it. To better categorize interpretability methods, we distinguish four types of mechanisms. While most methods explain the whole system (an \cmechinputoutput{input-output} mechanism), other methods explain input representations or hidden states (an \cmechinputinternal{input-internal} mechanism). Another mechanism type focuses on explaining the functionality of internal components such as neurons, attention heads, circuits, and more (an \cmechinternalinternal{internal-internal} mechanism). 

In addition, the mechanism property covers any abstraction of the mechanism states (see \S\ref{sub:terms}), for example, explaining the impact of concepts conveyed in the text input instead of explaining long and complex raw input. In this case, which is thoroughly discussed in the next subsection \S\ref{sub:concepts}, the explained mechanism is \cmechconceptoutput{concept-output}.

\textit{The choice of which mechanism to explain depends on the why}: the objective of the explanation and the stakeholder's needs. 
Stakeholders mostly utilize methods that explain the full system (an \textit{input-output} mechanism). However, many are interested in other mechanisms. For example, developers aim to understand the functionality of internal components such as neurons or layers to modify and edit factual knowledge encoded by them \citep{HaseBKG23}. Scientists might explore the representational space, for example, neuroscientists examine the brain by aligning model representations with brain activity \citep{Tikochinski2024IncrementalAO}, and social scientists cluster representations to monitor opinions, such as attitudes towards COVID-19 vaccines \citep{HristovaN22}.

\subsubsection{Raw Input or Abstracted Input}
\label{sub:concepts}

A common interpretability paradigm is feature attributions, where each input feature is assigned an importance score reflecting its relevance to the model prediction. In computer vision, the raw inputs consist of pixels, and feature attributions effectively highlight relevant areas that can be immediately and intuitively grasped \citep{AlqaraawiSWCB20, Muller2024}. 
In contrast, explaining the raw input in NLP, often a lengthy and complex text, presents distinct challenges. For end-users, assigning scores to each token can be overwhelming as the cognitive load increases with the text length. 

Instead, simplifying the system by abstracting the input to concepts or a summary, thus reducing the number of features explained, could lead to a better mental model of the system \citep{Poursabzi21}.
For example, concept counterfactual methods (see \S\ref{sub:counterfactuals}, \citet{FederOSR21} and \citet{gat2023faithful}) change a specific concept conveyed in the text. By contrasting the counterfactual predictions with the original prediction, we can gain digestible insights into how the concept impacts the prediction (a \cmechconceptoutput{concept-output} mechanism).
Moreover, due to the vast space of textual data, providing global explanations by explaining the raw input is challenging. 
In contrast, concept-level explanations naturally support global explanations.

\subsubsection{Scope: Local or Global}
\label{sub:local}
This categorization is based on the scope of the explanation: \clocalexp{local} or \cglobalexp{global}. A local explanation describes the mechanism for an individual instance. For example, feature attributions and attention visualizations (\S\ref{sub:features}). Conversely, global explanations describe the mechanism for the entire data distribution, for example, probing (\S\ref{sub:probing}) and mechanistic interpretability (\S\ref{sub:mechanistic}). 
Many local explanations can be generalized into global ones. For example, concept counterfactuals (\S\ref{sub:counterfactuals}) measure the causal effect of a concept on the prediction of an individual instance. A global average causal effect estimation can be derived by iterating the entire dataset and applying adjustments \citep{gat2023faithful}. 

The choice of scope, local or global, depends on the objectives of the explanation and its stakeholders. For instance, developers debugging edge cases may prefer local explanations. Conversely, when aiming to improve the functionality of model components, developers might lean towards global explanations offered by mechanistic interpretability. End-users, such as clients and customers, require local explanations since they are concerned with decisions directly affecting them; this local need is also reinforced by the ``right to explanation'' \citep{right_to_explanation}. Similarly, physicians using NLP systems must rely on local explanations. On the other hand, business decision-makers and scientists generally favor global explanations, which help identify broader trends and underlying patterns. From a social perspective, global explanations hold more significance. However, accumulating local evidence can progressively provide insights into global tendencies.

\subsection{\textit{How} Properties}
\label{sub:how_properties}

\subsubsection{Time: Post-hoc or Intrinsic}
\label{sub:posthoc}

This property distinguishes between methods based on the time the explanation is formed. \cposthoc{Post-hoc} methods produce explanations after the prediction and are typically external to the explained model. Conversely, \cintrinsic{intrinsic} methods are built-in; the explanation is generated during the prediction, and the model relies on it. Intrinsic methods include, for example, natural language explanations (\S\ref{sub:language_exp}) or self-explaining models (\S\ref{sub:concepts}) such as concept bottleneck models, which train a deep neural network to extract human-interpretable features, which are then used in a classic transparent model (e.g., logistic regression).

In the XAI literature, this distinction also defines the difference between explainable AI (post-hoc) and interpretable AI (intrinsic) \citep{Rudin2018, ArrietaRSBTBGGM20}. However, interpretable AI generally refers to transparent models (see \citep{Lipton18}), while in our categorization, intrinsic models can be opaque to some extent: in self-explaining methods, an opaque neural network extracts human-interpretable features; similarly, in natural language explanations, the explanation is generated by an opaque neural network.
Intrinsic methods aim to produce more faithful and understandable insights and could better serve all stakeholders. However, they may also limit model architecture and thus could potentially degrade system performance, although this is not always the case (see \citet{Badian2023SocialMI} for an example).

\subsubsection{Access: Model Specific or Agnostic}
\label{sub:specific_or_agnostic}

This property distinguishes interpretability methods based on their access to the explained model. \cmodelagnostic{Model-agnostic} methods do not assume any specific knowledge about the model and can only access its inputs and outputs. For example, diagnostic sets (\S\ref{sub:diagnostic}), perturbation-based attributions (\S\ref{sub:features}), or some counterfactual methods (\S\ref{sub:counterfactuals}). 
The latter two modify only the input and measure its impact on model prediction. 
On the other hand, \cmodelspecific{model-specific} methods require access to the explained model during the training time of the interpretability method. They can also access its internal components and representations. Hence, while a model-specific method can be applied only to one explained model, the same model-agnostic method can be applied to any model simultaneously. 

Unlike model-specific methods, model-agnostic methods can not explain internal mechanisms.
However, they can still be extremely valuable for some stakeholders. From an algorithmic perspective, they are useful during model selection and deployment. For example, developers juggling multiple models can easily rank them based on their vulnerability to confounding biases, such as gender bias. Moreover, regulators would prefer model-agnostic methods, utilizing a dedicated diagnostic set or a pool of counterfactuals to verify whether the model meets the required standards.

\subsubsection{Presenting Insights}
\label{sub:presenting}

The presentation of insights extracted by the interpretability method falls under the \textit{communicating} aspect of the explanation definition in \S\ref{sub:explanation}. There is extensive research in the XAI field that explores this aspect and examines how the presentation affects different stakeholders \citep{HohmanHCDD19, Schulze-Weddige21, BoveALTD22, Karran2022DesigningFC, ZytekLVV22}. Even though we do not delve into the stakeholder perspective, we still discuss this property since not all methods support every form of presentation. The design of interpretability methods and the choice of which to use depend on it.

The most common form of presentation is \cprescores{scores}, such as importance scores  (\S\ref{sub:features}), causal effects (\S\ref{sub:counterfactuals}) or metrics (\S\ref{sub:probing} and \S\ref{sub:diagnostic}). Scores are typically visualized using colors \citep{GatCRH22} or bar plots \citep{KokaljSLPR21}.
Another form is \cprevis{visualization}, which includes means such as heatmaps \citep{JoM20}, graphs \citep{Vig19}, and diagrams \citep{KatzB23}.
Others present similar or contrastive \cpreexamples{examples} to stakeholders, along with their prediction, aiding in speculating on \textit{why P and not Q?}. 
Such example presentations are found in counterfactual methods (\S\ref{sub:counterfactuals}) and KNN-based nets (\S\ref{sub:self_explain}). Finally, insights can also be conveyed through \cpretext{texts} written in natural language (e.g., \citet{menon-etal-2023-mantle} and \S\ref{sub:language_exp}).

\subsubsection{Faithfulness and Causality}
\label{sub:causality}

Note that some applications of interpretability methods are satisfied by correlational insights (\textit{what knowledge the model encodes}), e.g., in a case when scientists explore new hypotheses which will then be validated in a controlled experiment (see \citep{Lissak2024}). However, most applications seek to understand the reasons behind specific predictions. In this context, faithfulness becomes a crucial principle, demanding that explanations accurately reflect the system’s decision-making process \citep{Jacovi2020}. 
Unfaithful explanations, particularly those that seem plausible, can be misleading and dangerous and lead to potentially harmful decisions. As such, faithfulness is crucial in scenarios involving decision-makers and end-users. To ensure that explanations are faithful, establishing causality is essential \citep{FederKMPSWEGRRS22}. Indeed, \citet{gat2023faithful} theoretically demonstrated that non-causal methods often fail to provide faithful explanations.

A key approach to providing faithful explanations involves incorporating techniques from the causal inference literature, such as counterfactuals \citep{FederOSR21}, interventions \citep{WuGIPG23}, adjustment \citep{Wood-DoughtySD18}, and matching \citep{ZhangKSMMK23}. Therefore, an important property of an interpretability method is whether it is \ccausalbased{causal-based} or \cnotcausal{not}. We note that this categorization is not included in Table~\ref{tab:categorization} as it pertains more to specific methods rather than to a paradigm. For a comprehensive survey on faithfulness in NLP interpretability, see \citet{Lyu2022}.
\section{Common Interpretability Paradigms}
\label{sec:methods}

This section aims to establish a clear link between the properties introduced in \S\ref{sec:properties_discussion} and interpretability methods. 
To this end, we comprehensively review common interpretability paradigms, detailing relevant methods and works within each and explaining the paradigm's properties. Note that some methods may fall under multiple paradigms. 

Our classification of methods into paradigms is inspired by previous surveys on model analysis \citep{BelinkovG19}, local methods \citep{LuoIHP24}, post-hoc methods \citep{MadsenRC23}, faithful methods \citep{Lyu2022}, mechanistic interpretability \citep{RaukerHCH23, Bereska2024}, LLMs \citep{Singh2024, ZhaoCYLDCWYD24}, and others \citep{Danilevsky2020, Balkir2022, sajjad-etal-2022-neuron}. Furthermore, while the categorization of the properties captures the standard characterization each paradigm, there may be exceptions with some methods. 

\subsection{Feature Attributions}
\label{sub:features}

\begin{tcolorbox}[enhanced,breakable,colback=white, colframe=black!75!white,title=\textbf{Categorization},size=fbox, boxrule=1pt]
What: \mechinputoutput{} or \mechconceptoutput{}, \localexp{} \\ 
How: \posthoc{}, \modelspecific{} or \modelagnostic{}, \prescores{}
\end{tcolorbox}

Feature attribution methods measure the relevance (sometimes referred to as importance) of each input feature, primarily tokens or words, and are a widely used \clocalexp{local} interpretability paradigm. Each input feature is assigned a \cprescores{score} reflecting its relevance to a specific prediction, thus describing an \cmechinputoutput{input-output} mechanism. Various attribution methods have been developed, which can be mainly categorized into four types.

\textit{Perturbation-based} methods work by perturbing input examples, such as removing, masking, or altering input features at various levels, including tokens, embedding vectors, or hidden states \citep{WuCKL20, LiMJ16a}. Those are \cmodelagnostic{model-agnostic} methods since the perturbations are applied to the input. In contrast, the following methods are \cmodelspecific{model-specific}: \textit{Gradient-based} methods measure relevance via a regular backward pass (backpropagation) from the output through the model \citep{SmilkovTKVW17, SikdarBH20, GatCRH22, Enguehard23}. \textit{Propagation-based} methods define custom rules for different layer types \citep{MontavonBLSM19, VoitaST20, CheferGW21}. Other methods involve \textit{surrogate models}, such as LIME \citep{Ribeiro0G16} and SHAP \citep{LundbergL17}, which locally approximate a black-box model with a white-box surrogate model \citep{KokaljSLPR21, MoscaSTGG22}. Rarely, the features are mapped into concepts \citep{YehKALPR20}, describing a \cmechconceptoutput{concept-output} mechanism. 

We also include here \textit{attention-based} explanations, which aim to capture meaningful correlations between intermediate states of the instance \citep{JainW19, KovalevaRRR19, WiegreffeP19}, because typically the intermediate state is represented by its corresponding token. Usually, attention-based explanations are presented with visualizations such as heatmaps.

\subsection{Probing and Clustering}
\label{sub:probing}

\begin{tcolorbox}[enhanced,breakable,colback=white, colframe=black!75!white,title=\textbf{Categorization},size=fbox, boxrule=1pt]
What: \mechinputinternal{}, \globalexp{} \\
How: \posthoc{}, \modelspecific{}, \prescores{} or \pretext{}
\end{tcolorbox}
Probing typically involves training a classifier that takes the representations of the explained model and predicts some property \citep{Belinkov22}, making it a \cposthoc{post-hoc} \cmodelspecific{model-specific} method. Typically, the predicted concept is a syntactic or semantic property \citep{AdiKBLG17, BaroniBLKC18, HewittL19, LeporiM20, RavichanderBH21, AntvergB22, amini-etal-2023-naturalistic, VulicGLCPK23}. Probing methods usually answer questions of how extractable a property is from a representation or what knowledge a model encodes. Thus, it can \cglobalexp{globaly} describe the \cmechinputinternal{input-internal} mechanism. However, even though the model encodes some property, it does not mean it uses it for prediction \citep{Belinkov22}. Therefore, how we communicate probing insights to the stakeholders is important.

In the scope of probing, we also include \textit{clustering} methods. While most clustering methods are used to discover patterns in data, here, clustering is employed to explore the model's learned space and gain insights about what it has encoded. Clustering is considered the unsupervised counterpart of probing \citep{MichaelBT20, GuptaSGS22}, and they share the same categorization: both methods explore the \textit{input-internal} mechanisms of the system and are characterized as \textit{global}, \textit{post-hoc}, and \textit{model-specific}. After representations are clustered, explanations are provided through cluster descriptions defined by gold labels, top keywords, concepts, topic modeling, ontologies, or LLM-generated text \citep{AharoniG20, ZhangFCN22, Thompson2020, GuptaSGS22, SajjadDDAK022, AlamDDSK023, MousiDD23, WangSZ23, HawaslyDD24, Lissak2024}. Finally, we also include works that explain representation-based similarity using concepts and semantic aspects \citep{opitz-frank-2022-sbert}.

\subsection{Mechanistic Interpretability}
\label{sub:mechanistic}

\begin{tcolorbox}[enhanced,breakable,colback=white, colframe=black!75!white,title=\textbf{Categorization},size=fbox, boxrule=1pt]
What: \mechinternalinternal{}, \globalexp{} \\
How: \posthoc{}, \modelspecific{}, \previs{} or \pretext{}
\end{tcolorbox}

In contrast to probing, which is a top-down approach (i.e., we know in advance what we are looking for), mechanistic interpretability is a bottom-up approach that studies neural networks through analysis of the functionality of internal components of the NLP systems such as neurons, layers, and connections \citep{sajjad-etal-2022-neuron, RaukerHCH23, Bereska2024}. The goal of such methods is to \cglobalexp{globaly} explain one \cmechinternalinternal{internal-internal} mechanism of a \cmodelspecific{specific} model. Many mechanistic interpretability methods study how neurons respond to stimuli (real or synthetic examples) and \cprevis{visualize} or \cpretext{describe} the sensitivity of the neuron's activations \citep{FinlaysonMGSLB20, VigGBQNSS20, GeigerLIP21, GeigerWLRKIGP22, DaiDHSCW22, ConmyMLHG23, Garde2023, Gurnee2024}.

Other works perturb or intervene in neurons to study their functionality \citep{BauBSDDG19, GhorbaniZ20, WangVCSS23}, or mask network weights \citep{ZhaoLMJS20, CsordasSS21}. Some works focus on gradients instead of activations \citep{DurraniSDB20, Syed2023, Kramar2024} or train sparse autoencoders in an attempt to disentangle features, which are then described \citep{Cunningham2023, YuMP23}. 
Another line of work explores which information the internal states encode by projecting them into the vocabulary \citep{GevaCWG22, DarG0B23, Belrose2023, PalSYWB23, Sakarvadia2023, Ghandeharioun2024} or even by generating images \citep{Toker2024}. 

\subsection{Diagnostic Sets}
\label{sub:diagnostic}

\begin{tcolorbox}[enhanced,breakable,colback=white, colframe=black!75!white,title=\textbf{Categorization},size=fbox, boxrule=1pt]
What: \mechinputoutput{}, \globalexp{} \\
How: \posthoc{}, \modelagnostic{}, \prescores{}
\end{tcolorbox}

Diagnostic sets, also known as challenge sets, probing sets, or test suites, are specialized collections of data designed to analyze specific properties of the NLP system or challenging cases. These sets are typically curated manually to target specific aspects of system behavior within a predefined NLP task, enabling the identification of strengths, weaknesses, and biases \citep{BelinkovG19}.
Diagnostic sets are \cmodelagnostic{model-agnostic} since they are curated independently from the analyzed model. They support \cprescores{scoring} the model's predictive capabilities (\cmechinputoutput{input-output} mechanism) on subpopulations of interest, providing \cglobalexp{global} insights on how it works within them.
As one of the oldest techniques for analyzing NLP systems \cite{KingF90, LehmannORNLKFFEDCBBA96}, diagnostic sets have been reintroduced as essential tools for understanding NLP models \citep{HillRK15, LeviantR15, WangSMHLB19, VulicBPPLWMBMPR20, WangPNSMHLB19, Gardner2020} and LLMs \citep{Srivastava2022, McKenzie2023, Laskar2024ASS}. Rarely, diagnostic sets can be \cmodelspecific{model-specific}. For example, the diagnostic dataset curated by \citet{Gekhman24} involves examples not included in a specific LLM's pre-existing knowledge. Fine-tuning the same LLM using these examples increases hallucinations. 

Many diagnostic sets are employed to examine linguistic phenomena \citep{BurchardtMDHPW17, BurlotY17, Sennrich17, WhiteRDD17, GiulianelliHMHZ18, GulordavaBGLB18, JumeletH18, RavichanderHSTC20, NewmanAGH21, Sullivan24}, while others evaluate biases such as gender bias \citep{WaseemH16, WebsterRAB18, ZhaoWYOC18, De-ArteagaRWCBC19, DhamalaSKKPCG21, Doughman2022}, cultural bias \cite{Ventura2023, Chiu2024, Rao2024}, and political bias \citep{SmithHKPW22, Taubenfeld2024}. Beyond manually collecting diagnostic datasets or using simple rule-based programs, generative models are also being applied \citep{GoelRVTBR21, RibeiroWG021, RossWPPG22}. Importantly, these sets are crucial not only for evaluating the performance of NLP systems on specific examples or subpopulations but also serve as foundational elements in many probing and mechanistic interpretability methods.

\subsection{Counterfactuals and Adversarial Attacks}
\label{sub:counterfactuals}

\begin{tcolorbox}[enhanced,breakable,colback=white, colframe=black!75!white,title=\textbf{Categorization},size=fbox, boxrule=1pt]
{\normalsize What: \mechinputoutput{} or \mechconceptoutput{}, \\ 
\hspace*{0.9cm} \localexp{} or \globalexp{} \\
How: \posthoc{}, \modelagnostic{} or \modelspecific{}, \\
\hspace*{0.9cm} \prescores{} or \preexamples{}}
\end{tcolorbox}

The term \textit{counterfactual (CF)} is frequently used in the NLP literature, often referring to various concepts. In this subsection, we aim to align the community's understanding of this term and clearly distinguish between CF-based methods. In the context of NLP, we adopt the following definition, which captures the fundamental characteristic common to all CF-based methods:
\noindent\textit{``a counterfactual for a given textual example is a result of a targeted intervening on the text while holding everything else equal.''} \citep{CalderonBFR22, gat2023faithful}. 
The primary distinction among CF-based methods lies in the type of question the CFs aim to answer.

From a philosophical perspective, CFs answer \textit{what-if} questions: \textit{`If X had been different, then Y would be...'}. Presenting an \cpreexamples{alternation (CF)} of the input example to stakeholders allows for speculation on the \cmechinputoutput{input-output} mechanism: \textit{`Why prediction A and not B?'} \citep{Miller2017ExplanationIA, WuRHW20}.

From a causal inference perspective, CFs answer questions such as \textit{`How does C impact Y?'}, which can then help derive a \cprescores{score} quantifying the causal effect of some concept \textit{C} on the prediction: a \cmechconceptoutput{concept-output} mechanism \citep{AbrahamDFGGPRW22, FederKMPSWEGRRS22, WuDGZP23}.

\medskip\noindent\textbf{Contrastive Examples.}
These methods address \textit{what-if} questions and can explain a \clocalexp{local} prediction by \cpreexamples{presenting CFs} to stakeholders. They typically focus on minimally editing the text to change the model prediction. The edited texts are commonly known as \textit{contrastive examples}. Most approaches for generating contrastive examples are \cmodelagnostic{model-agnostic}. For instance, asking annotators to write them manually \citep{Gardner2020, KaushikHL20, SenASAA023}, utilizing a generative model and applying edit operations \citep{WuRHW20, RossWPPG22, Li2024, Nguyen2024}, or generating text until a proxy predictor indicates the label has changed \citep{RossMP21, ChemmengathALD22, FilandrianosDMZ23, TrevisoRGM23, Bhan2024MitigatingTT}.

\medskip\noindent\textbf{Adversarial Attacks.} A prominent \cmodelspecific{model-specific} approach for generating contrastive examples is known as \textit{adversarial attacks}, in which carefully crafted modifications barely noticeable to humans (e.g., a typo, extra space, or punctuation, etc...) are applied to the input and change the system predictions \citep{MorrisLYGJQ20, GoyalDKR23}. These attacks are typically generated through gradient-based token replacement \citep{EbrahimiRLD18, LiJDLW19, GuoSJK21}, and character-level perturbations \citep{BelinkovB18, YangCHWJ20, Rocamora2024RevisitingCA}. With LLMs, the focus is on adversarial prompts that break model alignment \citep{PerezHSCRAGMI22, ZhuDAN, Samvelyan, Paulus2024}. Note that most applications of contrastive examples in the NLP literature, particularly adversarial attacks, are for data augmentation to improve model generalization or red teaming \citep{Chen2021AnES, KaushikSHL21, DixitPHZ22, Balashankar00PT23, ZhaoPDYLCL23, SachdevaTG24, ZhangYZWY0TL24}.

\medskip\noindent\textbf{Concept Counterfactuals.} The second group of CF-based methods, which address \textit{How does C impact Y?} questions, is more theoretically grounded in the causal inference literature, making them more faithful \citep{Lyu2022, gat2023faithful}. Besides \cpreexamples{presenting} stakeholders with explanations similar to contrastive examples, which allows for speculation on what would have happened if a concept \textit{C} were different (e.g., a different gender of the writer), concept CFs can also be used to estimate the causal effect of high-level concepts on model predictions \citep{AbrahamDFGGPRW22, FederKMPSWEGRRS22}. This is typically done by calculating the difference between the model’s predictions for the original text and the counterfactual (CF) input.

In addition to providing a \clocalexp{local} \cprescores{score} for an individual instance, concept CFs can deliver a \cglobalexp{global} \cprescores{average causal effect} estimation by iterating through the entire dataset and applying certain adjustments \citep{gat2023faithful}. The objective of the global score, similar to diagnostic sets, is to examine model behavior on subgroups. However, the score derived from CFs offers greater fidelity by relying on causation rather than correlation \citep{Elazar2022, KeidarOJS22, LiLSDSLJJL22, Liu2023, WangMWZC23, MadaanHY23, Zhou023, Elazar0WZS24}.

Typically, a causal graph describing the input and output data-generating processes is provided, and an approximated counterfactual (CF) is generated by intervening on the concept of interest and adjusting for confounders \citep{FederOSR21, gat2023faithful}. \cmodelagnostic{Model-agnostic} methods focus on generating coherent, human-like CFs, either through controlled text generation \citep{CalderonBFR22, FangCBF23, HongBMAP23, HowardSLCS22, ZhengSVFB23} or by prompting LLMs \citep{gat2023faithful, FederWSSB23, ZhangFDA24}. An alternative to the computationally intensive generation process is \textit{causal matching}, where the example is paired with a similar control example that has a different concept value \citep{Roberts2020AdjustingFC, ZhangKSMMK23, gat2023faithful}. In contrast, \cmodelspecific{model-specific} methods typically intervene on the latent space of the explained model \citep{RavfogelEGTG20, FederOSR21, ElazarRJG21, HaghighatkhahFS22, KumarT022, WuDGZP23, ZhaoSF24}, or train a proxy model that mimics the CF behavior of the explained model \citep{WuDGZP23}.

\subsection{Natural Language Explanations}
\label{sub:language_exp}

\begin{tcolorbox}[enhanced,breakable,colback=white, colframe=black!75!white,title=\textbf{Categorization},size=fbox, boxrule=1pt]
What: \mechinputoutput{}, \localexp{} \\
How: \intrinsic{}, \modelspecific{}, \pretext{}
\end{tcolorbox}

We define \textit{Natural Language Explanations (NLE)} as any \cpretext{textual explanation} extracted or generated by an NLP system that is used for justifying its own prediction. We do not consider generative models used to explain other model predictions as an NLE method. Thus, all NLE methods are \cmodelspecific{model-specific}, \cintrinsic{intrinsic}, and \clocalexp{local} as they explain a single prediction. Usually, human-written explanations are used as an additional training signal for supervision \citep{WiegreffeM21, SunSMM22, KimJKJYSS23}. 

NLE can be \textit{abstractive} (by generating free-text) or \textit{extractive} (by highlighting spans of relevant text in the input). The term \textit{rationale} is often used in the extractive context to describe short and sufficient input spans for making a correct prediction \citep{ZaidanEP07}. In addition, and following \citet{CamburuRLB18, KumarT20, Lyu2022}, we divide NLE into \textit{explain-then-predict} and \textit{predict-and-explain} methods. 

The \textit{explain-then-predict} category comprises methods that extract or generate an explanation and then independently predict the output by conditioning solely on the explanation, typically by training explainer and predictor components separately \citep{LeiBJ16, BastingsAT19, CamburuSMLB20, JainWPW20}. 
The \textit{predict-and-explain} category includes methods that explain and predict simultaneously (i.e., the output is predicted based on both the input and the explanation, such as chain-of-thoughts (CoT)) or first predict and then provide an explanation \citep{LingYDB17, RajaniMXS19, Narang2020, MarasovicBDP22}, including explanations that reflect uncertainty \citep{Xiong2023, Zhou2024}. This category covers all the recent and commonly used CoT methods \citep{Chu2023, LyuHSZRWAC23}.

In the era of LLMs, which are used daily by numerous end-users, NLE (either through CoT or explicitly asking the LLM to explain its output) has become the de facto method for explaining LLM outputs, despite being considered unfaithful \citep{Lanham2023, TurpinMPB23}. Moreover, NLE helps address challenges in explaining generative models since many interpretability methods were designed to explain a single decision rather than a sequence of decisions (a generated text).

\subsection{Self-explaining Models}
\label{sub:self_explain}

\begin{tcolorbox}[enhanced,breakable,colback=white, colframe=black!75!white,title=\textbf{Categorization},size=fbox, boxrule=1pt]
What: \mechinputoutput{} or \mechinputconceptoutput{}, \\ 
\hspace*{0.9cm} \localexp{} or \globalexp{} \\
How: \intrinsic{}, \modelspecific{}, \prescores{} or \preexamples{} or \pretext{}
\end{tcolorbox}

Classic machine learning models, such as linear models, decision trees, Hidden Markov Models (HMMs), and Topic Models are often called transparent or whitebox models due to their simple structure and well-studied nature. These models represent the highest degree of self-explanation because explaining their decision-making process is relatively straightforward. Drawing inspiration from them, researchers attempt to design neural models with more structural transparency while maintaining their performance \citep{RajagopalBHT21, Das22, Su2023}.

An example is concept bottleneck models, which train a deep neural network to extract human-interpretable features and then apply a classic transparent that takes these features as an input, sometimes simultaneously. Concept bottleneck models describe relations of  \cmechinputconceptoutput{input-concepts and concepts-output}.
The interpretable features used for training the network can be manually annotated \citep{Koh2020, psychiatry, TanCWYLL24}, defined by domain experts and automatically extracted using an LLM \citep{Badian2023SocialMI}, or automatically discovered and annotated \citep{YehKALPR20, Ludan2023}. In concept bottleneck models, explanations can be \cglobalexp{global}, such as the \cprescores{linear regression weights} of concepts, or local. In the case of \clocalexp{local} explanations, they are provided with respect to the predicted concepts of a specific instance. KNN-based networks, for example, replace the final softmax classifier head with a KNN classifier at test time \citep{Papernot2018, WallaceFB18, SarwarZHDAN22}. The \clocalexp{local} explanations in KNN-based networks are \cpreexamples{example-based}.

Another prominent line of works focuses on neural module networks, which decompose the task into small interpretable steps, which are then presented to the stakeholder \citep{AndreasRDK16, HuARDS17, SantoroRBMPBL17, GuptaLR0020}. 
Similarly, other methods break down the input into ``atoms'' and then combine the atom-level solutions to reach a final decision \citep{StaceyMDR22, stacey-etal-2024-atomic}. Presenting such decompositions helps in understanding the decision-making process.

Note that models that \cpretext{extract or generate explanations} during their predictions are self-explaining models and are covered in \S\ref{sub:language_exp}.
\begin{figure}[t]
    \centering
    \includegraphics[width=0.495\textwidth]{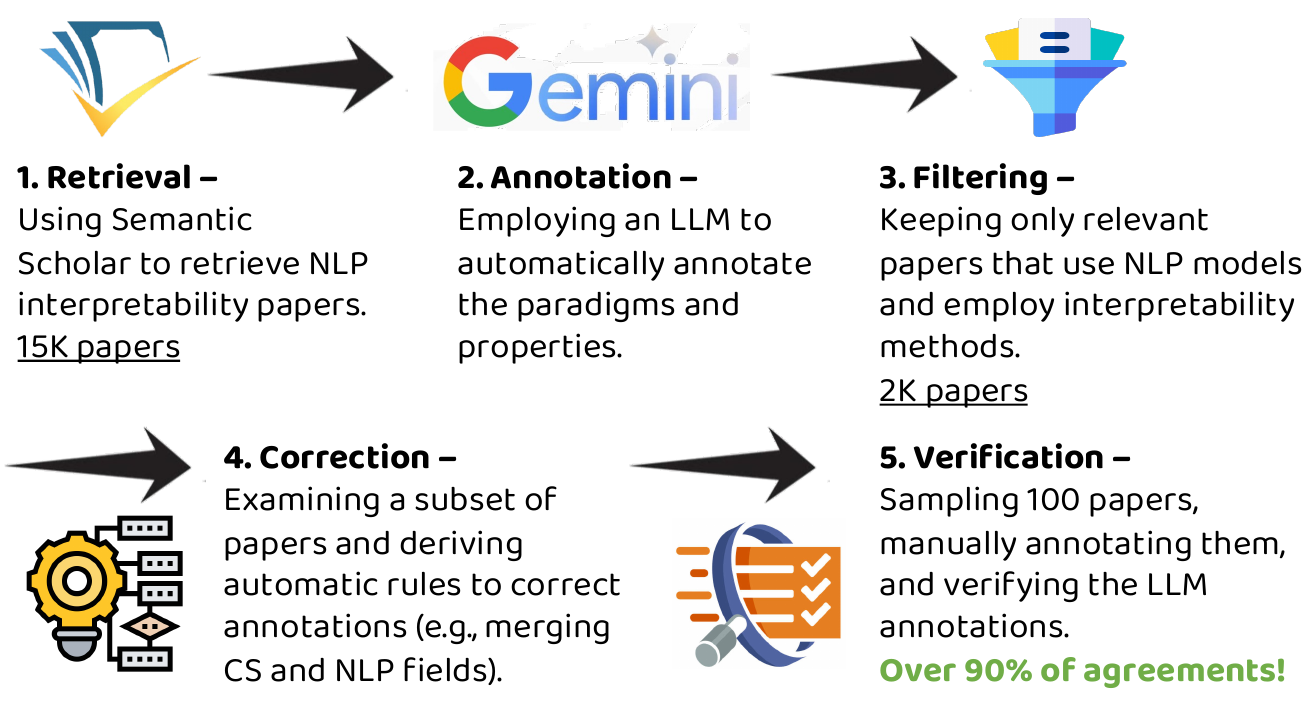}
    \caption{An illustration of our five-stage procedure for annotating NLP interpretability papers, with the stages fully detailed in Appendix \S\ref{sec:additional}.}
    \label{fig:annotations}
\vspace{-0.8em}
\end{figure}

\section{Trends in Model Interpretability}  
\label{sec:trends}

In this section, we analyze trends over the last decade in papers that propose or employ an interpretability method in the NLP field or fields outside of NLP. The analysis covers trends in interpretability method paradigms and their properties.

\subsection{Data}

Our data collection process consists of five stages and is illustrated in Figure~\ref{fig:annotations}. In the first stage, we utilized the Python client\footnote{\url{www.github.com/danielnsilva/semanticscholar}} of the Semantic Scholar API\footnote{\url{www.semanticscholar.org/product/api}} to retrieve 14,676 NLP interpretability papers by searching queries such as \texttt{NLP interpretability} (a full list of queries is provided in Box~\ref{box:queries}). 
Subsequently, we employed an LLM (\texttt{gemini-1.5-pro-preview-0514})\footnote{\url{www.ai.google.dev/\#gemini-api}} to determine the relevance of each paper based on its title and abstract. A paper is considered relevant if it relates to NLP research, employs NLP methods or models with text input, and proposes, utilizes, or discusses an interpretability method. After relevancy filtering, 2,009 papers remained (see Figure~\ref{fig:intro_papers} for their distribution across fields).

In addition, we used the LLM to annotate various attributes, including the research field, whether an LLM is employed, the paradigm of the interpretability method and its mechanism, scope, accessibility and whether it is causal-based or not. 
The zero-shot prompt is provided in Box~\ref{box:prompt}. See Appendix \S\ref{sec:additional} for additional details about our retrieval and annotation processes.

To verify the LLM annotations, we randomly sampled 100 papers, which one of the authors manually annotated. The agreement statistics are presented in Table~\ref{tab:iaa}. Notably, 96\% of the papers the LLM annotated as relevant were indeed relevant. Furthermore, over 90\% of the annotations across each property were correct. When excluding annotations labeled as `\textit{unknown}' (e.g., where the LLM indicated the method scope was unknown, but sufficient domain knowledge could infer it), over 95\% of the annotations were correct. To the best of our knowledge, this is the first paper to utilize an LLM successfully for such a task.

\begin{figure}[!t]
    \centering
    \includegraphics[width=0.485\textwidth]{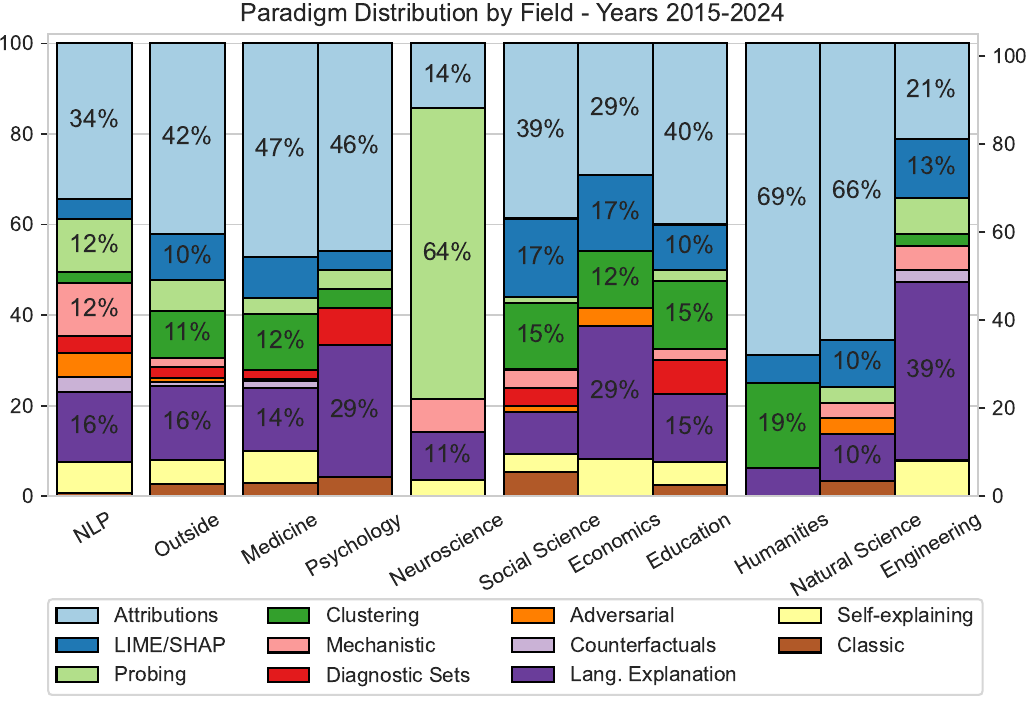}
    \caption{Distribution of NLP interpretability paradigms by research field, including papers in years 2015-24.}
    \label{fig:paradigm_field}
\end{figure}

\subsection{Results}

We present the results in the following figures and tables, all illustrating trends in the NLP field and external fields, thereby emphasizing differences between developers and non-developer stakeholders.\footnote{While developers may be stakeholders in fields outside of NLP, and vice versa, the primary distinction remains applicable. Most stakeholders in NLP are developers, while those in other fields are typically non-developers.}

\begin{figure}[t]
    \centering
    \includegraphics[width=0.485\textwidth]{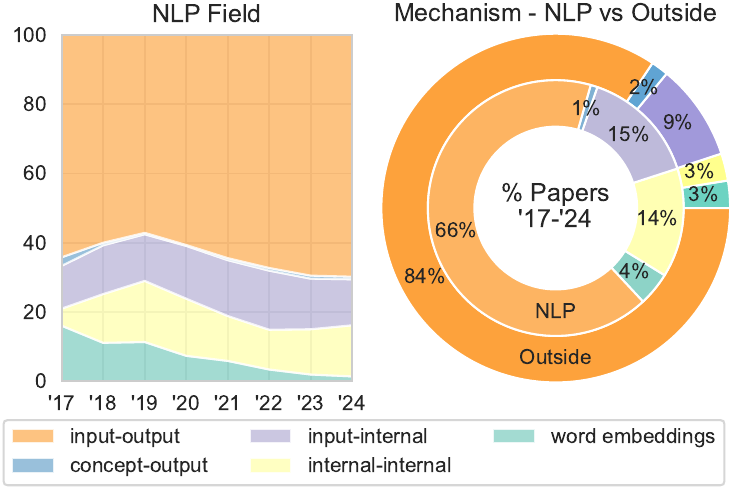}
    \caption{Trends in the explained mechanism. The left plot presents trends over time in the NLP field, showing the percentages of papers for each mechanism, as predicted by an LLM. The right plot presents pie charts with the percentage distribution of the mechanisms: the inner pie shows the distribution within the NLP field, and the outer pie shows for fields outside of NLP.}
    \label{fig:mechanism_trends}
\vspace{-0.8em}
\end{figure}

(1) Figure~\ref{fig:intro_papers} in \S\ref{sec:intro} presents the number of interpretability papers by research field and year.\footnote{Note that each year spans from June of the previous year to the following June.} (2) Figure~\ref{fig:paradigm_field} displays the distribution of interpretability method paradigms across each field, while (3) Figure~\ref{fig:paradigm_trends} illustrates trends over the last decade. (4) Figure~\ref{fig:mechanism_trends} presents the distribution of the explained mechanisms, and (5) Table~\ref{tab:properties} reports statistics on method properties. (6) Table~\ref{tab:llm_trends} emphasizes trends between papers that employ LLMs and those that do not. Finally, (7) Table~\ref{tab:paradigms_all} in the appendix provides the absolute number of papers and average citations for each paradigm.

Below, we discuss our key findings:

\begin{figure}[!t]
    \centering
    \includegraphics[width=0.485\textwidth]{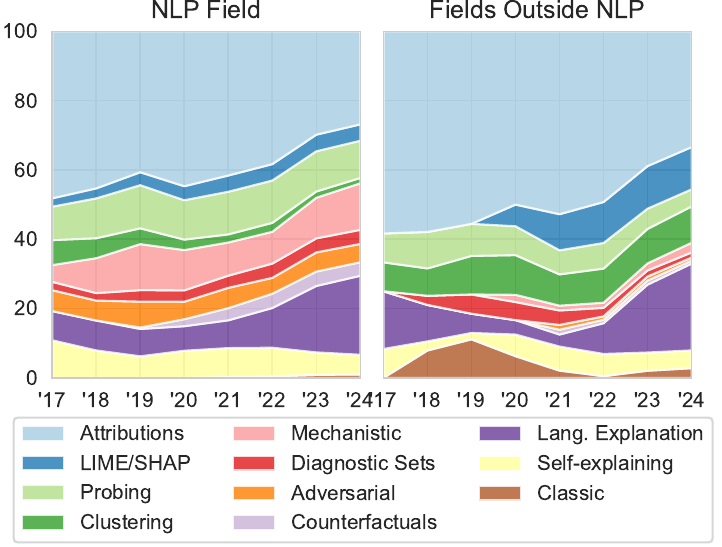}
    \caption{Trends in NLP interpretability paradigms over time in the NLP field (left plot) and in fields outside of NLP (right plot). The plots show the percentages of papers for each paradigm, as predicted by an LLM. The data smoothed using a one-year moving average.}
    \label{fig:paradigm_trends}
\vspace{-0.8em}
\end{figure}

\medskip\noindent\textbf{Inside: Stable trends in the NLP community.} 
Figure~\ref{fig:paradigm_trends} shows that paradigm trends within the NLP community are generally stable over time. However, two leading paradigms, Feature Attributions and Natural Language Explanations, demonstrate contrasting trends: the proportion of Feature Attribution papers has gradually decreased (from \textasciitilde45\% in 2017 to \textasciitilde30\% in 2024) while papers on Natural Language Explanations have increased (from \textasciitilde10\% in 2017 to \textasciitilde25\% in 2024). The latter rise is likely attributed to advancements in text generation capabilities, which will be discussed later. The next two most common paradigms—Probing and Mechanistic Interpretability, each account for about 12\% (see Figure~\ref{fig:paradigm_field}). 

Regarding the trends in mechanisms illustrated in Figure~\ref{fig:mechanism_trends}, the explanation of Word Embedding, which was very popular a decade ago, has diminished over the years. 
Currently, two-thirds of the papers explain the input-output mechanism.

\medskip\noindent\textbf{Inside vs Outside: Non-developers care less about model internals.} 
We observe notable differences when comparing paradigm distributions between the NLP field and outside of NLP. While Feature Attribution is the dominant paradigm in both, Mechanistic Interpretability and Adversarial Attacks hold a large share within NLP but are rarely seen outside of it. Conversely, Clustering and Surrogate Models (such as LIME and SHAP) are common outside of NLP but not frequently encountered in general NLP papers. 

We attribute these distinctions to two main reasons. The first reason is that non-developers care less about model internals and are more concerned with input-output mechanisms. This is evident in the right plot of Figure~\ref{fig:mechanism_trends}, where there are five times more internal-internal mechanism papers in the NLP field. Moreover, although 9\% of the papers outside of NLP explain an input-internal mechanism (representations), most involve field-specific techniques. For example, Probing is the most common paradigm in the neuroscience field (64\% of the papers, see Figure~\ref{fig:paradigm_field}), where researchers try to align model representations with brain activities \citep{goldstein2022shared, tikochinski2023perspective}.

The second reason is the ease of application and the level of support for these methods in popular code packages. These aspects are particularly important for non-developers. For instance, LIME and SHAP packages are widely used across many domains beyond NLP \citep{KaurNJCWV20}, and clustering or classic ML methods are readily available in popular data science packages like Scikit-learn.

\medskip\noindent\textbf{Outside NLP: Different fields, different needs.} 
The choice of interpretability method depends on the stakeholder's objectives and needs. Different research fields have distinct requirements, as clearly shown in Figure~\ref{fig:paradigm_field}, where paradigm distributions vary across the fields.
These differing needs are also reflected in method properties in Table~\ref{tab:properties}. For instance, in healthcare fields, local explanations are much more prominent.
This makes sense considering that the main stakeholders, patients and therapists, are interested in understanding individual decisions. 
Conversely, in neuroscience and social science, scientists aim to understand cognitive mechanisms or social phenomena, thus preferring global explanations.

\begin{table}[!t]
\centering
\normalsize
\begin{adjustbox}{width=0.475\textwidth}
\begin{tabular}{l  | C{0.08\linewidth}  C{0.11\linewidth} | C{0.11\linewidth}  C{0.15\linewidth} | C{0.08\linewidth}  C{0.08\linewidth}}
\toprule
& \multicolumn{2}{c}{\textbf{\cglobalexp{Scope}}} & \multicolumn{2}{|c}{\textbf{\cmodelspecific{Accessibility}}} & \multicolumn{2}{|c}{\textbf{\cnotcausal{Causal-based}}} \\
& \localexp & \globalexp & \modelspecific & \modelagnostic & \causalbased & \notcausal \\
\midrule
NLP  &   57.3 &    42.7 &            84.6 &            15.4 &     5.2 &        94.8 \\
Outside &   61.7 &    38.3 &            80.1 &            19.9 &     1.9 &        98.1 \\
$\hookrightarrow$ Healthcare       &   66.5 &    33.5 &            76.7 &            23.3 &     2.0 &        98.0 \\
$\hookrightarrow$ Neuroscience &   25.0 &    75.0 &            92.6 &             7.4 &     0.0 &       100 \\
$\hookrightarrow$ Social       &   57.7 &    42.3 &            76.9 &            23.1 &     2.0 &        98.0 \\
\bottomrule
\end{tabular}
\end{adjustbox}
\caption{Percentage of papers by properties (\S\ref{sec:properties}) across fields. \textit{Outside} encompasses all fields outside NLP and CS. \textit{Healthcare} includes Medicine and Psychology, while \textit{Social} includes Social Sciences, Economics and Education fields.}
\label{tab:properties}
\vspace{-0.8em}
\end{table}

\medskip\noindent\textbf{LLMs dramatically change the trends.} 
The introduction of LLMs in the last two years has drastically improved the capabilities of NLP models. These models have been widely adopted not only by NLP researchers but also by practitioners in various fields. This is evident in Table~\ref{tab:llm_trends}, where LLM papers have become prominent both within the NLP field (66.7\% of the papers in 2024) and outside of it (from 18.2\% in 2023 to 50.7\% in 2024). 

The widespread adoption of LLMs has shifted interpretability paradigms. Although paradigm trends in NLP were stable, the introduction of LLMs tripled the portion of Natural Language Explanation papers (30.8\%), likely due to the strong generation capabilities of LLMs. Outside NLP, this paradigm accounts for nearly half of the papers that employ LLMs (48.7\% compared to 6.2\% in non-LLM papers). This is another indication that non-developers favor methods that do not require advanced technical skills, as generating textual explanations can be done through simple prompting.

We anticipate more trend shifts in the LLM era, particularly toward methods that leverage strong generation capabilities, such as generating Counterfactuals and dedicated Diagnostic Sets, which is already evident in a 30\% increase in these paradigms.

\section{Conclusions and Recommendations}  
\label{sec:discussion}

In this half-position-half-survey paper, we reviewed hundreds of works on NLP model interpretability and analysis from the past decade. Unlike other surveys, we examined interpretability methods, paradigms, and properties from the stakeholders' perspective. 
Additionally, we conducted a first-of-its-kind large-scale trend analysis by exploring the usage of interpretability methods within the NLP community and in research fields outside of it. Our analysis reveals substantial diversity between research fields, particularly between NLP developers and non-developer stakeholders. To bridge these gaps and promote the adoption of NLP interpretability methods in other fields, we recommend the following steps for NLP researchers:

\begin{table}[!t]
\centering
\normalsize
\begin{adjustbox}{width=0.475\textwidth}
\begin{tabular}{l|l|cc|cc}
\toprule
\multicolumn{2}{c|}{} & \multicolumn{2}{c}{\textbf{NLP}} & \multicolumn{2}{|c}{\textbf{Outside}} \\
\multicolumn{2}{c|}{} & \textit{\underline{No LLMs}} & \textit{\underline{LLMs}} & \textit{\underline{No LLMs}} & \textit{\underline{LLMs}} \\
\midrule
\multirowcell{3}{\rotatebox[origin=c]{90}{\textbf{Year}}}  &  2022         &      97.5 & 2.5 &             100.0 &                  0.0 \\
&  2023         &     72.4 & 27.6 &                      81.8 &          18.2  \\
& 2024         &    33.3 &   66.7 &          49.3 &        50.7               \\
\midrule
\multirowcell{10}{\rotatebox[origin=c]{90}{\textbf{Paradigms} ('23 + '24)}}  &  
\cellcolor{SkyBlue!25!white} Attributions &  37.4 & \color{Red}{\textbf{19.4}}                 &       41.9 &  \color{Red}{\textbf{24.3}}             \\
&\cellcolor{RoyalBlue!35!white} LIME/SHAP &          6.3 &       3.3 &             17.5 &           \color{Red}{\textbf{4.3}} \\
&\cellcolor{LimeGreen!35!white} Probing &         11.4&      10.6 &              6.9 &           3.5 \\
&\cellcolor{Green!45!white} Clustering &          3.4 &           0.5         &             16.9 & \color{Red}{\textbf{2.6}} \\
&\cellcolor{Melon!35!white} Mechanistic &         10.6 &         \color{Green}{\textbf{15.2}}    &              2.5 &   2.6\\
&\cellcolor{Red!45!white} Diagnostic&          3.7 &        4.8       &              1.2 &  1.7  \\
&\cellcolor{Orange!45!white} Adversarial &          4.6 &       5.6          &              1.2 & 0.0  \\
&\cellcolor{Orchid!35!white} Counterfactuals&          3.4 &       4.3 &              0.0 &           1.7 \\
&\cellcolor{Fuchsia!55!white} Lang. Expl. &         10.9 &      \color{Green}{\textbf{30.8}} &              6.2 &          \color{Green}{\textbf{48.7}}\\
&\cellcolor{Goldenrod!45!white} Self-explain  &          7.1&       4.8  &              4.4 &           6.1\\
&\cellcolor{Brown!45!white} Classic  &          1.1 &       0.8&              1.2 &           4.3 \\
\bottomrule
\end{tabular}
\end{adjustbox}
\caption{Percentage of '23-'24 interpretability papers by field (\textbf{NLP} and fields \textbf{Outside} NLP) and by whether the paper employs an \textit{\underline{LLM}}. The top three rows present the distribution for each field and year (\textit{\underline{LLMs}} +\textit{\underline{No LLMs}}=100\%). The 11 bottom rows present the distribution by paradigms, each column summing to 100\%.}
\label{tab:llm_trends}
\vspace{-0.8em}
\end{table}

\medskip\noindent\textbf{Clearly define the stakeholders and applications of your work.} 
Researchers should explicitly state in the introduction who the stakeholders of their method are, the needs it addresses, its core properties, and its potential applications within and outside the NLP community. Articulating these aspects helps position the research within a broader context and ensures relevant audiences can effectively engage with the method. Additionally, demonstrating applications of interpretability methods in other fields can enhance their visibility and adoption. Publishing NLP research in interdisciplinary venues \citep{Ophir2020DeepNN, Badian2023SocialMI} fosters cross-domain collaboration and broadens the impact beyond NLP.

\medskip\noindent\textbf{Develop user-friendly code and write detailed guides for non-technical users.} 
Researchers outside the NLP community sometimes utilize specific methods due to specific needs (e.g., probing in neuroscience is used for aligning representations with brain activity). Yet, many utilize methods for the wrong reason: extensive familiarity with popular methods in non-NLP domains and with well-documented code in common data science libraries (e.g., SHAP, LIME, and Scikit-learn). 

To encourage the adoption of NLP interpretability methods beyond our community, researchers should prioritize developing user-friendly code accompanied by detailed guides for non-technical users. Additionally, the code should generate attractive and easy-to-understand visualizations. Making the methods more accessible can help integrate them into other scientific and industrial domains.

\medskip\noindent\textbf{Expand the reach of prevalent NLP interpretability paradigms.} Two paradigms have gained traction in NLP, particularly with the rise of LLMs: Natural Language Explanations and Mechanistic Interpretability. We found that natural language explanation methods are also extremely prevalent in non-NLP fields. We believe this rapid adoption is concerning, as their reliability remains a topic of ongoing debate in research. Our community should investigate the faithfulness of these methods \citep{Lanham2023, Parcalabescu2023, Bao2024, Wu2024} and determine whether they can replace traditional, extensively researched methods.

Conversely, while Mechanistic interpretability research is trending within the NLP community, explanations of internal model components are rarely used in other fields. Our community should explore whether and how mechanistic interpretability can be adapted more broadly \citep{Sharkey2025OpenPI}.

\medskip\noindent\textbf{We need more concept-level, self-explaining, and causal-based methods.}
In Appendix \S\ref{sub:concepts}, we highlight the potential of high-level concept explanations, particularly for non-expert stakeholders such as end-users, given the challenges of explaining lengthy raw textual inputs. Even though they can improve the accessibility of model insights \citep{Poursabzi21}, concept-level methods remain largely underutilized, accounting for only 2\% of the papers, as shown in Figure~\ref{fig:mechanism_trends}. 

Stakeholders using NLP models for decision-making require faithful explanations \citep{FederKMPSWEGRRS22}. In Appendix \S\ref{sub:causality}, we highlight the important role of causality in fostering faithfulness. Yet, Table~\ref{tab:properties} indicates that causal-based methods are rarely used (5.2\% in NLP and 1.9\% outside). 

Finally, building on the seminal calls of XAI researchers \citep{Rudin2018, ArrietaRSBTBGGM20}, we believe in self-explaining methods as a promising path toward the ``holy grail'' of NLP: achieving intrinsic interpretability while minimizing performance degradation. Yet, as Table~\ref{tab:llm_trends} indicates, only about 7\% of papers focus on self-explaining models, leaving them largely underexplored. 

\medskip\noindent\textbf{The LLM era presents new research opportunities.}
Despite the expectation that non-developers would benefit from concept-level, self-explaining, and causal-based methods, their adoption remains limited. We believe this is mainly due to the lack of research and development within the NLP community. This gap restricts the broader applicability of NLP models, particularly in domains where transparency and interpretability are essential. 

The increasing capabilities of LLMs provide an unprecedented opportunity to develop novel concept-level, self-explaining, and causal-based interpretability methods. Indeed, many of the works discussed in this paper demonstrate such potential (e.g., \citet{gat2023faithful} and \citet{stacey-etal-2024-atomic}). 
By expanding research in these directions, the NLP community can contribute to developing models that are more reliable, explainable, and accessible to a broader range of stakeholders.
\section{Limitations}  
\label{sec:limitations}


\medskip\noindent\textbf{Other Modalities.} The focus of our paper, while broad, centers on NLP and does not address other input modalities beyond text, such as visual or audio. These modalities, especially when considering the recent advancement of large multimodal models, could be vital for certain stakeholders, and it can be believed that the conclusions from our analysis would not be generalized to interpretability methods of vision and audio systems. 

\medskip\noindent\textbf{LLM Annotations.} Even though we manually verified the LLM annotations on a subset of 100 papers and observed high agreement rates with human annotations (over 95\%), it is possible that the LLM introduced potential biases. The statistics might have differed slightly if all 2000+ papers had been manually annotated. 
However, the manual annotation process is extremely time-consuming and requires high-domain expertise. This process involved reading full abstracts and assessing the nine annotation properties (900 annotations).
Therefore, while our findings benefit from high agreement rates between LLM and human annotations, they also emphasize the need for continuous human oversight and validation in studies that use automated tools for literature analysis \citep{Calderon2025TheAA}.


\section*{Acknowledgments}
NC is funded by the Clore Foundation’s PhD fellowship. We gratefully acknowledge the support provided by Google’s Gemma Academic Program,
which has significantly contributed to advancing our research. We would also like to thank the members of the DDS@Technion NLP group for their valuable feedback and advice.


\bibliography{custom}

\begin{thebibliography}{296}
\providecommand{\natexlab}[1]{#1}

\bibitem[{Abraham et~al.(2022)Abraham, D'Oosterlinck, Feder, Gat, Geiger, Potts, Reichart, and Wu}]{AbrahamDFGGPRW22}
Eldar~David Abraham, Karel D'Oosterlinck, Amir Feder, Yair~Ori Gat, Atticus Geiger, Christopher Potts, Roi Reichart, and Zhengxuan Wu. 2022.
\newblock \href {http://papers.nips.cc/paper\_files/paper/2022/hash/701ec28790b29a5bc33832b7bdc4c3b6-Abstract-Conference.html} {Cebab: Estimating the causal effects of real-world concepts on {NLP} model behavior}.
\newblock In \emph{Advances in Neural Information Processing Systems 35: Annual Conference on Neural Information Processing Systems 2022, NeurIPS 2022, New Orleans, LA, USA, November 28 - December 9, 2022}.

\bibitem[{Adi et~al.(2017)Adi, Kermany, Belinkov, Lavi, and Goldberg}]{AdiKBLG17}
Yossi Adi, Einat Kermany, Yonatan Belinkov, Ofer Lavi, and Yoav Goldberg. 2017.
\newblock \href {https://openreview.net/forum?id=BJh6Ztuxl} {Fine-grained analysis of sentence embeddings using auxiliary prediction tasks}.
\newblock In \emph{5th International Conference on Learning Representations, {ICLR} 2017, Toulon, France, April 24-26, 2017, Conference Track Proceedings}. OpenReview.net.

\bibitem[{Aharoni and Goldberg(2020)}]{AharoniG20}
Roee Aharoni and Yoav Goldberg. 2020.
\newblock \href {https://doi.org/10.18653/V1/2020.ACL-MAIN.692} {Unsupervised domain clusters in pretrained language models}.
\newblock In \emph{Proceedings of the 58th Annual Meeting of the Association for Computational Linguistics, {ACL} 2020, Online, July 5-10, 2020}, pages 7747--7763. Association for Computational Linguistics.

\bibitem[{Alam et~al.(2023)Alam, Dalvi, Durrani, Sajjad, Khan, and Xu}]{AlamDDSK023}
Firoj Alam, Fahim Dalvi, Nadir Durrani, Hassan Sajjad, Abdul~Rafae Khan, and Jia Xu. 2023.
\newblock \href {https://doi.org/10.1609/AAAI.V37I13.27057} {Conceptx: {A} framework for latent concept analysis}.
\newblock In \emph{Thirty-Seventh {AAAI} Conference on Artificial Intelligence, {AAAI} 2023, Thirty-Fifth Conference on Innovative Applications of Artificial Intelligence, {IAAI} 2023, Thirteenth Symposium on Educational Advances in Artificial Intelligence, {EAAI} 2023, Washington, DC, USA, February 7-14, 2023}, pages 16395--16397. {AAAI} Press.

\bibitem[{Aletras et~al.(2016)Aletras, Tsarapatsanis, Preotiuc-Pietro, and Lampos}]{Aletras2016PredictingJD}
Nikolaos Aletras, Dimitrios Tsarapatsanis, Daniel Preotiuc-Pietro, and Vasileios Lampos. 2016.
\newblock \href {https://api.semanticscholar.org/CorpusID:7630289} {Predicting judicial decisions of the european court of human rights: a natural language processing perspective}.
\newblock \emph{PeerJ Comput. Sci.}, 2:e93.

\bibitem[{Allen et~al.(2023)Allen, Gan, and Zheng}]{Allen2023InterpretableML}
Genevera~I. Allen, Luqin Gan, and Lili Zheng. 2023.
\newblock \href {https://doi.org/10.48550/ARXIV.2308.01475} {Interpretable machine learning for discovery: Statistical challenges {\textbackslash}{\&} opportunities}.
\newblock \emph{CoRR}, abs/2308.01475.

\bibitem[{Alqaraawi et~al.(2020)Alqaraawi, Schuessler, Wei{\ss}, Costanza, and Berthouze}]{AlqaraawiSWCB20}
Ahmed Alqaraawi, Martin Schuessler, Philipp Wei{\ss}, Enrico Costanza, and Nadia Berthouze. 2020.
\newblock \href {https://doi.org/10.1145/3377325.3377519} {Evaluating saliency map explanations for convolutional neural networks: a user study}.
\newblock In \emph{{IUI} '20: 25th International Conference on Intelligent User Interfaces, Cagliari, Italy, March 17-20, 2020}, pages 275--285. {ACM}.

\bibitem[{Amini et~al.(2023)Amini, Pimentel, Meister, and Cotterell}]{amini-etal-2023-naturalistic}
Afra Amini, Tiago Pimentel, Clara Meister, and Ryan Cotterell. 2023.
\newblock \href {https://doi.org/10.1162/tacl_a_00554} {Naturalistic causal probing for morpho-syntax}.
\newblock \emph{Transactions of the Association for Computational Linguistics}, 11:384--403.

\bibitem[{Andreas et~al.(2016)Andreas, Rohrbach, Darrell, and Klein}]{AndreasRDK16}
Jacob Andreas, Marcus Rohrbach, Trevor Darrell, and Dan Klein. 2016.
\newblock \href {https://doi.org/10.1109/CVPR.2016.12} {Neural module networks}.
\newblock In \emph{2016 {IEEE} Conference on Computer Vision and Pattern Recognition, {CVPR} 2016, Las Vegas, NV, USA, June 27-30, 2016}, pages 39--48. {IEEE} Computer Society.

\bibitem[{Antverg and Belinkov(2022)}]{AntvergB22}
Omer Antverg and Yonatan Belinkov. 2022.
\newblock \href {https://openreview.net/forum?id=8uz0EWPQIMu} {On the pitfalls of analyzing individual neurons in language models}.
\newblock In \emph{The Tenth International Conference on Learning Representations, {ICLR} 2022, Virtual Event, April 25-29, 2022}. OpenReview.net.

\bibitem[{Arad et~al.(2023)Arad, Orgad, and Belinkov}]{Arad2024}
Dana Arad, Hadas Orgad, and Yonatan Belinkov. 2023.
\newblock \href {https://doi.org/10.48550/ARXIV.2306.00738} {Refact: Updating text-to-image models by editing the text encoder}.
\newblock \emph{CoRR}, abs/2306.00738.

\bibitem[{Arrieta et~al.(2020)Arrieta, Rodr{\'{\i}}guez, Ser, Bennetot, Tabik, Barbado, Garc{\'{\i}}a, Gil{-}Lopez, Molina, Benjamins, Chatila, and Herrera}]{ArrietaRSBTBGGM20}
Alejandro~Barredo Arrieta, Natalia~D{\'{\i}}az Rodr{\'{\i}}guez, Javier~Del Ser, Adrien Bennetot, Siham Tabik, Alberto Barbado, Salvador Garc{\'{\i}}a, Sergio Gil{-}Lopez, Daniel Molina, Richard Benjamins, Raja Chatila, and Francisco Herrera. 2020.
\newblock \href {https://doi.org/10.1016/J.INFFUS.2019.12.012} {Explainable artificial intelligence {(XAI):} concepts, taxonomies, opportunities and challenges toward responsible {AI}}.
\newblock \emph{Inf. Fusion}, 58:82--115.

\bibitem[{Attanasio et~al.(2023)Attanasio, del Arco, Nozza, and Lauscher}]{AttanasioANL23}
Giuseppe Attanasio, Flor Miriam~Plaza del Arco, Debora Nozza, and Anne Lauscher. 2023.
\newblock \href {https://doi.org/10.18653/V1/2023.EMNLP-MAIN.243} {A tale of pronouns: Interpretability informs gender bias mitigation for fairer instruction-tuned machine translation}.
\newblock In \emph{Proceedings of the 2023 Conference on Empirical Methods in Natural Language Processing, {EMNLP} 2023, Singapore, December 6-10, 2023}, pages 3996--4014. Association for Computational Linguistics.

\bibitem[{Badian et~al.(2023)Badian, Ophir, Tikochinski, Calderon, Klomek, Fruchter, and Reichart}]{Badian2023SocialMI}
Yael Badian, Yaakov Ophir, Refael Tikochinski, Nitay Calderon, Anat~Brunstein Klomek, Eyal Fruchter, and Roi Reichart. 2023.
\newblock \href {https://api.semanticscholar.org/CorpusID:265393963} {Social media images can predict suicide risk using interpretable large language-vision models.}
\newblock \emph{The Journal of clinical psychiatry}, 85 1.

\bibitem[{Balashankar et~al.(2023)Balashankar, Wang, Qin, Packer, Thain, Chi, Chen, and Beutel}]{Balashankar00PT23}
Ananth Balashankar, Xuezhi Wang, Yao Qin, Ben Packer, Nithum Thain, Ed~H. Chi, Jilin Chen, and Alex Beutel. 2023.
\newblock \href {https://doi.org/10.18653/V1/2023.FINDINGS-EMNLP.10} {Improving classifier robustness through active generative counterfactual data augmentation}.
\newblock In \emph{Findings of the Association for Computational Linguistics: {EMNLP} 2023, Singapore, December 6-10, 2023}, pages 127--139. Association for Computational Linguistics.

\bibitem[{Balkir et~al.(2022)Balkir, Kiritchenko, Nejadgholi, and Fraser}]{Balkir2022}
Esma Balkir, Svetlana Kiritchenko, Isar Nejadgholi, and Kathleen~C. Fraser. 2022.
\newblock \href {https://doi.org/10.48550/ARXIV.2206.03945} {Challenges in applying explainability methods to improve the fairness of {NLP} models}.
\newblock \emph{CoRR}, abs/2206.03945.

\bibitem[{Bao et~al.(2024)Bao, Zhang, Yang, Wang, and Zhang}]{Bao2024}
Guangsheng Bao, Hongbo Zhang, Linyi Yang, Cunxiang Wang, and Yue Zhang. 2024.
\newblock \href {https://doi.org/10.48550/ARXIV.2402.16048} {Llms with chain-of-thought are non-causal reasoners}.
\newblock \emph{CoRR}, abs/2402.16048.

\bibitem[{Bastings et~al.(2019)Bastings, Aziz, and Titov}]{BastingsAT19}
Jasmijn Bastings, Wilker Aziz, and Ivan Titov. 2019.
\newblock \href {https://doi.org/10.18653/V1/P19-1284} {Interpretable neural predictions with differentiable binary variables}.
\newblock In \emph{Proceedings of the 57th Conference of the Association for Computational Linguistics, {ACL} 2019, Florence, Italy, July 28- August 2, 2019, Volume 1: Long Papers}, pages 2963--2977. Association for Computational Linguistics.

\bibitem[{Bau et~al.(2019)Bau, Belinkov, Sajjad, Durrani, Dalvi, and Glass}]{BauBSDDG19}
Anthony Bau, Yonatan Belinkov, Hassan Sajjad, Nadir Durrani, Fahim Dalvi, and James~R. Glass. 2019.
\newblock \href {https://openreview.net/forum?id=H1z-PsR5KX} {Identifying and controlling important neurons in neural machine translation}.
\newblock In \emph{7th International Conference on Learning Representations, {ICLR} 2019, New Orleans, LA, USA, May 6-9, 2019}. OpenReview.net.

\bibitem[{Belinkov(2022)}]{Belinkov22}
Yonatan Belinkov. 2022.
\newblock \href {https://doi.org/10.1162/COLI\_A\_00422} {Probing classifiers: Promises, shortcomings, and advances}.
\newblock \emph{Comput. Linguistics}, 48(1):207--219.

\bibitem[{Belinkov and Bisk(2018)}]{BelinkovB18}
Yonatan Belinkov and Yonatan Bisk. 2018.
\newblock \href {https://openreview.net/forum?id=BJ8vJebC-} {Synthetic and natural noise both break neural machine translation}.
\newblock In \emph{6th International Conference on Learning Representations, {ICLR} 2018, Vancouver, BC, Canada, April 30 - May 3, 2018, Conference Track Proceedings}. OpenReview.net.

\bibitem[{Belinkov and Glass(2019)}]{BelinkovG19}
Yonatan Belinkov and James~R. Glass. 2019.
\newblock \href {https://doi.org/10.1162/TACL\_A\_00254} {Analysis methods in neural language processing: {A} survey}.
\newblock \emph{Trans. Assoc. Comput. Linguistics}, 7:49--72.

\bibitem[{Belrose et~al.(2023)Belrose, Furman, Smith, Halawi, Ostrovsky, McKinney, Biderman, and Steinhardt}]{Belrose2023}
Nora Belrose, Zach Furman, Logan Smith, Danny Halawi, Igor Ostrovsky, Lev McKinney, Stella Biderman, and Jacob Steinhardt. 2023.
\newblock \href {https://doi.org/10.48550/ARXIV.2303.08112} {Eliciting latent predictions from transformers with the tuned lens}.
\newblock \emph{CoRR}, abs/2303.08112.

\bibitem[{Bereska and Gavves(2024)}]{Bereska2024}
Leonard Bereska and Efstratios Gavves. 2024.
\newblock \href {https://doi.org/10.48550/ARXIV.2404.14082} {Mechanistic interpretability for {AI} safety - {A} review}.
\newblock \emph{CoRR}, abs/2404.14082.

\bibitem[{Bhan et~al.(2024)Bhan, Vittaut, Achache, Legrand, Chesneau, Blangero, Murris, and Lesot}]{Bhan2024MitigatingTT}
Milan Bhan, Jean-Noel Vittaut, Nina Achache, Victor Legrand, Nicolas Chesneau, Annabelle Blangero, Juliette Murris, and Marie-Jeanne Lesot. 2024.
\newblock \href {https://api.semanticscholar.org/CorpusID:269790751} {Mitigating text toxicity with counterfactual generation}.

\bibitem[{Birhane et~al.(2023)Birhane, Kasirzadeh, Leslie, and Wachter}]{Birhane2023ScienceIT}
Abeba Birhane, Atoosa Kasirzadeh, David Leslie, and Sandra Wachter. 2023.
\newblock \href {https://api.semanticscholar.org/CorpusID:258361324} {Science in the age of large language models}.
\newblock \emph{Nature Reviews Physics}, 5:277--280.

\bibitem[{Blodgett et~al.(2020)Blodgett, Barocas, III, and Wallach}]{BlodgettBDW20}
Su~Lin Blodgett, Solon Barocas, Hal~Daum{\'{e}} III, and Hanna~M. Wallach. 2020.
\newblock \href {https://doi.org/10.18653/V1/2020.ACL-MAIN.485} {Language (technology) is power: {A} critical survey of "bias" in {NLP}}.
\newblock In \emph{Proceedings of the 58th Annual Meeting of the Association for Computational Linguistics, {ACL} 2020, Online, July 5-10, 2020}, pages 5454--5476. Association for Computational Linguistics.

\bibitem[{Bove et~al.(2022)Bove, Aigrain, Lesot, Tijus, and Detyniecki}]{BoveALTD22}
Clara Bove, Jonathan Aigrain, Marie{-}Jeanne Lesot, Charles Tijus, and Marcin Detyniecki. 2022.
\newblock \href {https://doi.org/10.1145/3490099.3511139} {Contextualization and exploration of local feature importance explanations to improve understanding and satisfaction of non-expert users}.
\newblock In \emph{{IUI} 2022: 27th International Conference on Intelligent User Interfaces, Helsinki, Finland, March 22 - 25, 2022}, pages 807--819. {ACM}.

\bibitem[{Bu{\c{c}}inca et~al.(2020)Bu{\c{c}}inca, Lin, Gajos, and Glassman}]{BucincaLGG20}
Zana Bu{\c{c}}inca, Phoebe Lin, Krzysztof~Z. Gajos, and Elena~L. Glassman. 2020.
\newblock \href {https://doi.org/10.1145/3377325.3377498} {Proxy tasks and subjective measures can be misleading in evaluating explainable {AI} systems}.
\newblock In \emph{{IUI} '20: 25th International Conference on Intelligent User Interfaces, Cagliari, Italy, March 17-20, 2020}, pages 454--464. {ACM}.

\bibitem[{Burchardt et~al.(2017)Burchardt, Macketanz, Dehdari, Heigold, Peter, and Williams}]{BurchardtMDHPW17}
Aljoscha Burchardt, Vivien Macketanz, Jon Dehdari, Georg Heigold, Jan{-}Thorsten Peter, and Philip Williams. 2017.
\newblock \href {http://ufal.mff.cuni.cz/pbml/108/art-burchardt-macketanz-dehdari-heigold-peter-williams.pdf} {A linguistic evaluation of rule-based, phrase-based, and neural {MT} engines}.
\newblock \emph{Prague Bull. Math. Linguistics}, 108:159--170.

\bibitem[{Burlot and Yvon(2017)}]{BurlotY17}
Franck Burlot and Fran{\c{c}}ois Yvon. 2017.
\newblock \href {https://doi.org/10.18653/V1/W17-4705} {Evaluating the morphological competence of machine translation systems}.
\newblock In \emph{Proceedings of the Second Conference on Machine Translation, {WMT} 2017, Copenhagen, Denmark, September 7-8, 2017}, pages 43--55. Association for Computational Linguistics.

\bibitem[{Calderon et~al.(2022)Calderon, Ben{-}David, Feder, and Reichart}]{CalderonBFR22}
Nitay Calderon, Eyal Ben{-}David, Amir Feder, and Roi Reichart. 2022.
\newblock \href {https://doi.org/10.18653/V1/2022.ACL-LONG.533} {Docogen: Domain counterfactual generation for low resource domain adaptation}.
\newblock In \emph{Proceedings of the 60th Annual Meeting of the Association for Computational Linguistics (Volume 1: Long Papers), {ACL} 2022, Dublin, Ireland, May 22-27, 2022}, pages 7727--7746. Association for Computational Linguistics.

\bibitem[{Calderon et~al.(2023)Calderon, Mukherjee, Reichart, and Kantor}]{CalderonMRK23}
Nitay Calderon, Subhabrata Mukherjee, Roi Reichart, and Amir Kantor. 2023.
\newblock \href {https://doi.org/10.18653/V1/2023.ACL-LONG.818} {A systematic study of knowledge distillation for natural language generation with pseudo-target training}.
\newblock In \emph{Proceedings of the 61st Annual Meeting of the Association for Computational Linguistics (Volume 1: Long Papers), {ACL} 2023, Toronto, Canada, July 9-14, 2023}, pages 14632--14659. Association for Computational Linguistics.

\bibitem[{Calderon et~al.(2024)Calderon, Porat, Ben-David, Chapanin, Gekhman, Oved, Shalumov, and Reichart}]{calderon2023measuring}
Nitay Calderon, Naveh Porat, Eyal Ben-David, Alexander Chapanin, Zorik Gekhman, Nadav Oved, Vitaly Shalumov, and Roi Reichart. 2024.
\newblock \href {https://arxiv.org/abs/2306.00168} {Measuring the robustness of nlp models to domain shifts}.
\newblock \emph{arXiv preprint arXiv:2306.00168}.

\bibitem[{Calderon et~al.(2025)Calderon, Reichart, and Dror}]{Calderon2025TheAA}
Nitay Calderon, Roi Reichart, and Rotem Dror. 2025.
\newblock \href {https://arxiv.org/abs/2501.10970} {The alternative annotator test for llm-as-a-judge: How to statistically justify replacing human annotators with llms}.

\bibitem[{Camburu et~al.(2018)Camburu, Rockt{\"{a}}schel, Lukasiewicz, and Blunsom}]{CamburuRLB18}
Oana{-}Maria Camburu, Tim Rockt{\"{a}}schel, Thomas Lukasiewicz, and Phil Blunsom. 2018.
\newblock \href {https://proceedings.neurips.cc/paper/2018/hash/4c7a167bb329bd92580a99ce422d6fa6-Abstract.html} {e-snli: Natural language inference with natural language explanations}.
\newblock In \emph{Advances in Neural Information Processing Systems 31: Annual Conference on Neural Information Processing Systems 2018, NeurIPS 2018, December 3-8, 2018, Montr{\'{e}}al, Canada}, pages 9560--9572.

\bibitem[{Camburu et~al.(2020)Camburu, Shillingford, Minervini, Lukasiewicz, and Blunsom}]{CamburuSMLB20}
Oana{-}Maria Camburu, Brendan Shillingford, Pasquale Minervini, Thomas Lukasiewicz, and Phil Blunsom. 2020.
\newblock \href {https://doi.org/10.18653/V1/2020.ACL-MAIN.382} {Make up your mind! adversarial generation of inconsistent natural language explanations}.
\newblock In \emph{Proceedings of the 58th Annual Meeting of the Association for Computational Linguistics, {ACL} 2020, Online, July 5-10, 2020}, pages 4157--4165. Association for Computational Linguistics.

\bibitem[{Chefer et~al.(2021)Chefer, Gur, and Wolf}]{CheferGW21}
Hila Chefer, Shir Gur, and Lior Wolf. 2021.
\newblock \href {https://doi.org/10.1109/CVPR46437.2021.00084} {Transformer interpretability beyond attention visualization}.
\newblock In \emph{{IEEE} Conference on Computer Vision and Pattern Recognition, {CVPR} 2021, virtual, June 19-25, 2021}, pages 782--791. Computer Vision Foundation / {IEEE}.

\bibitem[{Chemmengath et~al.(2022)Chemmengath, Azad, Luss, and Dhurandhar}]{ChemmengathALD22}
Saneem~A. Chemmengath, Amar~Prakash Azad, Ronny Luss, and Amit Dhurandhar. 2022.
\newblock \href {https://doi.org/10.18653/V1/2022.EMNLP-MAIN.484} {Let the {CAT} out of the bag: Contrastive attributed explanations for text}.
\newblock In \emph{Proceedings of the 2022 Conference on Empirical Methods in Natural Language Processing, {EMNLP} 2022, Abu Dhabi, United Arab Emirates, December 7-11, 2022}, pages 7190--7206. Association for Computational Linguistics.

\bibitem[{Chen et~al.(2021)Chen, Tam, Raffel, Bansal, and Yang}]{Chen2021AnES}
Jiaao Chen, Derek Tam, Colin Raffel, Mohit Bansal, and Diyi Yang. 2021.
\newblock \href {https://api.semanticscholar.org/CorpusID:235422524} {An empirical survey of data augmentation for limited data learning in nlp}.
\newblock \emph{Transactions of the Association for Computational Linguistics}, 11:191--211.

\bibitem[{Chiu et~al.(2024)Chiu, Jiang, Antoniak, Park, Li, Bhatia, Ravi, Tsvetkov, Shwartz, and Choi}]{Chiu2024}
Yu~Ying Chiu, Liwei Jiang, Maria Antoniak, Chan~Young Park, Shuyue~Stella Li, Mehar Bhatia, Sahithya Ravi, Yulia Tsvetkov, Vered Shwartz, and Yejin Choi. 2024.
\newblock \href {https://doi.org/10.48550/ARXIV.2404.06664} {Culturalteaming: Ai-assisted interactive red-teaming for challenging llms' (lack of) multicultural knowledge}.
\newblock \emph{CoRR}, abs/2404.06664.

\bibitem[{Choudhury and Shamszare(2023)}]{Choudhury2023InvestigatingTI}
Avishek Choudhury and Hamid Shamszare. 2023.
\newblock \href {https://api.semanticscholar.org/CorpusID:258922988} {Investigating the impact of user trust on the adoption and use of chatgpt: Survey analysis}.
\newblock \emph{Journal of Medical Internet Research}, 25.

\bibitem[{Chu et~al.(2023)Chu, Chen, Chen, Yu, He, Wang, Peng, Liu, Qin, and Liu}]{Chu2023}
Zheng Chu, Jingchang Chen, Qianglong Chen, Weijiang Yu, Tao He, Haotian Wang, Weihua Peng, Ming Liu, Bing Qin, and Ting Liu. 2023.
\newblock \href {https://doi.org/10.48550/ARXIV.2309.15402} {A survey of chain of thought reasoning: Advances, frontiers and future}.
\newblock \emph{CoRR}, abs/2309.15402.

\bibitem[{Clusmann et~al.(2023)Clusmann, Kolbinger, Muti, Carrero, Eckardt, Laleh, L{\"o}ffler, Schwarzkopf, Unger, Veldhuizen et~al.}]{clusmann2023future}
Jan Clusmann, Fiona~R Kolbinger, Hannah~Sophie Muti, Zunamys~I Carrero, Jan-Niklas Eckardt, Narmin~Ghaffari Laleh, Chiara Maria~Lavinia L{\"o}ffler, Sophie-Caroline Schwarzkopf, Michaela Unger, Gregory~P Veldhuizen, et~al. 2023.
\newblock \href {https://www.nature.com/articles/s43856-023-00370-1} {The future landscape of large language models in medicine}.
\newblock \emph{Communications medicine}, 3(1):141.

\bibitem[{Conmy et~al.(2023)Conmy, Mavor{-}Parker, Lynch, Heimersheim, and Garriga{-}Alonso}]{ConmyMLHG23}
Arthur Conmy, Augustine~N. Mavor{-}Parker, Aengus Lynch, Stefan Heimersheim, and Adri{\`{a}} Garriga{-}Alonso. 2023.
\newblock \href {http://papers.nips.cc/paper\_files/paper/2023/hash/34e1dbe95d34d7ebaf99b9bcaeb5b2be-Abstract-Conference.html} {Towards automated circuit discovery for mechanistic interpretability}.
\newblock In \emph{Advances in Neural Information Processing Systems 36: Annual Conference on Neural Information Processing Systems 2023, NeurIPS 2023, New Orleans, LA, USA, December 10 - 16, 2023}.

\bibitem[{Conneau et~al.(2018)Conneau, Kruszewski, Lample, Barrault, and Baroni}]{BaroniBLKC18}
Alexis Conneau, Germ{\'{a}}n Kruszewski, Guillaume Lample, Lo{\"{\i}}c Barrault, and Marco Baroni. 2018.
\newblock \href {https://doi.org/10.18653/V1/P18-1198} {What you can cram into a single {\textbackslash}{\textdollar}{\&}!{\#}* vector: Probing sentence embeddings for linguistic properties}.
\newblock In \emph{Proceedings of the 56th Annual Meeting of the Association for Computational Linguistics, {ACL} 2018, Melbourne, Australia, July 15-20, 2018, Volume 1: Long Papers}, pages 2126--2136. Association for Computational Linguistics.

\bibitem[{Csord{\'{a}}s et~al.(2021)Csord{\'{a}}s, van Steenkiste, and Schmidhuber}]{CsordasSS21}
R{\'{o}}bert Csord{\'{a}}s, Sjoerd van Steenkiste, and J{\"{u}}rgen Schmidhuber. 2021.
\newblock \href {https://openreview.net/forum?id=7uVcpu-gMD} {Are neural nets modular? inspecting functional modularity through differentiable weight masks}.
\newblock In \emph{9th International Conference on Learning Representations, {ICLR} 2021, Virtual Event, Austria, May 3-7, 2021}. OpenReview.net.

\bibitem[{Cunningham et~al.(2023)Cunningham, Ewart, Riggs, Huben, and Sharkey}]{Cunningham2023}
Hoagy Cunningham, Aidan Ewart, Logan Riggs, Robert Huben, and Lee Sharkey. 2023.
\newblock \href {https://doi.org/10.48550/ARXIV.2309.08600} {Sparse autoencoders find highly interpretable features in language models}.
\newblock \emph{CoRR}, abs/2309.08600.

\bibitem[{Dai et~al.(2022)Dai, Dong, Hao, Sui, Chang, and Wei}]{DaiDHSCW22}
Damai Dai, Li~Dong, Yaru Hao, Zhifang Sui, Baobao Chang, and Furu Wei. 2022.
\newblock \href {https://doi.org/10.18653/V1/2022.ACL-LONG.581} {Knowledge neurons in pretrained transformers}.
\newblock In \emph{Proceedings of the 60th Annual Meeting of the Association for Computational Linguistics (Volume 1: Long Papers), {ACL} 2022, Dublin, Ireland, May 22-27, 2022}, pages 8493--8502. Association for Computational Linguistics.

\bibitem[{Danilevsky et~al.(2020)Danilevsky, Qian, Aharonov, Katsis, Kawas, and Sen}]{Danilevsky2020}
Marina Danilevsky, Kun Qian, Ranit Aharonov, Yannis Katsis, Ban Kawas, and Prithviraj Sen. 2020.
\newblock \href {https://aclanthology.org/2020.aacl-main.46/} {A survey of the state of explainable {AI} for natural language processing}.
\newblock In \emph{Proceedings of the 1st Conference of the Asia-Pacific Chapter of the Association for Computational Linguistics and the 10th International Joint Conference on Natural Language Processing, {AACL/IJCNLP} 2020, Suzhou, China, December 4-7, 2020}, pages 447--459. Association for Computational Linguistics.

\bibitem[{Dar et~al.(2023)Dar, Geva, Gupta, and Berant}]{DarG0B23}
Guy Dar, Mor Geva, Ankit Gupta, and Jonathan Berant. 2023.
\newblock \href {https://doi.org/10.18653/V1/2023.ACL-LONG.893} {Analyzing transformers in embedding space}.
\newblock In \emph{Proceedings of the 61st Annual Meeting of the Association for Computational Linguistics (Volume 1: Long Papers), {ACL} 2023, Toronto, Canada, July 9-14, 2023}, pages 16124--16170. Association for Computational Linguistics.

\bibitem[{Das et~al.(2022)Das, Gupta, Kovatchev, Lease, and Li}]{Das22}
Anubrata Das, Chitrank Gupta, Venelin Kovatchev, Matthew Lease, and Junyi~Jessy Li. 2022.
\newblock \href {https://doi.org/10.18653/V1/2022.ACL-LONG.213} {Prototex: Explaining model decisions with prototype tensors}.
\newblock In \emph{Proceedings of the 60th Annual Meeting of the Association for Computational Linguistics (Volume 1: Long Papers), {ACL} 2022, Dublin, Ireland, May 22-27, 2022}, pages 2986--2997. Association for Computational Linguistics.

\bibitem[{De{-}Arteaga et~al.(2019)De{-}Arteaga, Romanov, Wallach, Chayes, Borgs, Chouldechova, Geyik, Kenthapadi, and Kalai}]{De-ArteagaRWCBC19}
Maria De{-}Arteaga, Alexey Romanov, Hanna~M. Wallach, Jennifer~T. Chayes, Christian Borgs, Alexandra Chouldechova, Sahin~Cem Geyik, Krishnaram Kenthapadi, and Adam~Tauman Kalai. 2019.
\newblock \href {https://doi.org/10.1145/3287560.3287572} {Bias in bios: {A} case study of semantic representation bias in a high-stakes setting}.
\newblock In \emph{Proceedings of the Conference on Fairness, Accountability, and Transparency, FAT* 2019, Atlanta, GA, USA, January 29-31, 2019}, pages 120--128. {ACM}.

\bibitem[{Dev et~al.(2021)Dev, Monajatipoor, Ovalle, Subramonian, Phillips, and Chang}]{DevMOSPC21}
Sunipa Dev, Masoud Monajatipoor, Anaelia Ovalle, Arjun Subramonian, Jeff~M. Phillips, and Kai{-}Wei Chang. 2021.
\newblock \href {https://doi.org/10.18653/V1/2021.EMNLP-MAIN.150} {Harms of gender exclusivity and challenges in non-binary representation in language technologies}.
\newblock In \emph{Proceedings of the 2021 Conference on Empirical Methods in Natural Language Processing, {EMNLP} 2021, Virtual Event / Punta Cana, Dominican Republic, 7-11 November, 2021}, pages 1968--1994. Association for Computational Linguistics.

\bibitem[{Dhamala et~al.(2021)Dhamala, Sun, Kumar, Krishna, Pruksachatkun, Chang, and Gupta}]{DhamalaSKKPCG21}
Jwala Dhamala, Tony Sun, Varun Kumar, Satyapriya Krishna, Yada Pruksachatkun, Kai{-}Wei Chang, and Rahul Gupta. 2021.
\newblock \href {https://doi.org/10.1145/3442188.3445924} {{BOLD:} dataset and metrics for measuring biases in open-ended language generation}.
\newblock In \emph{FAccT '21: 2021 {ACM} Conference on Fairness, Accountability, and Transparency, Virtual Event / Toronto, Canada, March 3-10, 2021}, pages 862--872. {ACM}.

\bibitem[{Dixit et~al.(2022)Dixit, Paranjape, Hajishirzi, and Zettlemoyer}]{DixitPHZ22}
Tanay Dixit, Bhargavi Paranjape, Hannaneh Hajishirzi, and Luke Zettlemoyer. 2022.
\newblock \href {https://doi.org/10.18653/V1/2022.FINDINGS-EMNLP.216} {{CORE:} {A} retrieve-then-edit framework for counterfactual data generation}.
\newblock In \emph{Findings of the Association for Computational Linguistics: {EMNLP} 2022, Abu Dhabi, United Arab Emirates, December 7-11, 2022}, pages 2964--2984. Association for Computational Linguistics.

\bibitem[{Doshi-Velez and Kim(2017)}]{doshi2017towards}
Finale Doshi-Velez and Been Kim. 2017.
\newblock \href {https://arxiv.org/abs/1702.08608} {Towards a rigorous science of interpretable machine learning}.
\newblock \emph{arXiv preprint arXiv:1702.08608}.

\bibitem[{Doughman and Khreich(2022)}]{Doughman2022}
Jad Doughman and Wael Khreich. 2022.
\newblock \href {https://arxiv.org/abs/2201.08675} {Gender bias in text: Labeled datasets and lexicons}.
\newblock \emph{CoRR}, abs/2201.08675.

\bibitem[{Durrani et~al.(2020)Durrani, Sajjad, Dalvi, and Belinkov}]{DurraniSDB20}
Nadir Durrani, Hassan Sajjad, Fahim Dalvi, and Yonatan Belinkov. 2020.
\newblock \href {https://doi.org/10.18653/V1/2020.EMNLP-MAIN.395} {Analyzing individual neurons in pre-trained language models}.
\newblock In \emph{Proceedings of the 2020 Conference on Empirical Methods in Natural Language Processing, {EMNLP} 2020, Online, November 16-20, 2020}, pages 4865--4880. Association for Computational Linguistics.

\bibitem[{Ebrahimi et~al.(2018)Ebrahimi, Rao, Lowd, and Dou}]{EbrahimiRLD18}
Javid Ebrahimi, Anyi Rao, Daniel Lowd, and Dejing Dou. 2018.
\newblock \href {https://doi.org/10.18653/V1/P18-2006} {Hotflip: White-box adversarial examples for text classification}.
\newblock In \emph{Proceedings of the 56th Annual Meeting of the Association for Computational Linguistics, {ACL} 2018, Melbourne, Australia, July 15-20, 2018, Volume 2: Short Papers}, pages 31--36. Association for Computational Linguistics.

\bibitem[{Editorials(2023)}]{2023ToolsSA}
Nature Editorials. 2023.
\newblock \href {https://api.semanticscholar.org/CorpusID:256229858} {Tools such as chatgpt threaten transparent science; here are our ground rules for their use}.
\newblock \emph{Nature}, 613:612.

\bibitem[{El-Haj et~al.(2019)El-Haj, Rayson, Walker, Young, and Simaki}]{el2019search}
Mahmoud El-Haj, Paul Rayson, Martin Walker, Steven Young, and Vasiliki Simaki. 2019.
\newblock \href {https://onlinelibrary.wiley.com/doi/abs/10.1111/jbfa.12378} {In search of meaning: Lessons, resources and next steps for computational analysis of financial discourse}.
\newblock \emph{Journal of Business Finance \& Accounting}, 46(3-4):265--306.

\bibitem[{Elazar et~al.(2022)Elazar, Kassner, Ravfogel, Feder, Ravichander, Mosbach, Belinkov, Sch{\"{u}}tze, and Goldberg}]{Elazar2022}
Yanai Elazar, Nora Kassner, Shauli Ravfogel, Amir Feder, Abhilasha Ravichander, Marius Mosbach, Yonatan Belinkov, Hinrich Sch{\"{u}}tze, and Yoav Goldberg. 2022.
\newblock \href {https://doi.org/10.48550/ARXIV.2207.14251} {Measuring causal effects of data statistics on language model's 'factual' predictions}.
\newblock \emph{CoRR}, abs/2207.14251.

\bibitem[{Elazar et~al.(2021)Elazar, Ravfogel, Jacovi, and Goldberg}]{ElazarRJG21}
Yanai Elazar, Shauli Ravfogel, Alon Jacovi, and Yoav Goldberg. 2021.
\newblock \href {https://doi.org/10.1162/TACL\_A\_00359} {Amnesic probing: Behavioral explanation with amnesic counterfactuals}.
\newblock \emph{Trans. Assoc. Comput. Linguistics}, 9:160--175.

\bibitem[{Elazar et~al.(2024)Elazar, Zhang, Wadden, Zhang, and Smith}]{Elazar0WZS24}
Yanai Elazar, Jiayao Zhang, David Wadden, Bo~Zhang, and Noah~A. Smith. 2024.
\newblock \href {https://proceedings.mlr.press/v236/elazar24a.html} {Estimating the causal effect of early arxiving on paper acceptance}.
\newblock In \emph{Causal Learning and Reasoning, 1-3 April 2024, Los Angeles, California, {USA}}, volume 236 of \emph{Proceedings of Machine Learning Research}, pages 913--933. {PMLR}.

\bibitem[{Eloundou et~al.(2023)Eloundou, Manning, Mishkin, and Rock}]{Eloundou2023}
Tyna Eloundou, Sam Manning, Pamela Mishkin, and Daniel Rock. 2023.
\newblock \href {https://doi.org/10.48550/ARXIV.2303.10130} {Gpts are gpts: An early look at the labor market impact potential of large language models}.
\newblock \emph{CoRR}, abs/2303.10130.

\bibitem[{Enguehard(2023)}]{Enguehard23}
Joseph Enguehard. 2023.
\newblock \href {https://doi.org/10.18653/V1/2023.FINDINGS-ACL.477} {Sequential integrated gradients: a simple but effective method for explaining language models}.
\newblock In \emph{Findings of the Association for Computational Linguistics: {ACL} 2023, Toronto, Canada, July 9-14, 2023}, pages 7555--7565. Association for Computational Linguistics.

\bibitem[{Fang et~al.(2023)Fang, Cohn, Baldwin, and Frermann}]{FangCBF23}
Biaoyan Fang, Trevor Cohn, Timothy Baldwin, and Lea Frermann. 2023.
\newblock \href {https://doi.org/10.18653/V1/2023.IJCNLP-SHORT.22} {It's not only what you say, it's also who it's said to: Counterfactual analysis of interactive behavior in the courtroom}.
\newblock In \emph{Proceedings of the 13th International Joint Conference on Natural Language Processing and the 3rd Conference of the Asia-Pacific Chapter of the Association for Computational Linguistics, {IJCNLP} 2023 - Volume 2: Short Papers, Nusa Dua, Bali, November 1-4, 2023}, pages 197--207. Association for Computational Linguistics.

\bibitem[{Feder et~al.(2022)Feder, Keith, Manzoor, Pryzant, Sridhar, Wood{-}Doughty, Eisenstein, Grimmer, Reichart, Roberts, Stewart, Veitch, and Yang}]{FederKMPSWEGRRS22}
Amir Feder, Katherine~A. Keith, Emaad Manzoor, Reid Pryzant, Dhanya Sridhar, Zach Wood{-}Doughty, Jacob Eisenstein, Justin Grimmer, Roi Reichart, Margaret~E. Roberts, Brandon~M. Stewart, Victor Veitch, and Diyi Yang. 2022.
\newblock \href {https://transacl.org/ojs/index.php/tacl/article/view/4005} {Causal inference in natural language processing: Estimation, prediction, interpretation and beyond}.
\newblock \emph{Trans. Assoc. Comput. Linguistics}, 10:1138--1158.

\bibitem[{Feder et~al.(2021)Feder, Oved, Shalit, and Reichart}]{FederOSR21}
Amir Feder, Nadav Oved, Uri Shalit, and Roi Reichart. 2021.
\newblock \href {https://doi.org/10.1162/COLI\_A\_00404} {Causalm: Causal model explanation through counterfactual language models}.
\newblock \emph{Comput. Linguistics}, 47(2):333--386.

\bibitem[{Feder et~al.(2023)Feder, Wald, Shi, Saria, and Blei}]{FederWSSB23}
Amir Feder, Yoav Wald, Claudia Shi, Suchi Saria, and David~M. Blei. 2023.
\newblock \href {http://papers.nips.cc/paper\_files/paper/2023/hash/df88b275bef31ac96c85f0c4013734fc-Abstract-Conference.html} {Causal-structure driven augmentations for text {OOD} generalization}.
\newblock In \emph{Advances in Neural Information Processing Systems 36: Annual Conference on Neural Information Processing Systems 2023, NeurIPS 2023, New Orleans, LA, USA, December 10 - 16, 2023}.

\bibitem[{Filandrianos et~al.(2023)Filandrianos, Dervakos, Menis{-}Mastromichalakis, Zerva, and Stamou}]{FilandrianosDMZ23}
George Filandrianos, Edmund Dervakos, Orfeas Menis{-}Mastromichalakis, Chrysoula Zerva, and Giorgos Stamou. 2023.
\newblock \href {https://doi.org/10.18653/V1/2023.FINDINGS-ACL.606} {Counterfactuals of counterfactuals: a back-translation-inspired approach to analyse counterfactual editors}.
\newblock In \emph{Findings of the Association for Computational Linguistics: {ACL} 2023, Toronto, Canada, July 9-14, 2023}, pages 9507--9525. Association for Computational Linguistics.

\bibitem[{Finlayson et~al.(2021)Finlayson, Mueller, Gehrmann, Shieber, Linzen, and Belinkov}]{FinlaysonMGSLB20}
Matthew Finlayson, Aaron Mueller, Sebastian Gehrmann, Stuart~M. Shieber, Tal Linzen, and Yonatan Belinkov. 2021.
\newblock \href {https://doi.org/10.18653/V1/2021.ACL-LONG.144} {Causal analysis of syntactic agreement mechanisms in neural language models}.
\newblock In \emph{Proceedings of the 59th Annual Meeting of the Association for Computational Linguistics and the 11th International Joint Conference on Natural Language Processing, {ACL/IJCNLP} 2021, (Volume 1: Long Papers), Virtual Event, August 1-6, 2021}, pages 1828--1843. Association for Computational Linguistics.

\bibitem[{Gade et~al.(2020)Gade, Geyik, Kenthapadi, Mithal, and Taly}]{GadeGKMT20}
Krishna Gade, Sahin~Cem Geyik, Krishnaram Kenthapadi, Varun Mithal, and Ankur Taly. 2020.
\newblock \href {https://doi.org/10.1145/3351095.3375664} {Explainable {AI} in industry: practical challenges and lessons learned: implications tutorial}.
\newblock In \emph{FAT* '20: Conference on Fairness, Accountability, and Transparency, Barcelona, Spain, January 27-30, 2020}, page 699. {ACM}.

\bibitem[{Garde et~al.(2023)Garde, Kran, and Barez}]{Garde2023}
Albert Garde, Esben Kran, and Fazl Barez. 2023.
\newblock \href {https://doi.org/10.48550/ARXIV.2310.01870} {Deepdecipher: Accessing and investigating neuron activation in large language models}.
\newblock \emph{CoRR}, abs/2310.01870.

\bibitem[{Gardner et~al.(2020)Gardner, Artzi, Basmova, Berant, Bogin, and et~al.}]{Gardner2020}
Matt Gardner, Yoav Artzi, Victoria Basmova, Jonathan Berant, Ben Bogin, and et~al. 2020.
\newblock \href {https://doi.org/10.18653/V1/2020.FINDINGS-EMNLP.117} {Evaluating models' local decision boundaries via contrast sets}.
\newblock In \emph{Findings of the Association for Computational Linguistics: {EMNLP} 2020, Online Event, 16-20 November 2020}, volume {EMNLP} 2020 of \emph{Findings of {ACL}}, pages 1307--1323. Association for Computational Linguistics.

\bibitem[{Gat et~al.(2022)Gat, Calderon, Reichart, and Hazan}]{GatCRH22}
Itai Gat, Nitay Calderon, Roi Reichart, and Tamir Hazan. 2022.
\newblock \href {https://proceedings.mlr.press/v162/gat22a.html} {A functional information perspective on model interpretation}.
\newblock In \emph{International Conference on Machine Learning, {ICML} 2022, 17-23 July 2022, Baltimore, Maryland, {USA}}, volume 162 of \emph{Proceedings of Machine Learning Research}, pages 7266--7278. {PMLR}.

\bibitem[{Gat et~al.(2023)Gat, Calderon, Feder, Chapanin, Sharma, and Reichart}]{gat2023faithful}
Yair~Ori Gat, Nitay Calderon, Amir Feder, Alexander Chapanin, Amit Sharma, and Roi Reichart. 2023.
\newblock \href {https://doi.org/10.48550/arXiv.2310.00603} {Faithful explanations of black-box nlp models using llm-generated counterfactuals}.
\newblock In \emph{The Twelfth International Conference on Learning Representations}.

\bibitem[{Geiger et~al.(2021)Geiger, Lu, Icard, and Potts}]{GeigerLIP21}
Atticus Geiger, Hanson Lu, Thomas Icard, and Christopher Potts. 2021.
\newblock \href {https://proceedings.neurips.cc/paper/2021/hash/4f5c422f4d49a5a807eda27434231040-Abstract.html} {Causal abstractions of neural networks}.
\newblock In \emph{Advances in Neural Information Processing Systems 34: Annual Conference on Neural Information Processing Systems 2021, NeurIPS 2021, December 6-14, 2021, virtual}, pages 9574--9586.

\bibitem[{Geiger et~al.(2022)Geiger, Wu, Lu, Rozner, Kreiss, Icard, Goodman, and Potts}]{GeigerWLRKIGP22}
Atticus Geiger, Zhengxuan Wu, Hanson Lu, Josh Rozner, Elisa Kreiss, Thomas Icard, Noah~D. Goodman, and Christopher Potts. 2022.
\newblock \href {https://proceedings.mlr.press/v162/geiger22a.html} {Inducing causal structure for interpretable neural networks}.
\newblock In \emph{International Conference on Machine Learning, {ICML} 2022, 17-23 July 2022, Baltimore, Maryland, {USA}}, volume 162 of \emph{Proceedings of Machine Learning Research}, pages 7324--7338. {PMLR}.

\bibitem[{Gekhman et~al.(2024)Gekhman, Yona, Aharoni, Eyal, Feder, Reichart, and Herzig}]{Gekhman24}
Zorik Gekhman, Gal Yona, Roee Aharoni, Matan Eyal, Amir Feder, Roi Reichart, and Jonathan Herzig. 2024.
\newblock \href {https://doi.org/10.48550/ARXIV.2405.05904} {Does fine-tuning llms on new knowledge encourage hallucinations?}
\newblock \emph{CoRR}, abs/2405.05904.

\bibitem[{Gennaro and Ash(2022)}]{gennaro2022emotion}
Gloria Gennaro and Elliott Ash. 2022.
\newblock \href {https://academic.oup.com/ej/article-abstract/132/643/1037/6490125} {Emotion and reason in political language}.
\newblock \emph{The Economic Journal}, 132(643):1037--1059.

\bibitem[{Geva et~al.(2022)Geva, Caciularu, Wang, and Goldberg}]{GevaCWG22}
Mor Geva, Avi Caciularu, Kevin~Ro Wang, and Yoav Goldberg. 2022.
\newblock \href {https://doi.org/10.18653/V1/2022.EMNLP-MAIN.3} {Transformer feed-forward layers build predictions by promoting concepts in the vocabulary space}.
\newblock In \emph{Proceedings of the 2022 Conference on Empirical Methods in Natural Language Processing, {EMNLP} 2022, Abu Dhabi, United Arab Emirates, December 7-11, 2022}, pages 30--45. Association for Computational Linguistics.

\bibitem[{Ghaeini et~al.(2019)Ghaeini, Fern, Shahbazi, and Tadepalli}]{GhaeiniFST19}
Reza Ghaeini, Xiaoli~Z. Fern, Hamed Shahbazi, and Prasad Tadepalli. 2019.
\newblock \href {https://doi.org/10.18653/V1/N19-1404} {Saliency learning: Teaching the model where to pay attention}.
\newblock In \emph{Proceedings of the 2019 Conference of the North American Chapter of the Association for Computational Linguistics: Human Language Technologies, {NAACL-HLT} 2019, Minneapolis, MN, USA, June 2-7, 2019, Volume 1 (Long and Short Papers)}, pages 4016--4025. Association for Computational Linguistics.

\bibitem[{Ghandeharioun et~al.(2024)Ghandeharioun, Caciularu, Pearce, Dixon, and Geva}]{Ghandeharioun2024}
Asma Ghandeharioun, Avi Caciularu, Adam Pearce, Lucas Dixon, and Mor Geva. 2024.
\newblock \href {https://doi.org/10.48550/ARXIV.2401.06102} {Patchscopes: {A} unifying framework for inspecting hidden representations of language models}.
\newblock \emph{CoRR}, abs/2401.06102.

\bibitem[{Ghorbani and Zou(2020)}]{GhorbaniZ20}
Amirata Ghorbani and James~Y. Zou. 2020.
\newblock \href {https://proceedings.neurips.cc/paper/2020/hash/41c542dfe6e4fc3deb251d64cf6ed2e4-Abstract.html} {Neuron shapley: Discovering the responsible neurons}.
\newblock In \emph{Advances in Neural Information Processing Systems 33: Annual Conference on Neural Information Processing Systems 2020, NeurIPS 2020, December 6-12, 2020, virtual}.

\bibitem[{Giulianelli et~al.(2018)Giulianelli, Harding, Mohnert, Hupkes, and Zuidema}]{GiulianelliHMHZ18}
Mario Giulianelli, Jack Harding, Florian Mohnert, Dieuwke Hupkes, and Willem~H. Zuidema. 2018.
\newblock \href {https://doi.org/10.18653/V1/W18-5426} {Under the hood: Using diagnostic classifiers to investigate and improve how language models track agreement information}.
\newblock In \emph{Proceedings of the Workshop: Analyzing and Interpreting Neural Networks for NLP, BlackboxNLP@EMNLP 2018, Brussels, Belgium, November 1, 2018}, pages 240--248. Association for Computational Linguistics.

\bibitem[{Goel et~al.(2021)Goel, Rajani, Vig, Taschdjian, Bansal, and R{\'{e}}}]{GoelRVTBR21}
Karan Goel, Nazneen~Fatema Rajani, Jesse Vig, Zachary Taschdjian, Mohit Bansal, and Christopher R{\'{e}}. 2021.
\newblock \href {https://doi.org/10.18653/V1/2021.NAACL-DEMOS.6} {Robustness gym: Unifying the {NLP} evaluation landscape}.
\newblock In \emph{Proceedings of the 2021 Conference of the North American Chapter of the Association for Computational Linguistics: Human Language Technologies: Demonstrations, {NAACL-HLT} 2021, Online, June 6-11, 2021}, pages 42--55. Association for Computational Linguistics.

\bibitem[{Goldstein et~al.(2022)Goldstein, Zada, Buchnik, Schain, Price, Aubrey, Nastase, Feder, Emanuel, Cohen et~al.}]{goldstein2022shared}
Ariel Goldstein, Zaid Zada, Eliav Buchnik, Mariano Schain, Amy Price, Bobbi Aubrey, Samuel~A Nastase, Amir Feder, Dotan Emanuel, Alon Cohen, et~al. 2022.
\newblock \href {https://www.nature.com/articles/s41593-022-01026-4} {Shared computational principles for language processing in humans and deep language models}.
\newblock \emph{Nature neuroscience}, 25(3):369--380.

\bibitem[{Goodman and Flaxman(2017)}]{right_to_explanation}
Bryce Goodman and Seth~R. Flaxman. 2017.
\newblock \href {https://doi.org/10.1609/AIMAG.V38I3.2741} {European union regulations on algorithmic decision-making and a "right to explanation"}.
\newblock \emph{{AI} Mag.}, 38(3):50--57.

\bibitem[{Goyal et~al.(2023)Goyal, Doddapaneni, Khapra, and Ravindran}]{GoyalDKR23}
Shreya Goyal, Sumanth Doddapaneni, Mitesh~M. Khapra, and Balaraman Ravindran. 2023.
\newblock \href {https://doi.org/10.1145/3593042} {A survey of adversarial defenses and robustness in {NLP}}.
\newblock \emph{{ACM} Comput. Surv.}, 55(14s):332:1--332:39.

\bibitem[{Gulordava et~al.(2018)Gulordava, Bojanowski, Grave, Linzen, and Baroni}]{GulordavaBGLB18}
Kristina Gulordava, Piotr Bojanowski, Edouard Grave, Tal Linzen, and Marco Baroni. 2018.
\newblock \href {https://doi.org/10.18653/V1/N18-1108} {Colorless green recurrent networks dream hierarchically}.
\newblock In \emph{Proceedings of the 2018 Conference of the North American Chapter of the Association for Computational Linguistics: Human Language Technologies, {NAACL-HLT} 2018, New Orleans, Louisiana, USA, June 1-6, 2018, Volume 1 (Long Papers)}, pages 1195--1205. Association for Computational Linguistics.

\bibitem[{Guo et~al.(2021)Guo, Sablayrolles, J{\'{e}}gou, and Kiela}]{GuoSJK21}
Chuan Guo, Alexandre Sablayrolles, Herv{\'{e}} J{\'{e}}gou, and Douwe Kiela. 2021.
\newblock \href {https://doi.org/10.18653/V1/2021.EMNLP-MAIN.464} {Gradient-based adversarial attacks against text transformers}.
\newblock In \emph{Proceedings of the 2021 Conference on Empirical Methods in Natural Language Processing, {EMNLP} 2021, Virtual Event / Punta Cana, Dominican Republic, 7-11 November, 2021}, pages 5747--5757. Association for Computational Linguistics.

\bibitem[{Gupta et~al.(2020)Gupta, Lin, Roth, Singh, and Gardner}]{GuptaLR0020}
Nitish Gupta, Kevin Lin, Dan Roth, Sameer Singh, and Matt Gardner. 2020.
\newblock \href {https://openreview.net/forum?id=SygWvAVFPr} {Neural module networks for reasoning over text}.
\newblock In \emph{8th International Conference on Learning Representations, {ICLR} 2020, Addis Ababa, Ethiopia, April 26-30, 2020}. OpenReview.net.

\bibitem[{Gupta et~al.(2022)Gupta, Shi, Gimpel, and Sachan}]{GuptaSGS22}
Vikram Gupta, Haoyue Shi, Kevin Gimpel, and Mrinmaya Sachan. 2022.
\newblock \href {https://doi.org/10.1609/AAAI.V36I10.21317} {Deep clustering of text representations for supervision-free probing of syntax}.
\newblock In \emph{Thirty-Sixth {AAAI} Conference on Artificial Intelligence, {AAAI} 2022, Thirty-Fourth Conference on Innovative Applications of Artificial Intelligence, {IAAI} 2022, The Twelveth Symposium on Educational Advances in Artificial Intelligence, {EAAI} 2022 Virtual Event, February 22 - March 1, 2022}, pages 10720--10728. {AAAI} Press.

\bibitem[{Gurnee et~al.(2024)Gurnee, Horsley, Guo, Kheirkhah, Sun, Hathaway, Nanda, and Bertsimas}]{Gurnee2024}
Wes Gurnee, Theo Horsley, Zifan~Carl Guo, Tara~Rezaei Kheirkhah, Qinyi Sun, Will Hathaway, Neel Nanda, and Dimitris Bertsimas. 2024.
\newblock \href {https://doi.org/10.48550/ARXIV.2401.12181} {Universal neurons in {GPT2} language models}.
\newblock \emph{CoRR}, abs/2401.12181.

\bibitem[{Haghighatkhah et~al.(2022)Haghighatkhah, Fokkens, Sommerauer, Speckmann, and Verbeek}]{HaghighatkhahFS22}
Pantea Haghighatkhah, Antske Fokkens, Pia Sommerauer, Bettina Speckmann, and Kevin Verbeek. 2022.
\newblock \href {https://doi.org/10.18653/V1/2022.EMNLP-MAIN.575} {Better hit the nail on the head than beat around the bush: Removing protected attributes with a single projection}.
\newblock In \emph{Proceedings of the 2022 Conference on Empirical Methods in Natural Language Processing, {EMNLP} 2022, Abu Dhabi, United Arab Emirates, December 7-11, 2022}, pages 8395--8416. Association for Computational Linguistics.

\bibitem[{Hartmann et~al.(2022)Hartmann, Heitmann, Siebert, and Schamp}]{Hartmann2022MoreTA}
Jochen Hartmann, Mark Heitmann, Christian Siebert, and Christina Schamp. 2022.
\newblock \href {https://api.semanticscholar.org/CorpusID:249903898} {More than a feeling: Accuracy and application of sentiment analysis}.
\newblock \emph{International Journal of Research in Marketing}.

\bibitem[{Hase et~al.(2023)Hase, Bansal, Kim, and Ghandeharioun}]{HaseBKG23}
Peter Hase, Mohit Bansal, Been Kim, and Asma Ghandeharioun. 2023.
\newblock \href {http://papers.nips.cc/paper\_files/paper/2023/hash/3927bbdcf0e8d1fa8aa23c26f358a281-Abstract-Conference.html} {Does localization inform editing? surprising differences in causality-based localization vs. knowledge editing in language models}.
\newblock In \emph{Advances in Neural Information Processing Systems 36: Annual Conference on Neural Information Processing Systems 2023, NeurIPS 2023, New Orleans, LA, USA, December 10 - 16, 2023}.

\bibitem[{Hawasly et~al.(2024)Hawasly, Dalvi, and Durrani}]{HawaslyDD24}
Majd Hawasly, Fahim Dalvi, and Nadir Durrani. 2024.
\newblock \href {https://aclanthology.org/2024.eacl-long.48} {Scaling up discovery of latent concepts in deep {NLP} models}.
\newblock In \emph{Proceedings of the 18th Conference of the European Chapter of the Association for Computational Linguistics, {EACL} 2024 - Volume 1: Long Papers, St. Julian's, Malta, March 17-22, 2024}, pages 793--806. Association for Computational Linguistics.

\bibitem[{Hewitt and Liang(2019)}]{HewittL19}
John Hewitt and Percy Liang. 2019.
\newblock \href {https://doi.org/10.18653/V1/D19-1275} {Designing and interpreting probes with control tasks}.
\newblock In \emph{Proceedings of the 2019 Conference on Empirical Methods in Natural Language Processing and the 9th International Joint Conference on Natural Language Processing, {EMNLP-IJCNLP} 2019, Hong Kong, China, November 3-7, 2019}, pages 2733--2743. Association for Computational Linguistics.

\bibitem[{Hill et~al.(2015)Hill, Reichart, and Korhonen}]{HillRK15}
Felix Hill, Roi Reichart, and Anna Korhonen. 2015.
\newblock \href {https://doi.org/10.1162/COLI\_A\_00237} {Simlex-999: Evaluating semantic models with (genuine) similarity estimation}.
\newblock \emph{Comput. Linguistics}, 41(4):665--695.

\bibitem[{Hohman et~al.(2019)Hohman, Head, Caruana, DeLine, and Drucker}]{HohmanHCDD19}
Fred Hohman, Andrew Head, Rich Caruana, Robert DeLine, and Steven~Mark Drucker. 2019.
\newblock \href {https://doi.org/10.1145/3290605.3300809} {Gamut: {A} design probe to understand how data scientists understand machine learning models}.
\newblock In \emph{Proceedings of the 2019 {CHI} Conference on Human Factors in Computing Systems, {CHI} 2019, Glasgow, Scotland, UK, May 04-09, 2019}, page 579. {ACM}.

\bibitem[{Hong et~al.(2023)Hong, Bhardwaj, Majumder, Aditya, and Poria}]{HongBMAP23}
Pengfei Hong, Rishabh Bhardwaj, Navonil Majumder, Somak Aditya, and Soujanya Poria. 2023.
\newblock \href {https://doi.org/10.18653/V1/2023.FINDINGS-ACL.231} {A robust information-masking approach for domain counterfactual generation}.
\newblock In \emph{Findings of the Association for Computational Linguistics: {ACL} 2023, Toronto, Canada, July 9-14, 2023}, pages 3756--3769. Association for Computational Linguistics.

\bibitem[{Howard et~al.(2022)Howard, Singer, Lal, Choi, and Swayamdipta}]{HowardSLCS22}
Phillip Howard, Gadi Singer, Vasudev Lal, Yejin Choi, and Swabha Swayamdipta. 2022.
\newblock \href {https://doi.org/10.18653/V1/2022.FINDINGS-EMNLP.371} {Neurocounterfactuals: Beyond minimal-edit counterfactuals for richer data augmentation}.
\newblock In \emph{Findings of the Association for Computational Linguistics: {EMNLP} 2022, Abu Dhabi, United Arab Emirates, December 7-11, 2022}, pages 5056--5072. Association for Computational Linguistics.

\bibitem[{Hristova and Netov(2022)}]{HristovaN22}
Gloria Hristova and Nikolay Netov. 2022.
\newblock \href {https://doi.org/10.1109/BIGDATA55660.2022.10020466} {Media coverage and public perception of distance learning during the {COVID-19} pandemic: {A} topic modeling approach based on bertopic}.
\newblock In \emph{{IEEE} International Conference on Big Data, Big Data 2022, Osaka, Japan, December 17-20, 2022}, pages 2259--2264. {IEEE}.

\bibitem[{Hu et~al.(2017)Hu, Andreas, Rohrbach, Darrell, and Saenko}]{HuARDS17}
Ronghang Hu, Jacob Andreas, Marcus Rohrbach, Trevor Darrell, and Kate Saenko. 2017.
\newblock \href {https://doi.org/10.1109/ICCV.2017.93} {Learning to reason: End-to-end module networks for visual question answering}.
\newblock In \emph{{IEEE} International Conference on Computer Vision, {ICCV} 2017, Venice, Italy, October 22-29, 2017}, pages 804--813. {IEEE} Computer Society.

\bibitem[{Jacovi and Goldberg(2020)}]{Jacovi2020}
Alon Jacovi and Yoav Goldberg. 2020.
\newblock \href {https://doi.org/10.18653/V1/2020.ACL-MAIN.386} {Towards faithfully interpretable {NLP} systems: How should we define and evaluate faithfulness?}
\newblock In \emph{Proceedings of the 58th Annual Meeting of the Association for Computational Linguistics, {ACL} 2020, Online, July 5-10, 2020}, pages 4198--4205. Association for Computational Linguistics.

\bibitem[{Jain and Wallace(2019)}]{JainW19}
Sarthak Jain and Byron~C. Wallace. 2019.
\newblock \href {https://doi.org/10.18653/V1/N19-1357} {Attention is not explanation}.
\newblock In \emph{Proceedings of the 2019 Conference of the North American Chapter of the Association for Computational Linguistics: Human Language Technologies, {NAACL-HLT} 2019, Minneapolis, MN, USA, June 2-7, 2019, Volume 1 (Long and Short Papers)}, pages 3543--3556. Association for Computational Linguistics.

\bibitem[{Jain et~al.(2020)Jain, Wiegreffe, Pinter, and Wallace}]{JainWPW20}
Sarthak Jain, Sarah Wiegreffe, Yuval Pinter, and Byron~C. Wallace. 2020.
\newblock \href {https://doi.org/10.18653/V1/2020.ACL-MAIN.409} {Learning to faithfully rationalize by construction}.
\newblock In \emph{Proceedings of the 58th Annual Meeting of the Association for Computational Linguistics, {ACL} 2020, Online, July 5-10, 2020}, pages 4459--4473. Association for Computational Linguistics.

\bibitem[{Jo and Myaeng(2020)}]{JoM20}
Jae{-}young Jo and Sung{-}Hyon Myaeng. 2020.
\newblock \href {https://doi.org/10.18653/V1/2020.ACL-MAIN.311} {Roles and utilization of attention heads in transformer-based neural language models}.
\newblock In \emph{Proceedings of the 58th Annual Meeting of the Association for Computational Linguistics, {ACL} 2020, Online, July 5-10, 2020}, pages 3404--3417. Association for Computational Linguistics.

\bibitem[{Joshi et~al.(2022)Joshi, Chan, Liu, Nie, Sanjabi, Firooz, and Ren}]{JoshiCLNSF022}
Brihi Joshi, Aaron Chan, Ziyi Liu, Shaoliang Nie, Maziar Sanjabi, Hamed Firooz, and Xiang Ren. 2022.
\newblock \href {https://doi.org/10.18653/V1/2022.FINDINGS-EMNLP.242} {Er-test: Evaluating explanation regularization methods for language models}.
\newblock In \emph{Findings of the Association for Computational Linguistics: {EMNLP} 2022, Abu Dhabi, United Arab Emirates, December 7-11, 2022}, pages 3315--3336. Association for Computational Linguistics.

\bibitem[{Jumelet and Hupkes(2018)}]{JumeletH18}
Jaap Jumelet and Dieuwke Hupkes. 2018.
\newblock \href {https://doi.org/10.18653/V1/W18-5424} {Do language models understand anything? on the ability of lstms to understand negative polarity items}.
\newblock In \emph{Proceedings of the Workshop: Analyzing and Interpreting Neural Networks for NLP, BlackboxNLP@EMNLP 2018, Brussels, Belgium, November 1, 2018}, pages 222--231. Association for Computational Linguistics.

\bibitem[{Kabir et~al.(2024)Kabir, Li, and Zhang}]{KabirL024}
Samia Kabir, Lixiang Li, and Tianyi Zhang. 2024.
\newblock \href {https://doi.org/10.1145/3613904.3642111} {{STILE:} exploring and debugging social biases in pre-trained text representations}.
\newblock In \emph{Proceedings of the {CHI} Conference on Human Factors in Computing Systems, {CHI} 2024, Honolulu, HI, USA, May 11-16, 2024}, pages 293:1--293:20. {ACM}.

\bibitem[{Karran et~al.(2022)Karran, Demazure, Hudon, S{\'e}n{\'e}cal, and L{\'e}ger}]{Karran2022DesigningFC}
Alexander~John Karran, Th{\'e}ophile Demazure, Antoine Hudon, Sylvain S{\'e}n{\'e}cal, and Pierre-Majorique L{\'e}ger. 2022.
\newblock \href {https://api.semanticscholar.org/CorpusID:250034557} {Designing for confidence: The impact of visualizing artificial intelligence decisions}.
\newblock \emph{Frontiers in Neuroscience}, 16.

\bibitem[{Kasneci et~al.(2023)Kasneci, Se{\ss}ler, K{\"u}chemann, Bannert, Dementieva, and et~al.}]{Kasneci2023ChatGPTFG}
Enkelejda Kasneci, Kathrin Se{\ss}ler, Stefan K{\"u}chemann, Maria Bannert, Daryna Dementieva, and et~al. 2023.
\newblock \href {https://api.semanticscholar.org/CorpusID:257445349} {Chatgpt for good? on opportunities and challenges of large language models for education}.
\newblock \emph{Learning and Individual Differences}.

\bibitem[{Katz and Belinkov(2023)}]{KatzB23}
Shahar Katz and Yonatan Belinkov. 2023.
\newblock \href {https://doi.org/10.18653/V1/2023.FINDINGS-EMNLP.939} {{VISIT:} visualizing and interpreting the semantic information flow of transformers}.
\newblock In \emph{Findings of the Association for Computational Linguistics: {EMNLP} 2023, Singapore, December 6-10, 2023}, pages 14094--14113. Association for Computational Linguistics.

\bibitem[{Kaur et~al.(2020)Kaur, Nori, Jenkins, Caruana, Wallach, and Vaughan}]{KaurNJCWV20}
Harmanpreet Kaur, Harsha Nori, Samuel Jenkins, Rich Caruana, Hanna~M. Wallach, and Jennifer~Wortman Vaughan. 2020.
\newblock \href {https://doi.org/10.1145/3313831.3376219} {Interpreting interpretability: Understanding data scientists' use of interpretability tools for machine learning}.
\newblock In \emph{{CHI} '20: {CHI} Conference on Human Factors in Computing Systems, Honolulu, HI, USA, April 25-30, 2020}, pages 1--14. {ACM}.

\bibitem[{Kaur et~al.(2021)Kaur, Nori, Jenkins, Caruana, Wallach, and Vaughan}]{KaurNJCWV21}
Harmanpreet Kaur, Harsha Nori, Samuel Jenkins, Rich Caruana, Hanna~M. Wallach, and Jennifer~Wortman Vaughan. 2021.
\newblock \href {https://github.com/valahu/dash/blob/main/KDD2021/paper\_5\_invited\_Kaur.pdf} {Interpreting interpretability: Understanding data scientists' use of interpretability tools for machine learning}.
\newblock In \emph{3rd Workshop on Data Science with Human in the Loop, DaSH@KDD, Virtual Conference, August 15, 2021}.

\bibitem[{Kaushik et~al.(2020)Kaushik, Hovy, and Lipton}]{KaushikHL20}
Divyansh Kaushik, Eduard~H. Hovy, and Zachary~Chase Lipton. 2020.
\newblock \href {https://openreview.net/forum?id=Sklgs0NFvr} {Learning the difference that makes {A} difference with counterfactually-augmented data}.
\newblock In \emph{8th International Conference on Learning Representations, {ICLR} 2020, Addis Ababa, Ethiopia, April 26-30, 2020}. OpenReview.net.

\bibitem[{Kaushik et~al.(2021)Kaushik, Setlur, Hovy, and Lipton}]{KaushikSHL21}
Divyansh Kaushik, Amrith Setlur, Eduard~H. Hovy, and Zachary~Chase Lipton. 2021.
\newblock \href {https://openreview.net/forum?id=HHiiQKWsOcV} {Explaining the efficacy of counterfactually augmented data}.
\newblock In \emph{9th International Conference on Learning Representations, {ICLR} 2021, Virtual Event, Austria, May 3-7, 2021}. OpenReview.net.

\bibitem[{Keidar et~al.(2022)Keidar, Opedal, Jin, and Sachan}]{KeidarOJS22}
Daphna Keidar, Andreas Opedal, Zhijing Jin, and Mrinmaya Sachan. 2022.
\newblock \href {https://doi.org/10.18653/V1/2022.ACL-LONG.101} {Slangvolution: {A} causal analysis of semantic change and frequency dynamics in slang}.
\newblock In \emph{Proceedings of the 60th Annual Meeting of the Association for Computational Linguistics (Volume 1: Long Papers), {ACL} 2022, Dublin, Ireland, May 22-27, 2022}, pages 1422--1442. Association for Computational Linguistics.

\bibitem[{Kim et~al.(2023)Kim, Joo, Kim, Jang, Ye, Shin, and Seo}]{KimJKJYSS23}
Seungone Kim, Se~June Joo, Doyoung Kim, Joel Jang, Seonghyeon Ye, Jamin Shin, and Minjoon Seo. 2023.
\newblock \href {https://doi.org/10.18653/V1/2023.EMNLP-MAIN.782} {The cot collection: Improving zero-shot and few-shot learning of language models via chain-of-thought fine-tuning}.
\newblock In \emph{Proceedings of the 2023 Conference on Empirical Methods in Natural Language Processing, {EMNLP} 2023, Singapore, December 6-10, 2023}, pages 12685--12708. Association for Computational Linguistics.

\bibitem[{King and Falkedal(1990)}]{KingF90}
Margaret King and Kirsten Falkedal. 1990.
\newblock \href {https://aclanthology.org/C90-2037/} {Using test suites in evaluation of machine translation systems}.
\newblock In \emph{13th International Conference on Computational Linguistics, {COLING} 1990, University of Helsinki, Finland, August 20-25, 1990}, pages 211--216.

\bibitem[{Koh et~al.(2020)Koh, Nguyen, Tang, Mussmann, Pierson, Kim, and Liang}]{Koh2020}
Pang~Wei Koh, Thao Nguyen, Yew~Siang Tang, Stephen Mussmann, Emma Pierson, Been Kim, and Percy Liang. 2020.
\newblock \href {https://arxiv.org/abs/2007.04612} {Concept bottleneck models}.
\newblock \emph{CoRR}, abs/2007.04612.

\bibitem[{Kokalj et~al.(2021)Kokalj, Skrlj, Lavrac, Pollak, and Robnik{-}Sikonja}]{KokaljSLPR21}
Enja Kokalj, Blaz Skrlj, Nada Lavrac, Senja Pollak, and Marko Robnik{-}Sikonja. 2021.
\newblock \href {https://aclanthology.org/2021.hackashop-1.3/} {{BERT} meets shapley: Extending {SHAP} explanations to transformer-based classifiers}.
\newblock In \emph{Proceedings of the {EACL} Hackashop on News Media Content Analysis and Automated Report Generation, {EACL} 2021, Online, April 19, 2021}, pages 16--21. Association for Computational Linguistics.

\bibitem[{Kovaleva et~al.(2019)Kovaleva, Romanov, Rogers, and Rumshisky}]{KovalevaRRR19}
Olga Kovaleva, Alexey Romanov, Anna Rogers, and Anna Rumshisky. 2019.
\newblock \href {https://doi.org/10.18653/V1/D19-1445} {Revealing the dark secrets of {BERT}}.
\newblock In \emph{Proceedings of the 2019 Conference on Empirical Methods in Natural Language Processing and the 9th International Joint Conference on Natural Language Processing, {EMNLP-IJCNLP} 2019, Hong Kong, China, November 3-7, 2019}, pages 4364--4373. Association for Computational Linguistics.

\bibitem[{Kram{\'{a}}r et~al.(2024)Kram{\'{a}}r, Lieberum, Shah, and Nanda}]{Kramar2024}
J{\'{a}}nos Kram{\'{a}}r, Tom Lieberum, Rohin Shah, and Neel Nanda. 2024.
\newblock \href {https://doi.org/10.48550/ARXIV.2403.00745} {Atp*: An efficient and scalable method for localizing {LLM} behaviour to components}.
\newblock \emph{CoRR}, abs/2403.00745.

\bibitem[{Krishnan(2019)}]{Krishnan2019AgainstIA}
Maya Krishnan. 2019.
\newblock \href {https://api.semanticscholar.org/CorpusID:202287978} {Against interpretability: a critical examination of the interpretability problem in machine learning}.
\newblock \emph{Philosophy \& Technology}, 33:487 -- 502.

\bibitem[{Kumar et~al.(2022)Kumar, Tan, and Sharma}]{KumarT022}
Abhinav Kumar, Chenhao Tan, and Amit Sharma. 2022.
\newblock \href {http://papers.nips.cc/paper\_files/paper/2022/hash/725f5e8036cc08adeba4a7c3bcbc6f2c-Abstract-Conference.html} {Probing classifiers are unreliable for concept removal and detection}.
\newblock In \emph{Advances in Neural Information Processing Systems 35: Annual Conference on Neural Information Processing Systems 2022, NeurIPS 2022, New Orleans, LA, USA, November 28 - December 9, 2022}.

\bibitem[{Kumar and Talukdar(2020)}]{KumarT20}
Sawan Kumar and Partha~P. Talukdar. 2020.
\newblock \href {https://doi.org/10.18653/V1/2020.ACL-MAIN.771} {{NILE} : Natural language inference with faithful natural language explanations}.
\newblock In \emph{Proceedings of the 58th Annual Meeting of the Association for Computational Linguistics, {ACL} 2020, Online, July 5-10, 2020}, pages 8730--8742. Association for Computational Linguistics.

\bibitem[{Lai and Tan(2019)}]{LaiT19}
Vivian Lai and Chenhao Tan. 2019.
\newblock \href {https://doi.org/10.1145/3287560.3287590} {On human predictions with explanations and predictions of machine learning models: {A} case study on deception detection}.
\newblock In \emph{Proceedings of the Conference on Fairness, Accountability, and Transparency, FAT* 2019, Atlanta, GA, USA, January 29-31, 2019}, pages 29--38. {ACM}.

\bibitem[{Lanham et~al.(2023)Lanham, Chen, Radhakrishnan, Steiner, Denison, and et~al.}]{Lanham2023}
Tamera Lanham, Anna Chen, Ansh Radhakrishnan, Benoit Steiner, Carson Denison, and et~al. 2023.
\newblock \href {https://doi.org/10.48550/ARXIV.2307.13702} {Measuring faithfulness in chain-of-thought reasoning}.
\newblock \emph{CoRR}, abs/2307.13702.

\bibitem[{Laskar et~al.(2024)Laskar, Alqahtani, Bari, Rahman, and et~al.}]{Laskar2024ASS}
Md~Tahmid~Rahman Laskar, Sawsan Alqahtani, M~Saiful Bari, Mizanur Rahman, and et~al. 2024.
\newblock \href {https://api.semanticscholar.org/CorpusID:271038835} {A systematic survey and critical review on evaluating large language models: Challenges, limitations, and recommendations}.

\bibitem[{Lazer et~al.(2020)Lazer, Pentland, Watts, Aral, Athey, Contractor, Freelon, Gonzalez-Bailon, King, Margetts et~al.}]{lazer2020computational}
David~MJ Lazer, Alex Pentland, Duncan~J Watts, Sinan Aral, Susan Athey, Noshir Contractor, Deen Freelon, Sandra Gonzalez-Bailon, Gary King, Helen Margetts, et~al. 2020.
\newblock \href {https://www.science.org/doi/full/10.1126/science.aaz8170?casa_token=mzc9aF3KetwAAAAA%3AI5WZnAUxZk_hmSSWH3g0BE1DDnpwVp3ISVXA07qFV8OO5ObBaBkLSN4PK3FTT7jw1j-Lg2FANRBPtw} {Computational social science: Obstacles and opportunities}.
\newblock \emph{Science}, 369(6507):1060--1062.

\bibitem[{Lehmann et~al.(1996)Lehmann, Oepen, Regnier{-}Prost, Netter, Lux, Klein, Falkedal, Fouvry, Estival, Dauphin, Compagnion, Baur, Balkan, and Arnold}]{LehmannORNLKFFEDCBBA96}
Sabine Lehmann, Stephan Oepen, Sylvie Regnier{-}Prost, Klaus Netter, Veronika Lux, Judith Klein, Kirsten Falkedal, Frederik Fouvry, Dominique Estival, Eva Dauphin, Herve Compagnion, Judith Baur, Lorna Balkan, and Doug Arnold. 1996.
\newblock \href {https://aclanthology.org/C96-2120/} {{TSNLP} - test suites for natural language processing}.
\newblock In \emph{16th International Conference on Computational Linguistics, Proceedings of the Conference, {COLING} 1996, Center for Sprogteknologi, Copenhagen, Denmark, August 5-9, 1996}, pages 711--716.

\bibitem[{Lei et~al.(2016)Lei, Barzilay, and Jaakkola}]{LeiBJ16}
Tao Lei, Regina Barzilay, and Tommi~S. Jaakkola. 2016.
\newblock \href {https://doi.org/10.18653/V1/D16-1011} {Rationalizing neural predictions}.
\newblock In \emph{Proceedings of the 2016 Conference on Empirical Methods in Natural Language Processing, {EMNLP} 2016, Austin, Texas, USA, November 1-4, 2016}, pages 107--117. The Association for Computational Linguistics.

\bibitem[{Lepori and McCoy(2020)}]{LeporiM20}
Michael~A. Lepori and R.~Thomas McCoy. 2020.
\newblock \href {https://doi.org/10.18653/V1/2020.COLING-MAIN.325} {Picking bert's brain: Probing for linguistic dependencies in contextualized embeddings using representational similarity analysis}.
\newblock In \emph{Proceedings of the 28th International Conference on Computational Linguistics, {COLING} 2020, Barcelona, Spain (Online), December 8-13, 2020}, pages 3637--3651. International Committee on Computational Linguistics.

\bibitem[{Lertvittayakumjorn and Toni(2021)}]{Lertvittayakumjorn21}
Piyawat Lertvittayakumjorn and Francesca Toni. 2021.
\newblock \href {https://doi.org/10.1162/TACL\_A\_00440} {Explanation-based human debugging of {NLP} models: {A} survey}.
\newblock \emph{Trans. Assoc. Comput. Linguistics}, 9:1508--1528.

\bibitem[{Leviant and Reichart(2015)}]{LeviantR15}
Ira Leviant and Roi Reichart. 2015.
\newblock \href {https://arxiv.org/abs/1508.00106} {Judgment language matters: Multilingual vector space models for judgment language aware lexical semantics}.
\newblock \emph{CoRR}, abs/1508.00106.

\bibitem[{Li et~al.(2019)Li, Ji, Du, Li, and Wang}]{LiJDLW19}
Jinfeng Li, Shouling Ji, Tianyu Du, Bo~Li, and Ting Wang. 2019.
\newblock \href {https://www.ndss-symposium.org/ndss-paper/textbugger-generating-adversarial-text-against-real-world-applications/} {Textbugger: Generating adversarial text against real-world applications}.
\newblock In \emph{26th Annual Network and Distributed System Security Symposium, {NDSS} 2019, San Diego, California, USA, February 24-27, 2019}. The Internet Society.

\bibitem[{Li et~al.(2016)Li, Monroe, and Jurafsky}]{LiMJ16a}
Jiwei Li, Will Monroe, and Dan Jurafsky. 2016.
\newblock \href {https://arxiv.org/abs/1612.08220} {Understanding neural networks through representation erasure}.
\newblock \emph{CoRR}, abs/1612.08220.

\bibitem[{Li et~al.(2022)Li, Li, Shang, Dong, Sun, Liu, Ji, Jiang, and Liu}]{LiLSDSLJJL22}
Shaobo Li, Xiaoguang Li, Lifeng Shang, Zhenhua Dong, Chengjie Sun, Bingquan Liu, Zhenzhou Ji, Xin Jiang, and Qun Liu. 2022.
\newblock \href {https://doi.org/10.18653/V1/2022.FINDINGS-ACL.136} {How pre-trained language models capture factual knowledge? {A} causal-inspired analysis}.
\newblock In \emph{Findings of the Association for Computational Linguistics: {ACL} 2022, Dublin, Ireland, May 22-27, 2022}, pages 1720--1732. Association for Computational Linguistics.

\bibitem[{Li et~al.(2024)Li, Xu, Miao, Zhou, and Qian}]{Li2024}
Yongqi Li, Mayi Xu, Xin Miao, Shen Zhou, and Tieyun Qian. 2024.
\newblock \href {https://aclanthology.org/2024.lrec-main.1156} {Prompting large language models for counterfactual generation: An empirical study}.
\newblock In \emph{Proceedings of the 2024 Joint International Conference on Computational Linguistics, Language Resources and Evaluation, {LREC/COLING} 2024, 20-25 May, 2024, Torino, Italy}, pages 13201--13221. {ELRA} and {ICCL}.

\bibitem[{Liang et~al.(2024)Liang, Izzo, Zhang, Lepp, Cao, Zhao, Chen, Ye, Liu, Huang, McFarland, and Zou}]{Liang2024}
Weixin Liang, Zachary Izzo, Yaohui Zhang, Haley Lepp, Hancheng Cao, Xuandong Zhao, Lingjiao Chen, Haotian Ye, Sheng Liu, Zhi Huang, Daniel~A. McFarland, and James~Y. Zou. 2024.
\newblock \href {https://doi.org/10.48550/ARXIV.2403.07183} {Monitoring ai-modified content at scale: {A} case study on the impact of chatgpt on {AI} conference peer reviews}.
\newblock \emph{CoRR}, abs/2403.07183.

\bibitem[{Ling et~al.(2017)Ling, Yogatama, Dyer, and Blunsom}]{LingYDB17}
Wang Ling, Dani Yogatama, Chris Dyer, and Phil Blunsom. 2017.
\newblock \href {https://doi.org/10.18653/V1/P17-1015} {Program induction by rationale generation: Learning to solve and explain algebraic word problems}.
\newblock In \emph{Proceedings of the 55th Annual Meeting of the Association for Computational Linguistics, {ACL} 2017, Vancouver, Canada, July 30 - August 4, Volume 1: Long Papers}, pages 158--167. Association for Computational Linguistics.

\bibitem[{Lipton(2018)}]{Lipton18}
Zachary~C. Lipton. 2018.
\newblock \href {https://doi.org/10.1145/3233231} {The mythos of model interpretability}.
\newblock \emph{Commun. {ACM}}, 61(10):36--43.

\bibitem[{Lissak et~al.(2024{\natexlab{a}})Lissak, Calderon, Shenkman, Ophir, Fruchter, Klomek, and Reichart}]{Lissak2024_b}
Shir Lissak, Nitay Calderon, Geva Shenkman, Yaakov Ophir, Eyal Fruchter, Anat~Brunstein Klomek, and Roi Reichart. 2024{\natexlab{a}}.
\newblock \href {https://doi.org/10.48550/ARXIV.2402.11886} {The colorful future of llms: Evaluating and improving llms as emotional supporters for queer youth}.
\newblock \emph{CoRR}, abs/2402.11886.

\bibitem[{Lissak et~al.(2024{\natexlab{b}})Lissak, Ophir, Tikochinski, Brunstein~Klomek, Sisso, Fruchter, and Reichart}]{Lissak2024}
Shir Lissak, Yaakov Ophir, Refael Tikochinski, Anat Brunstein~Klomek, Itay Sisso, Eyal Fruchter, and Roi Reichart. 2024{\natexlab{b}}.
\newblock \href {https://doi.org/10.3389/fpsyt.2024.1328122} {Bored to death: Artificial intelligence research reveals the role of boredom in suicide behavior}.
\newblock \emph{Frontiers in Psychiatry}, 15.

\bibitem[{Liu et~al.(2023)Liu, Chaudhary, and Wang}]{Liu2023}
Haoyang Liu, Maheep Chaudhary, and Haohan Wang. 2023.
\newblock \href {https://doi.org/10.48550/ARXIV.2307.16851} {Towards trustworthy and aligned machine learning: {A} data-centric survey with causality perspectives}.
\newblock \emph{CoRR}, abs/2307.16851.

\bibitem[{Ludan et~al.(2023)Ludan, Lyu, Yang, Dugan, Yatskar, and Callison{-}Burch}]{Ludan2023}
Josh~Magnus Ludan, Qing Lyu, Yue Yang, Liam Dugan, Mark Yatskar, and Chris Callison{-}Burch. 2023.
\newblock \href {https://doi.org/10.48550/ARXIV.2310.19660} {Interpretable-by-design text classification with iteratively generated concept bottleneck}.
\newblock \emph{CoRR}, abs/2310.19660.

\bibitem[{Lundberg and Lee(2017)}]{LundbergL17}
Scott~M. Lundberg and Su{-}In Lee. 2017.
\newblock \href {https://proceedings.neurips.cc/paper/2017/hash/8a20a8621978632d76c43dfd28b67767-Abstract.html} {A unified approach to interpreting model predictions}.
\newblock In \emph{Advances in Neural Information Processing Systems 30: Annual Conference on Neural Information Processing Systems 2017, December 4-9, 2017, Long Beach, CA, {USA}}, pages 4765--4774.

\bibitem[{Luo et~al.(2024)Luo, Ivison, Han, and Poon}]{LuoIHP24}
Siwen Luo, Hamish Ivison, Soyeon~Caren Han, and Josiah Poon. 2024.
\newblock \href {https://doi.org/10.1145/3649450} {Local interpretations for explainable natural language processing: {A} survey}.
\newblock \emph{{ACM} Comput. Surv.}, 56(9):232:1--232:36.

\bibitem[{Lyu et~al.(2022)Lyu, Apidianaki, and Callison{-}Burch}]{Lyu2022}
Qing Lyu, Marianna Apidianaki, and Chris Callison{-}Burch. 2022.
\newblock \href {https://doi.org/10.48550/ARXIV.2209.11326} {Towards faithful model explanation in {NLP:} {A} survey}.
\newblock \emph{CoRR}, abs/2209.11326.

\bibitem[{Lyu et~al.(2023)Lyu, Havaldar, Stein, Zhang, Rao, Wong, Apidianaki, and Callison{-}Burch}]{LyuHSZRWAC23}
Qing Lyu, Shreya Havaldar, Adam Stein, Li~Zhang, Delip Rao, Eric Wong, Marianna Apidianaki, and Chris Callison{-}Burch. 2023.
\newblock \href {https://doi.org/10.18653/V1/2023.IJCNLP-MAIN.20} {Faithful chain-of-thought reasoning}.
\newblock In \emph{Proceedings of the 13th International Joint Conference on Natural Language Processing and the 3rd Conference of the Asia-Pacific Chapter of the Association for Computational Linguistics, {IJCNLP} 2023 -Volume 1: Long Papers, Nusa Dua, Bali, November 1 - 4, 2023}, pages 305--329. Association for Computational Linguistics.

\bibitem[{MacCarthy(2019)}]{maccarthy2019examination}
Mark MacCarthy. 2019.
\newblock \href {https://papers.ssrn.com/sol3/papers.cfm?abstract_id=3615731} {An examination of the algorithmic accountability act of 2019}.
\newblock \emph{Available at SSRN 3615731}.

\bibitem[{Madaan et~al.(2023)Madaan, Hermann, and Yazdanbakhsh}]{MadaanHY23}
Aman Madaan, Katherine Hermann, and Amir Yazdanbakhsh. 2023.
\newblock \href {https://doi.org/10.18653/V1/2023.FINDINGS-EMNLP.101} {What makes chain-of-thought prompting effective? {A} counterfactual study}.
\newblock In \emph{Findings of the Association for Computational Linguistics: {EMNLP} 2023, Singapore, December 6-10, 2023}, pages 1448--1535. Association for Computational Linguistics.

\bibitem[{Madsen et~al.(2023)Madsen, Reddy, and Chandar}]{MadsenRC23}
Andreas Madsen, Siva Reddy, and Sarath Chandar. 2023.
\newblock \href {https://doi.org/10.1145/3546577} {Post-hoc interpretability for neural {NLP:} {A} survey}.
\newblock \emph{{ACM} Comput. Surv.}, 55(8):155:1--155:42.

\bibitem[{Marasovic et~al.(2022)Marasovic, Beltagy, Downey, and Peters}]{MarasovicBDP22}
Ana Marasovic, Iz~Beltagy, Doug Downey, and Matthew~E. Peters. 2022.
\newblock \href {https://doi.org/10.18653/V1/2022.FINDINGS-NAACL.31} {Few-shot self-rationalization with natural language prompts}.
\newblock In \emph{Findings of the Association for Computational Linguistics: {NAACL} 2022, Seattle, WA, United States, July 10-15, 2022}, pages 410--424. Association for Computational Linguistics.

\bibitem[{McKenzie et~al.(2023)McKenzie, Lyzhov, Pieler, Parrish, Mueller, and et~al.}]{McKenzie2023}
Ian~R. McKenzie, Alexander Lyzhov, Michael Pieler, Alicia Parrish, Aaron Mueller, and et~al. 2023.
\newblock \href {https://doi.org/10.48550/ARXIV.2306.09479} {Inverse scaling: When bigger isn't better}.
\newblock \emph{CoRR}, abs/2306.09479.

\bibitem[{Meng et~al.(2022)Meng, Bau, Andonian, and Belinkov}]{MengBAB22}
Kevin Meng, David Bau, Alex Andonian, and Yonatan Belinkov. 2022.
\newblock \href {http://papers.nips.cc/paper\_files/paper/2022/hash/6f1d43d5a82a37e89b0665b33bf3a182-Abstract-Conference.html} {Locating and editing factual associations in {GPT}}.
\newblock In \emph{Advances in Neural Information Processing Systems 35: Annual Conference on Neural Information Processing Systems 2022, NeurIPS 2022, New Orleans, LA, USA, November 28 - December 9, 2022}.

\bibitem[{Menon et~al.(2023)Menon, Zaman, and Srivastava}]{menon-etal-2023-mantle}
Rakesh Menon, Kerem Zaman, and Shashank Srivastava. 2023.
\newblock \href {https://doi.org/10.18653/v1/2023.emnlp-main.832} {{M}a{N}t{LE}: Model-agnostic natural language explainer}.
\newblock In \emph{Proceedings of the 2023 Conference on Empirical Methods in Natural Language Processing}, pages 13493--13511, Singapore. Association for Computational Linguistics.

\bibitem[{Michael et~al.(2020)Michael, Botha, and Tenney}]{MichaelBT20}
Julian Michael, Jan~A. Botha, and Ian Tenney. 2020.
\newblock \href {https://doi.org/10.18653/V1/2020.EMNLP-MAIN.552} {Asking without telling: Exploring latent ontologies in contextual representations}.
\newblock In \emph{Proceedings of the 2020 Conference on Empirical Methods in Natural Language Processing, {EMNLP} 2020, Online, November 16-20, 2020}, pages 6792--6812. Association for Computational Linguistics.

\bibitem[{Miller et~al.(2016)Miller, Johns, Mok, Gowda, Sirkin, Lee, and Ju}]{miller2016behavioral}
David Miller, Mishel Johns, Brian Mok, Nikhil Gowda, David Sirkin, Key Lee, and Wendy Ju. 2016.
\newblock \href {https://journals.sagepub.com/doi/abs/10.1177/1541931213601422} {Behavioral measurement of trust in automation: the trust fall}.
\newblock In \emph{Proceedings of the human factors and ergonomics society annual meeting}, volume~60, pages 1849--1853. SAGE Publications Sage CA: Los Angeles, CA.

\bibitem[{Miller(2017)}]{Miller2017ExplanationIA}
Tim Miller. 2017.
\newblock \href {https://api.semanticscholar.org/CorpusID:36024272} {Explanation in artificial intelligence: Insights from the social sciences}.
\newblock \emph{Artif. Intell.}, 267:1--38.

\bibitem[{Montavon et~al.(2019)Montavon, Binder, Lapuschkin, Samek, and M{\"{u}}ller}]{MontavonBLSM19}
Gr{\'{e}}goire Montavon, Alexander Binder, Sebastian Lapuschkin, Wojciech Samek, and Klaus{-}Robert M{\"{u}}ller. 2019.
\newblock \href {https://doi.org/10.1007/978-3-030-28954-6\_10} {Layer-wise relevance propagation: An overview}.
\newblock In Wojciech Samek, Gr{\'{e}}goire Montavon, Andrea Vedaldi, Lars~Kai Hansen, and Klaus{-}Robert M{\"{u}}ller, editors, \emph{Explainable {AI:} Interpreting, Explaining and Visualizing Deep Learning}, volume 11700 of \emph{Lecture Notes in Computer Science}, pages 193--209. Springer.

\bibitem[{Morris et~al.(2020)Morris, Lifland, Yoo, Grigsby, Jin, and Qi}]{MorrisLYGJQ20}
John~X. Morris, Eli Lifland, Jin~Yong Yoo, Jake Grigsby, Di~Jin, and Yanjun Qi. 2020.
\newblock \href {https://doi.org/10.18653/V1/2020.EMNLP-DEMOS.16} {Textattack: {A} framework for adversarial attacks, data augmentation, and adversarial training in {NLP}}.
\newblock In \emph{Proceedings of the 2020 Conference on Empirical Methods in Natural Language Processing: System Demonstrations, {EMNLP} 2020 - Demos, Online, November 16-20, 2020}, pages 119--126. Association for Computational Linguistics.

\bibitem[{Mosbach et~al.(2024)Mosbach, Gautam, Browne, Klakow, and Geva}]{MosbachGBKG24}
Marius Mosbach, Vagrant Gautam, Tom{\'{a}}s~Vergara Browne, Dietrich Klakow, and Mor Geva. 2024.
\newblock \href {https://aclanthology.org/2024.emnlp-main.181} {From insights to actions: The impact of interpretability and analysis research on {NLP}}.
\newblock In \emph{Proceedings of the 2024 Conference on Empirical Methods in Natural Language Processing, {EMNLP} 2024, Miami, FL, USA, November 12-16, 2024}, pages 3078--3105. Association for Computational Linguistics.

\bibitem[{Mosca et~al.(2022)Mosca, Szigeti, Tragianni, Gallagher, and Groh}]{MoscaSTGG22}
Edoardo Mosca, Ferenc Szigeti, Stella Tragianni, Daniel Gallagher, and Georg Groh. 2022.
\newblock \href {https://aclanthology.org/2022.coling-1.406} {Shap-based explanation methods: {A} review for {NLP} interpretability}.
\newblock In \emph{Proceedings of the 29th International Conference on Computational Linguistics, {COLING} 2022, Gyeongju, Republic of Korea, October 12-17, 2022}, pages 4593--4603. International Committee on Computational Linguistics.

\bibitem[{Mousi et~al.(2023)Mousi, Durrani, and Dalvi}]{MousiDD23}
Basel Mousi, Nadir Durrani, and Fahim Dalvi. 2023.
\newblock \href {https://doi.org/10.18653/V1/2023.EMNLP-MAIN.196} {Can llms facilitate interpretation of pre-trained language models?}
\newblock In \emph{Proceedings of the 2023 Conference on Empirical Methods in Natural Language Processing, {EMNLP} 2023, Singapore, December 6-10, 2023}, pages 3248--3268. Association for Computational Linguistics.

\bibitem[{M{\"{u}}ller(2024)}]{Muller2024}
Romy M{\"{u}}ller. 2024.
\newblock \href {https://doi.org/10.48550/ARXIV.2404.16042} {How explainable {AI} affects human performance: {A} systematic review of the behavioural consequences of saliency maps}.
\newblock \emph{CoRR}, abs/2404.16042.

\bibitem[{Murdoch et~al.(2019)Murdoch, Singh, Kumbier, Abbasi-Asl, and Yu}]{murdoch2019definitions}
W~James Murdoch, Chandan Singh, Karl Kumbier, Reza Abbasi-Asl, and Bin Yu. 2019.
\newblock \href {https://www.pnas.org/doi/abs/10.1073/pnas.1900654116} {Definitions, methods, and applications in interpretable machine learning}.
\newblock \emph{Proceedings of the National Academy of Sciences}, 116(44):22071--22080.

\bibitem[{Narang et~al.(2020)Narang, Raffel, Lee, Roberts, Fiedel, and Malkan}]{Narang2020}
Sharan Narang, Colin Raffel, Katherine Lee, Adam Roberts, Noah Fiedel, and Karishma Malkan. 2020.
\newblock \href {https://arxiv.org/abs/2004.14546} {Wt5?! training text-to-text models to explain their predictions}.
\newblock \emph{CoRR}, abs/2004.14546.

\bibitem[{Newman et~al.(2021)Newman, Ang, Gong, and Hewitt}]{NewmanAGH21}
Benjamin Newman, Kai{-}Siang Ang, Julia Gong, and John Hewitt. 2021.
\newblock \href {https://doi.org/10.18653/V1/2021.NAACL-MAIN.290} {Refining targeted syntactic evaluation of language models}.
\newblock In \emph{Proceedings of the 2021 Conference of the North American Chapter of the Association for Computational Linguistics: Human Language Technologies, {NAACL-HLT} 2021, Online, June 6-11, 2021}, pages 3710--3723. Association for Computational Linguistics.

\bibitem[{Nguyen et~al.(2024)Nguyen, Youssef, Schl{\"{o}}tterer, and Seifert}]{Nguyen2024}
Van~Bach Nguyen, Paul Youssef, J{\"{o}}rg Schl{\"{o}}tterer, and Christin Seifert. 2024.
\newblock \href {https://doi.org/10.48550/ARXIV.2405.00722} {Llms for generating and evaluating counterfactuals: {A} comprehensive study}.
\newblock \emph{CoRR}, abs/2405.00722.

\bibitem[{Ophir et~al.(2020)Ophir, Tikochinski, Asterhan, Sisso, and Reichart}]{Ophir2020DeepNN}
Yaakov Ophir, Refael Tikochinski, Christa S.~C. Asterhan, Itay Sisso, and Roi Reichart. 2020.
\newblock \href {https://api.semanticscholar.org/CorpusID:222209343} {Deep neural networks detect suicide risk from textual facebook posts}.
\newblock \emph{Scientific Reports}, 10.

\bibitem[{Ophir et~al.(2022)Ophir, Tikochinski, Brunstein~Klomek, and Reichart}]{psychology}
Yaakov Ophir, Refael Tikochinski, Anat Brunstein~Klomek, and Roi Reichart. 2022.
\newblock \href {https://journals.sagepub.com/doi/abs/10.1177/21677026211022013?casa_token=Ouqa49hvnL4AAAAA:IcU6FX8Lqp4Qa8h_Zq5TbOJTAr9pFRvPyuu7YWtn5sU9QT_UrDQmPCg-_V8n0REQtRT-Ak70X3C_} {The hitchhiker’s guide to computational linguistics in suicide prevention}.
\newblock \emph{Clinical Psychological Science}, 10(2):212--235.

\bibitem[{Opitz and Frank(2022)}]{opitz-frank-2022-sbert}
Juri Opitz and Anette Frank. 2022.
\newblock \href {https://doi.org/10.18653/v1/2022.aacl-main.48} {{SBERT} studies meaning representations: Decomposing sentence embeddings into explainable semantic features}.
\newblock In \emph{Proceedings of the 2nd Conference of the Asia-Pacific Chapter of the Association for Computational Linguistics and the 12th International Joint Conference on Natural Language Processing (Volume 1: Long Papers)}, pages 625--638, Online only. Association for Computational Linguistics.

\bibitem[{Orgad et~al.(2022)Orgad, Goldfarb{-}Tarrant, and Belinkov}]{OrgadGB22}
Hadas Orgad, Seraphina Goldfarb{-}Tarrant, and Yonatan Belinkov. 2022.
\newblock \href {https://doi.org/10.18653/V1/2022.NAACL-MAIN.188} {How gender debiasing affects internal model representations, and why it matters}.
\newblock In \emph{Proceedings of the 2022 Conference of the North American Chapter of the Association for Computational Linguistics: Human Language Technologies, {NAACL} 2022, Seattle, WA, United States, July 10-15, 2022}, pages 2602--2628. Association for Computational Linguistics.

\bibitem[{Pal et~al.(2023)Pal, Sun, Yuan, Wallace, and Bau}]{PalSYWB23}
Koyena Pal, Jiuding Sun, Andrew Yuan, Byron~C. Wallace, and David Bau. 2023.
\newblock \href {https://doi.org/10.18653/V1/2023.CONLL-1.37} {Future lens: Anticipating subsequent tokens from a single hidden state}.
\newblock In \emph{Proceedings of the 27th Conference on Computational Natural Language Learning, CoNLL 2023, Singapore, December 6-7, 2023}, pages 548--560. Association for Computational Linguistics.

\bibitem[{Papernot and McDaniel(2018)}]{Papernot2018}
Nicolas Papernot and Patrick~D. McDaniel. 2018.
\newblock \href {https://arxiv.org/abs/1803.04765} {Deep k-nearest neighbors: Towards confident, interpretable and robust deep learning}.
\newblock \emph{CoRR}, abs/1803.04765.

\bibitem[{Parasuraman and Riley(1997)}]{Parasuraman1997}
Raja Parasuraman and Victor Riley. 1997.
\newblock \href {https://doi.org/10.1518/001872097778543886} {Humans and automation: Use, misuse, disuse, abuse}.
\newblock \emph{Human Factors}, 39(2):230--253.

\bibitem[{Parcalabescu and Frank(2023)}]{Parcalabescu2023}
Letitia Parcalabescu and Anette Frank. 2023.
\newblock \href {https://doi.org/10.48550/ARXIV.2311.07466} {On measuring faithfulness of natural language explanations}.
\newblock \emph{CoRR}, abs/2311.07466.

\bibitem[{Paulus et~al.(2024)Paulus, Zharmagambetov, Guo, Amos, and Tian}]{Paulus2024}
Anselm Paulus, Arman Zharmagambetov, Chuan Guo, Brandon Amos, and Yuandong Tian. 2024.
\newblock \href {https://doi.org/10.48550/ARXIV.2404.16873} {Advprompter: Fast adaptive adversarial prompting for llms}.
\newblock \emph{CoRR}, abs/2404.16873.

\bibitem[{Pavón(2023)}]{Pavon_2024}
Pedro Pavón. 2023.
\newblock \href {https://about.fb.com/news/2023/02/increasing-our-ads-transparency/} {Increasing our ads transparency}.

\bibitem[{Perez et~al.(2022)Perez, Huang, Song, Cai, Ring, Aslanides, Glaese, McAleese, and Irving}]{PerezHSCRAGMI22}
Ethan Perez, Saffron Huang, H.~Francis Song, Trevor Cai, Roman Ring, John Aslanides, Amelia Glaese, Nat McAleese, and Geoffrey Irving. 2022.
\newblock \href {https://doi.org/10.18653/V1/2022.EMNLP-MAIN.225} {Red teaming language models with language models}.
\newblock In \emph{Proceedings of the 2022 Conference on Empirical Methods in Natural Language Processing, {EMNLP} 2022, Abu Dhabi, United Arab Emirates, December 7-11, 2022}, pages 3419--3448. Association for Computational Linguistics.

\bibitem[{Poursabzi{-}Sangdeh et~al.(2021)Poursabzi{-}Sangdeh, Goldstein, Hofman, Vaughan, and Wallach}]{Poursabzi21}
Forough Poursabzi{-}Sangdeh, Daniel~G. Goldstein, Jake~M. Hofman, Jennifer~Wortman Vaughan, and Hanna~M. Wallach. 2021.
\newblock \href {https://doi.org/10.1145/3411764.3445315} {Manipulating and measuring model interpretability}.
\newblock In \emph{{CHI} '21: {CHI} Conference on Human Factors in Computing Systems, Virtual Event / Yokohama, Japan, May 8-13, 2021}, pages 237:1--237:52. {ACM}.

\bibitem[{Rajagopal et~al.(2021)Rajagopal, Balachandran, Hovy, and Tsvetkov}]{RajagopalBHT21}
Dheeraj Rajagopal, Vidhisha Balachandran, Eduard~H. Hovy, and Yulia Tsvetkov. 2021.
\newblock \href {https://doi.org/10.18653/V1/2021.EMNLP-MAIN.64} {{SELFEXPLAIN:} {A} self-explaining architecture for neural text classifiers}.
\newblock In \emph{Proceedings of the 2021 Conference on Empirical Methods in Natural Language Processing, {EMNLP} 2021, Virtual Event / Punta Cana, Dominican Republic, 7-11 November, 2021}, pages 836--850. Association for Computational Linguistics.

\bibitem[{Rajani et~al.(2019)Rajani, McCann, Xiong, and Socher}]{RajaniMXS19}
Nazneen~Fatema Rajani, Bryan McCann, Caiming Xiong, and Richard Socher. 2019.
\newblock \href {https://doi.org/10.18653/V1/P19-1487} {Explain yourself! leveraging language models for commonsense reasoning}.
\newblock In \emph{Proceedings of the 57th Conference of the Association for Computational Linguistics, {ACL} 2019, Florence, Italy, July 28- August 2, 2019, Volume 1: Long Papers}, pages 4932--4942. Association for Computational Linguistics.

\bibitem[{Rao et~al.(2024)Rao, Yerukola, Shah, Reinecke, and Sap}]{Rao2024}
Abhinav Rao, Akhila Yerukola, Vishwa Shah, Katharina Reinecke, and Maarten Sap. 2024.
\newblock \href {https://doi.org/10.48550/ARXIV.2404.12464} {{NORMAD:} {A} benchmark for measuring the cultural adaptability of large language models}.
\newblock \emph{CoRR}, abs/2404.12464.

\bibitem[{R{\"{a}}uker et~al.(2023)R{\"{a}}uker, Ho, Casper, and Hadfield{-}Menell}]{RaukerHCH23}
Tilman R{\"{a}}uker, Anson Ho, Stephen Casper, and Dylan Hadfield{-}Menell. 2023.
\newblock \href {https://doi.org/10.1109/SATML54575.2023.00039} {Toward transparent {AI:} {A} survey on interpreting the inner structures of deep neural networks}.
\newblock In \emph{2023 {IEEE} Conference on Secure and Trustworthy Machine Learning, SaTML 2023, Raleigh, NC, USA, February 8-10, 2023}, pages 464--483. {IEEE}.

\bibitem[{Ravfogel et~al.(2020)Ravfogel, Elazar, Gonen, Twiton, and Goldberg}]{RavfogelEGTG20}
Shauli Ravfogel, Yanai Elazar, Hila Gonen, Michael Twiton, and Yoav Goldberg. 2020.
\newblock \href {https://doi.org/10.18653/V1/2020.ACL-MAIN.647} {Null it out: Guarding protected attributes by iterative nullspace projection}.
\newblock In \emph{Proceedings of the 58th Annual Meeting of the Association for Computational Linguistics, {ACL} 2020, Online, July 5-10, 2020}, pages 7237--7256. Association for Computational Linguistics.

\bibitem[{Ravichander et~al.(2021)Ravichander, Belinkov, and Hovy}]{RavichanderBH21}
Abhilasha Ravichander, Yonatan Belinkov, and Eduard~H. Hovy. 2021.
\newblock \href {https://doi.org/10.18653/V1/2021.EACL-MAIN.295} {Probing the probing paradigm: Does probing accuracy entail task relevance?}
\newblock In \emph{Proceedings of the 16th Conference of the European Chapter of the Association for Computational Linguistics: Main Volume, {EACL} 2021, Online, April 19 - 23, 2021}, pages 3363--3377. Association for Computational Linguistics.

\bibitem[{Ravichander et~al.(2020)Ravichander, Hovy, Suleman, Trischler, and Cheung}]{RavichanderHSTC20}
Abhilasha Ravichander, Eduard~H. Hovy, Kaheer Suleman, Adam Trischler, and Jackie Chi~Kit Cheung. 2020.
\newblock \href {https://aclanthology.org/2020.starsem-1.10/} {On the systematicity of probing contextualized word representations: The case of hypernymy in {BERT}}.
\newblock In \emph{Proceedings of the Ninth Joint Conference on Lexical and Computational Semantics, *SEM@COLING 2020, Barcelona, Spain (Online), December 12-13, 2020}, pages 88--102. Association for Computational Linguistics.

\bibitem[{Rezaii et~al.(2022)Rezaii, Wolff, and Price}]{psychiatry}
Neguine Rezaii, Phillip Wolff, and Bruce~H Price. 2022.
\newblock \href {https://www.cambridge.org/core/journals/the-british-journal-of-psychiatry/article/natural-language-processing-in-psychiatry-the-promises-and-perils-of-a-transformative-approach/1EF95BF39C8C663026E5ED49C4FFC5EB} {Natural language processing in psychiatry: the promises and perils of a transformative approach}.
\newblock \emph{The British Journal of Psychiatry}, 220(5):251--253.

\bibitem[{Ribeiro et~al.(2016)Ribeiro, Singh, and Guestrin}]{Ribeiro0G16}
Marco~T{\'{u}}lio Ribeiro, Sameer Singh, and Carlos Guestrin. 2016.
\newblock \href {https://doi.org/10.1145/2939672.2939778} {"why should {I} trust you?": Explaining the predictions of any classifier}.
\newblock In \emph{Proceedings of the 22nd {ACM} {SIGKDD} International Conference on Knowledge Discovery and Data Mining, San Francisco, CA, USA, August 13-17, 2016}, pages 1135--1144. {ACM}.

\bibitem[{Ribeiro et~al.(2018)Ribeiro, Singh, and Guestrin}]{Ribeiro0G18}
Marco~T{\'{u}}lio Ribeiro, Sameer Singh, and Carlos Guestrin. 2018.
\newblock \href {https://doi.org/10.1609/AAAI.V32I1.11491} {Anchors: High-precision model-agnostic explanations}.
\newblock In \emph{Proceedings of the Thirty-Second {AAAI} Conference on Artificial Intelligence, (AAAI-18), the 30th innovative Applications of Artificial Intelligence (IAAI-18), and the 8th {AAAI} Symposium on Educational Advances in Artificial Intelligence (EAAI-18), New Orleans, Louisiana, USA, February 2-7, 2018}, pages 1527--1535. {AAAI} Press.

\bibitem[{Ribeiro et~al.(2021)Ribeiro, Wu, Guestrin, and Singh}]{RibeiroWG021}
Marco~T{\'{u}}lio Ribeiro, Tongshuang Wu, Carlos Guestrin, and Sameer Singh. 2021.
\newblock \href {https://doi.org/10.24963/IJCAI.2021/659} {Beyond accuracy: Behavioral testing of {NLP} models with checklist (extended abstract)}.
\newblock In \emph{Proceedings of the Thirtieth International Joint Conference on Artificial Intelligence, {IJCAI} 2021, Virtual Event / Montreal, Canada, 19-27 August 2021}, pages 4824--4828. ijcai.org.

\bibitem[{Roberts et~al.(2020)Roberts, Stewart, and Nielsen}]{Roberts2020AdjustingFC}
Margaret~E. Roberts, Brandon~M Stewart, and Richard~A. Nielsen. 2020.
\newblock \href {https://api.semanticscholar.org/CorpusID:3538372} {Adjusting for confounding with text matching}.
\newblock \emph{American Journal of Political Science}, 64:887--903.

\bibitem[{Rocamora et~al.(2024)Rocamora, Wu, Liu, Chrysos, and Cevher}]{Rocamora2024RevisitingCA}
Elias~Abad Rocamora, Yongtao Wu, Fanghui Liu, Grigorios~G. Chrysos, and Volkan Cevher. 2024.
\newblock \href {https://api.semanticscholar.org/CorpusID:269613847} {Revisiting character-level adversarial attacks}.

\bibitem[{Roscher et~al.(2020)Roscher, Bohn, Duarte, and Garcke}]{RoscherBDG20}
Ribana Roscher, Bastian Bohn, Marco~F. Duarte, and Jochen Garcke. 2020.
\newblock \href {https://doi.org/10.1109/ACCESS.2020.2976199} {Explainable machine learning for scientific insights and discoveries}.
\newblock \emph{{IEEE} Access}, 8:42200--42216.

\bibitem[{Ross et~al.(2021)Ross, Marasovic, and Peters}]{RossMP21}
Alexis Ross, Ana Marasovic, and Matthew~E. Peters. 2021.
\newblock \href {https://doi.org/10.18653/V1/2021.FINDINGS-ACL.336} {Explaining {NLP} models via minimal contrastive editing (mice)}.
\newblock In \emph{Findings of the Association for Computational Linguistics: {ACL/IJCNLP} 2021, Online Event, August 1-6, 2021}, volume {ACL/IJCNLP} 2021 of \emph{Findings of {ACL}}, pages 3840--3852. Association for Computational Linguistics.

\bibitem[{Ross et~al.(2022)Ross, Wu, Peng, Peters, and Gardner}]{RossWPPG22}
Alexis Ross, Tongshuang Wu, Hao Peng, Matthew~E. Peters, and Matt Gardner. 2022.
\newblock \href {https://doi.org/10.18653/V1/2022.ACL-LONG.228} {Tailor: Generating and perturbing text with semantic controls}.
\newblock In \emph{Proceedings of the 60th Annual Meeting of the Association for Computational Linguistics (Volume 1: Long Papers), {ACL} 2022, Dublin, Ireland, May 22-27, 2022}, pages 3194--3213. Association for Computational Linguistics.

\bibitem[{Ruder et~al.(2022)Ruder, Vulic, and S{\o}gaard}]{RuderVS22}
Sebastian Ruder, Ivan Vulic, and Anders S{\o}gaard. 2022.
\newblock \href {https://doi.org/10.18653/V1/2022.FINDINGS-ACL.184} {Square one bias in {NLP:} towards a multi-dimensional exploration of the research manifold}.
\newblock In \emph{Findings of the Association for Computational Linguistics: {ACL} 2022, Dublin, Ireland, May 22-27, 2022}, pages 2340--2354. Association for Computational Linguistics.

\bibitem[{Rudin(2018)}]{Rudin2018}
Cynthia Rudin. 2018.
\newblock \href {https://arxiv.org/abs/1811.10154} {Please stop explaining black box models for high stakes decisions}.
\newblock \emph{CoRR}, abs/1811.10154.

\bibitem[{Sachdeva et~al.(2024)Sachdeva, Tutek, and Gurevych}]{SachdevaTG24}
Rachneet Sachdeva, Martin Tutek, and Iryna Gurevych. 2024.
\newblock \href {https://aclanthology.org/2024.eacl-long.113} {Catfood: Counterfactual augmented training for improving out-of-domain performance and calibration}.
\newblock In \emph{Proceedings of the 18th Conference of the European Chapter of the Association for Computational Linguistics, {EACL} 2024 - Volume 1: Long Papers, St. Julian's, Malta, March 17-22, 2024}, pages 1876--1898. Association for Computational Linguistics.

\bibitem[{Saeed and Omlin(2023)}]{Waddah2O23}
Waddah Saeed and Christian~W. Omlin. 2023.
\newblock \href {https://doi.org/10.1016/J.KNOSYS.2023.110273} {Explainable {AI} {(XAI):} {A} systematic meta-survey of current challenges and future opportunities}.
\newblock \emph{Knowl. Based Syst.}, 263:110273.

\bibitem[{Sajjad et~al.(2022{\natexlab{a}})Sajjad, Durrani, and Dalvi}]{sajjad-etal-2022-neuron}
Hassan Sajjad, Nadir Durrani, and Fahim Dalvi. 2022{\natexlab{a}}.
\newblock \href {https://doi.org/10.1162/tacl_a_00519} {Neuron-level interpretation of deep {NLP} models: A survey}.
\newblock \emph{Transactions of the Association for Computational Linguistics}, 10:1285--1303.

\bibitem[{Sajjad et~al.(2022{\natexlab{b}})Sajjad, Durrani, Dalvi, Alam, Khan, and Xu}]{SajjadDDAK022}
Hassan Sajjad, Nadir Durrani, Fahim Dalvi, Firoj Alam, Abdul~Rafae Khan, and Jia Xu. 2022{\natexlab{b}}.
\newblock \href {https://doi.org/10.18653/V1/2022.NAACL-MAIN.225} {Analyzing encoded concepts in transformer language models}.
\newblock In \emph{Proceedings of the 2022 Conference of the North American Chapter of the Association for Computational Linguistics: Human Language Technologies, {NAACL} 2022, Seattle, WA, United States, July 10-15, 2022}, pages 3082--3101. Association for Computational Linguistics.

\bibitem[{Sakarvadia et~al.(2023)Sakarvadia, Khan, Ajith, Grzenda, Hudson, Bauer, Chard, and Foster}]{Sakarvadia2023}
Mansi Sakarvadia, Arham Khan, Aswathy Ajith, Daniel Grzenda, Nathaniel Hudson, Andr{\'{e}} Bauer, Kyle Chard, and Ian~T. Foster. 2023.
\newblock \href {https://doi.org/10.48550/ARXIV.2310.16270} {Attention lens: {A} tool for mechanistically interpreting the attention head information retrieval mechanism}.
\newblock \emph{CoRR}, abs/2310.16270.

\bibitem[{Samuel(2023)}]{Samuel2023ResponseTT}
Jim Samuel. 2023.
\newblock \href {https://api.semanticscholar.org/CorpusID:258274564} {Response to the march 2023 'pause giant ai experiments: An open letter' by yoshua bengio, signed by stuart russell, elon musk, steve wozniak, yuval noah harari and others…}.
\newblock \emph{SSRN Electronic Journal}.

\bibitem[{Samvelyan et~al.(2024)Samvelyan, Raparthy, Lupu, Hambro, Markosyan, Bhatt, Mao, Jiang, Parker{-}Holder, Foerster, Rockt{\"{a}}schel, and Raileanu}]{Samvelyan}
Mikayel Samvelyan, Sharath~Chandra Raparthy, Andrei Lupu, Eric Hambro, Aram~H. Markosyan, Manish Bhatt, Yuning Mao, Minqi Jiang, Jack Parker{-}Holder, Jakob~N. Foerster, Tim Rockt{\"{a}}schel, and Roberta Raileanu. 2024.
\newblock \href {https://doi.org/10.48550/ARXIV.2402.16822} {Rainbow teaming: Open-ended generation of diverse adversarial prompts}.
\newblock \emph{CoRR}, abs/2402.16822.

\bibitem[{Santoro et~al.(2017)Santoro, Raposo, Barrett, Malinowski, Pascanu, Battaglia, and Lillicrap}]{SantoroRBMPBL17}
Adam Santoro, David Raposo, David G.~T. Barrett, Mateusz Malinowski, Razvan Pascanu, Peter~W. Battaglia, and Tim Lillicrap. 2017.
\newblock \href {https://proceedings.neurips.cc/paper/2017/hash/e6acf4b0f69f6f6e60e9a815938aa1ff-Abstract.html} {A simple neural network module for relational reasoning}.
\newblock In \emph{Advances in Neural Information Processing Systems 30: Annual Conference on Neural Information Processing Systems 2017, December 4-9, 2017, Long Beach, CA, {USA}}, pages 4967--4976.

\bibitem[{Santosh et~al.(2024)Santosh, Baumgartner, St{\"{u}}rmer, Grabmair, and Niklaus}]{SantoshBSGN24}
T.~Y. S.~S. Santosh, Nina Baumgartner, Matthias St{\"{u}}rmer, Matthias Grabmair, and Joel Niklaus. 2024.
\newblock \href {https://aclanthology.org/2024.lrec-main.1434} {Towards explainability and fairness in swiss judgement prediction: Benchmarking on a multilingual dataset}.
\newblock In \emph{Proceedings of the 2024 Joint International Conference on Computational Linguistics, Language Resources and Evaluation, {LREC/COLING} 2024, 20-25 May, 2024, Torino, Italy}, pages 16500--16513. {ELRA} and {ICCL}.

\bibitem[{Sanz{-}Guerrero and Arroyo(2024)}]{Sanz2024}
Mario Sanz{-}Guerrero and Javier Arroyo. 2024.
\newblock \href {https://doi.org/10.48550/ARXIV.2401.16458} {Credit risk meets large language models: Building a risk indicator from loan descriptions in {P2P} lending}.
\newblock \emph{CoRR}, abs/2401.16458.

\bibitem[{Sarwar et~al.(2022)Sarwar, Zlatkova, Hardalov, Dinkov, Augenstein, and Nakov}]{SarwarZHDAN22}
Sheikh~Muhammad Sarwar, Dimitrina Zlatkova, Momchil Hardalov, Yoan Dinkov, Isabelle Augenstein, and Preslav Nakov. 2022.
\newblock \href {https://doi.org/10.1162/TACL\_A\_00472} {A neighborhood framework for resource-lean content flagging}.
\newblock \emph{Trans. Assoc. Comput. Linguistics}, 10:484--502.

\bibitem[{Schulze{-}Weddige and Zylowski(2021)}]{Schulze-Weddige21}
Sophia Schulze{-}Weddige and Thorsten Zylowski. 2021.
\newblock \href {https://doi.org/10.1007/978-3-030-95531-1\_31} {User study on the effects explainable {AI} visualizations on non-experts}.
\newblock In \emph{ArtsIT, Interactivity and Game Creation - Creative Heritage. New Perspectives from Media Arts and Artificial Intelligence. 10th {EAI} International Conference, ArtsIT 2021, Virtual Event, December 2-3, 2021, Proceedings}, volume 422 of \emph{Lecture Notes of the Institute for Computer Sciences, Social Informatics and Telecommunications Engineering}, pages 457--467. Springer.

\bibitem[{Sen et~al.(2023)Sen, Assenmacher, Samory, Augenstein, van~der Aalst, and Wagner}]{SenASAA023}
Indira Sen, Dennis Assenmacher, Mattia Samory, Isabelle Augenstein, Wil M.~P. van~der Aalst, and Claudia Wagner. 2023.
\newblock \href {https://doi.org/10.18653/V1/2023.EMNLP-MAIN.649} {People make better edits: Measuring the efficacy of llm-generated counterfactually augmented data for harmful language detection}.
\newblock In \emph{Proceedings of the 2023 Conference on Empirical Methods in Natural Language Processing, {EMNLP} 2023, Singapore, December 6-10, 2023}, pages 10480--10504. Association for Computational Linguistics.

\bibitem[{Sennrich(2017)}]{Sennrich17}
Rico Sennrich. 2017.
\newblock \href {https://doi.org/10.18653/V1/E17-2060} {How grammatical is character-level neural machine translation? assessing {MT} quality with contrastive translation pairs}.
\newblock In \emph{Proceedings of the 15th Conference of the European Chapter of the Association for Computational Linguistics, {EACL} 2017, Valencia, Spain, April 3-7, 2017, Volume 2: Short Papers}, pages 376--382. Association for Computational Linguistics.

\bibitem[{Shapira et~al.(2023)Shapira, Apel, Tennenholtz, and Reichart}]{Shapira2023}
Eilam Shapira, Reut Apel, Moshe Tennenholtz, and Roi Reichart. 2023.
\newblock \href {https://doi.org/10.48550/ARXIV.2305.10361} {Human choice prediction in language-based non-cooperative games: Simulation-based off-policy evaluation}.
\newblock \emph{CoRR}, abs/2305.10361.

\bibitem[{Shapira et~al.(2024)Shapira, Madmon, Reichart, and Tennenholtz}]{Shapira2024}
Eilam Shapira, Omer Madmon, Roi Reichart, and Moshe Tennenholtz. 2024.
\newblock \href {https://doi.org/10.48550/ARXIV.2401.17435} {Can large language models replace economic choice prediction labs?}
\newblock \emph{CoRR}, abs/2401.17435.

\bibitem[{Sharkey et~al.(2025)Sharkey, Chughtai, Batson, Lindsey, Wu, Bushnaq, Goldowsky-Dill, Heimersheim, Ortega, Bloom, Biderman, Garriga-Alonso, Conmy, Nanda, Rumbelow, Wattenberg, Schoots, Miller, Michaud, Casper, Tegmark, Saunders, Bau, Todd, Geiger, Geva, Hoogland, Murfet, and McGrath}]{Sharkey2025OpenPI}
Lee Sharkey, Bilal Chughtai, Joshua Batson, Jack Lindsey, Jeff Wu, Lucius Bushnaq, Nicholas Goldowsky-Dill, Stefan Heimersheim, Alejandro Ortega, Joseph Bloom, Stella Biderman, Adri{\`a} Garriga-Alonso, Arthur Conmy, Neel Nanda, Jessica Rumbelow, Martin Wattenberg, Nandi Schoots, Joseph Miller, Eric~J. Michaud, Stephen Casper, Max Tegmark, William Saunders, David Bau, Eric Todd, Atticus Geiger, Mor Geva, Jesse Hoogland, Daniel Murfet, and Thomas McGrath. 2025.
\newblock \href {https://arxiv.org/abs/2501.16496} {Open problems in mechanistic interpretability}.

\bibitem[{Sikdar et~al.(2021)Sikdar, Bhattacharya, and Heese}]{SikdarBH20}
Sandipan Sikdar, Parantapa Bhattacharya, and Kieran Heese. 2021.
\newblock \href {https://doi.org/10.18653/V1/2021.ACL-LONG.71} {Integrated directional gradients: Feature interaction attribution for neural {NLP} models}.
\newblock In \emph{Proceedings of the 59th Annual Meeting of the Association for Computational Linguistics and the 11th International Joint Conference on Natural Language Processing, {ACL/IJCNLP} 2021, (Volume 1: Long Papers), Virtual Event, August 1-6, 2021}, pages 865--878. Association for Computational Linguistics.

\bibitem[{Singh et~al.(2024)Singh, Inala, Galley, Caruana, and Gao}]{Singh2024}
Chandan Singh, Jeevana~Priya Inala, Michel Galley, Rich Caruana, and Jianfeng Gao. 2024.
\newblock \href {https://doi.org/10.48550/ARXIV.2402.01761} {Rethinking interpretability in the era of large language models}.
\newblock \emph{CoRR}, abs/2402.01761.

\bibitem[{Singhal et~al.(2023)Singhal, Azizi, Tu, Mahdavi, Wei, Chung, Scales, Tanwani, Cole-Lewis, Pfohl et~al.}]{singhal2023large}
Karan Singhal, Shekoofeh Azizi, Tao Tu, S~Sara Mahdavi, Jason Wei, Hyung~Won Chung, Nathan Scales, Ajay Tanwani, Heather Cole-Lewis, Stephen Pfohl, et~al. 2023.
\newblock \href {https://www.nature.com/articles/s41586-023-06291-2} {Large language models encode clinical knowledge}.
\newblock \emph{Nature}, 620(7972):172--180.

\bibitem[{Smilkov et~al.(2017)Smilkov, Thorat, Kim, Vi{\'{e}}gas, and Wattenberg}]{SmilkovTKVW17}
Daniel Smilkov, Nikhil Thorat, Been Kim, Fernanda~B. Vi{\'{e}}gas, and Martin Wattenberg. 2017.
\newblock \href {https://arxiv.org/abs/1706.03825} {Smoothgrad: removing noise by adding noise}.
\newblock \emph{CoRR}, abs/1706.03825.

\bibitem[{Smith et~al.(2022)Smith, Hall, Kambadur, Presani, and Williams}]{SmithHKPW22}
Eric~Michael Smith, Melissa Hall, Melanie Kambadur, Eleonora Presani, and Adina Williams. 2022.
\newblock \href {https://doi.org/10.18653/V1/2022.EMNLP-MAIN.625} {"i'm sorry to hear that": Finding new biases in language models with a holistic descriptor dataset}.
\newblock In \emph{Proceedings of the 2022 Conference on Empirical Methods in Natural Language Processing, {EMNLP} 2022, Abu Dhabi, United Arab Emirates, December 7-11, 2022}, pages 9180--9211. Association for Computational Linguistics.

\bibitem[{Srivastava et~al.(2022)Srivastava, Rastogi, Rao, Shoeb, Abid, and et~al.}]{Srivastava2022}
Aarohi Srivastava, Abhinav Rastogi, Abhishek Rao, Abu Awal~Md Shoeb, Abubakar Abid, and et~al. 2022.
\newblock \href {https://doi.org/10.48550/ARXIV.2206.04615} {Beyond the imitation game: Quantifying and extrapolating the capabilities of language models}.
\newblock \emph{CoRR}, abs/2206.04615.

\bibitem[{Stacey et~al.(2024)Stacey, Minervini, Dubossarsky, Camburu, and Rei}]{stacey-etal-2024-atomic}
Joe Stacey, Pasquale Minervini, Haim Dubossarsky, Oana-Maria Camburu, and Marek Rei. 2024.
\newblock \href {https://doi.org/10.18653/v1/2024.emnlp-main.569} {Atomic inference for {NLI} with generated facts as atoms}.
\newblock In \emph{Proceedings of the 2024 Conference on Empirical Methods in Natural Language Processing}, pages 10188--10204, Miami, Florida, USA. Association for Computational Linguistics.

\bibitem[{Stacey et~al.(2022)Stacey, Minervini, Dubossarsky, and Rei}]{StaceyMDR22}
Joe Stacey, Pasquale Minervini, Haim Dubossarsky, and Marek Rei. 2022.
\newblock \href {https://doi.org/10.18653/V1/2022.EMNLP-MAIN.251} {Logical reasoning with span-level predictions for interpretable and robust {NLI} models}.
\newblock In \emph{Proceedings of the 2022 Conference on Empirical Methods in Natural Language Processing, {EMNLP} 2022, Abu Dhabi, United Arab Emirates, December 7-11, 2022}, pages 3809--3823. Association for Computational Linguistics.

\bibitem[{Su et~al.(2023)Su, Li, and Lease}]{Su2023}
Yiheng Su, Junyi~Jessy Li, and Matthew Lease. 2023.
\newblock \href {https://doi.org/10.48550/ARXIV.2311.08644} {Interpretable by design: Wrapper boxes combine neural performance with faithful explanations}.
\newblock \emph{CoRR}, abs/2311.08644.

\bibitem[{Sullivan(2024)}]{Sullivan24}
Michael Sullivan. 2024.
\newblock \href {https://aclanthology.org/2024.eacl-long.116} {It is not true that transformers are inductive learners: Probing {NLI} models with external negation}.
\newblock In \emph{Proceedings of the 18th Conference of the European Chapter of the Association for Computational Linguistics, {EACL} 2024 - Volume 1: Long Papers, St. Julian's, Malta, March 17-22, 2024}, pages 1924--1945. Association for Computational Linguistics.

\bibitem[{Sun et~al.(2022)Sun, Swayamdipta, May, and Ma}]{SunSMM22}
Jiao Sun, Swabha Swayamdipta, Jonathan May, and Xuezhe Ma. 2022.
\newblock \href {https://doi.org/10.18653/V1/2022.FINDINGS-EMNLP.432} {Investigating the benefits of free-form rationales}.
\newblock In \emph{Findings of the Association for Computational Linguistics: {EMNLP} 2022, Abu Dhabi, United Arab Emirates, December 7-11, 2022}, pages 5867--5882. Association for Computational Linguistics.

\bibitem[{Syed et~al.(2023)Syed, Rager, and Conmy}]{Syed2023}
Aaquib Syed, Can Rager, and Arthur Conmy. 2023.
\newblock \href {https://doi.org/10.48550/ARXIV.2310.10348} {Attribution patching outperforms automated circuit discovery}.
\newblock \emph{CoRR}, abs/2310.10348.

\bibitem[{Tan et~al.(2014)Tan, Lee, and Pang}]{TanLP14}
Chenhao Tan, Lillian Lee, and Bo~Pang. 2014.
\newblock \href {https://doi.org/10.3115/V1/P14-1017} {The effect of wording on message propagation: Topic- and author-controlled natural experiments on twitter}.
\newblock In \emph{Proceedings of the 52nd Annual Meeting of the Association for Computational Linguistics, {ACL} 2014, June 22-27, 2014, Baltimore, MD, USA, Volume 1: Long Papers}, pages 175--185. The Association for Computer Linguistics.

\bibitem[{Tan et~al.(2024)Tan, Cheng, Wang, Yuan, Li, and Liu}]{TanCWYLL24}
Zhen Tan, Lu~Cheng, Song Wang, Bo~Yuan, Jundong Li, and Huan Liu. 2024.
\newblock \href {https://doi.org/10.1007/978-981-97-2259-4\_5} {Interpreting pretrained language models via concept bottlenecks}.
\newblock In \emph{Advances in Knowledge Discovery and Data Mining - 28th Pacific-Asia Conference on Knowledge Discovery and Data Mining, {PAKDD} 2024, Taipei, Taiwan, May 7-10, 2024, Proceedings, Part {III}}, volume 14647 of \emph{Lecture Notes in Computer Science}, pages 56--74. Springer.

\bibitem[{Taubenfeld et~al.(2024)Taubenfeld, Dover, Reichart, and Goldstein}]{Taubenfeld2024}
Amir Taubenfeld, Yaniv Dover, Roi Reichart, and Ariel Goldstein. 2024.
\newblock \href {https://doi.org/10.48550/ARXIV.2402.04049} {Systematic biases in {LLM} simulations of debates}.
\newblock \emph{CoRR}, abs/2402.04049.

\bibitem[{Thirunavukarasu et~al.(2023)Thirunavukarasu, Ting, Elangovan, Gutierrez, Tan, and Ting}]{thirunavukarasu2023large}
Arun~James Thirunavukarasu, Darren Shu~Jeng Ting, Kabilan Elangovan, Laura Gutierrez, Ting~Fang Tan, and Daniel Shu~Wei Ting. 2023.
\newblock \href {https://idp.nature.com/authorize/casa?redirect_uri=https://www.nature.com/articles/s41591-023-02448-8&casa_token=QTsnMmzDanoAAAAA:0LUAEXlkWCe2mEg_0gqJt5nb1HS0oDxYcLUkLnYx8WR0D9Nn9_Uo2v8TF4Sj-oaLD67Lye9I1cE2pJum} {Large language models in medicine}.
\newblock \emph{Nature medicine}, 29(8):1930--1940.

\bibitem[{Thompson and Mimno(2020)}]{Thompson2020}
Laure Thompson and David Mimno. 2020.
\newblock \href {https://arxiv.org/abs/2010.12626} {Topic modeling with contextualized word representation clusters}.
\newblock \emph{CoRR}, abs/2010.12626.

\bibitem[{Tikochinski et~al.(2024)Tikochinski, Goldstein, Meiri, Hasson, and Reichart}]{Tikochinski2024IncrementalAO}
Refael Tikochinski, Ariel Goldstein, Yoav Meiri, Uri Hasson, and Roi Reichart. 2024.
\newblock \href {https://api.semanticscholar.org/CorpusID:267035486} {Incremental accumulation of linguistic context in artificial and biological neural networks}.
\newblock \emph{bioRxiv}.

\bibitem[{Tikochinski et~al.(2023)Tikochinski, Goldstein, Yeshurun, Hasson, and Reichart}]{tikochinski2023perspective}
Refael Tikochinski, Ariel Goldstein, Yaara Yeshurun, Uri Hasson, and Roi Reichart. 2023.
\newblock \href {https://academic.oup.com/cercor/article-abstract/33/12/7830/7080913} {Perspective changes in human listeners are aligned with the contextual transformation of the word embedding space}.
\newblock \emph{Cerebral Cortex}, 33(12):7830--7842.

\bibitem[{Toker et~al.(2024)Toker, Orgad, Ventura, Arad, and Belinkov}]{Toker2024}
Michael Toker, Hadas Orgad, Mor Ventura, Dana Arad, and Yonatan Belinkov. 2024.
\newblock \href {https://doi.org/10.48550/ARXIV.2403.05846} {Diffusion lens: Interpreting text encoders in text-to-image pipelines}.
\newblock \emph{CoRR}, abs/2403.05846.

\bibitem[{Treviso et~al.(2023)Treviso, Ross, Guerreiro, and Martins}]{TrevisoRGM23}
Marcos~V. Treviso, Alexis Ross, Nuno~Miguel Guerreiro, and Andr{\'{e}} F.~T. Martins. 2023.
\newblock \href {https://doi.org/10.18653/V1/2023.ACL-LONG.842} {{CREST:} {A} joint framework for rationalization and counterfactual text generation}.
\newblock In \emph{Proceedings of the 61st Annual Meeting of the Association for Computational Linguistics (Volume 1: Long Papers), {ACL} 2023, Toronto, Canada, July 9-14, 2023}, pages 15109--15126. Association for Computational Linguistics.

\bibitem[{Tu et~al.(2024)Tu, Palepu, Schaekermann, Saab, Freyberg, Tanno, Wang, Li, Amin, Tomasev, Azizi, Singhal, Cheng, Hou, Webson, Kulkarni, Mahdavi, Semturs, Gottweis, Barral, Chou, Corrado, Matias, Karthikesalingam, and Natarajan}]{Tu2024}
Tao Tu, Anil Palepu, Mike Schaekermann, Khaled Saab, Jan Freyberg, Ryutaro Tanno, Amy Wang, Brenna Li, Mohamed Amin, Nenad Tomasev, Shekoofeh Azizi, Karan Singhal, Yong Cheng, Le~Hou, Albert Webson, Kavita Kulkarni, S.~Sara Mahdavi, Christopher Semturs, Juraj Gottweis, Joelle~K. Barral, Katherine Chou, Gregory~S. Corrado, Yossi Matias, Alan Karthikesalingam, and Vivek Natarajan. 2024.
\newblock \href {https://doi.org/10.48550/ARXIV.2401.05654} {Towards conversational diagnostic {AI}}.
\newblock \emph{CoRR}, abs/2401.05654.

\bibitem[{Turpin et~al.(2023)Turpin, Michael, Perez, and Bowman}]{TurpinMPB23}
Miles Turpin, Julian Michael, Ethan Perez, and Samuel~R. Bowman. 2023.
\newblock \href {http://papers.nips.cc/paper\_files/paper/2023/hash/ed3fea9033a80fea1376299fa7863f4a-Abstract-Conference.html} {Language models don't always say what they think: Unfaithful explanations in chain-of-thought prompting}.
\newblock In \emph{Advances in Neural Information Processing Systems 36: Annual Conference on Neural Information Processing Systems 2023, NeurIPS 2023, New Orleans, LA, USA, December 10 - 16, 2023}.

\bibitem[{Ventura et~al.(2023)Ventura, Ben{-}David, Korhonen, and Reichart}]{Ventura2023}
Mor Ventura, Eyal Ben{-}David, Anna Korhonen, and Roi Reichart. 2023.
\newblock \href {https://doi.org/10.48550/ARXIV.2310.01929} {Navigating cultural chasms: Exploring and unlocking the cultural {POV} of text-to-image models}.
\newblock \emph{CoRR}, abs/2310.01929.

\bibitem[{Vig(2019)}]{Vig19}
Jesse Vig. 2019.
\newblock \href {https://doi.org/10.18653/V1/P19-3007} {A multiscale visualization of attention in the transformer model}.
\newblock In \emph{Proceedings of the 57th Conference of the Association for Computational Linguistics, {ACL} 2019, Florence, Italy, July 28 - August 2, 2019, Volume 3: System Demonstrations}, pages 37--42. Association for Computational Linguistics.

\bibitem[{Vig et~al.(2020)Vig, Gehrmann, Belinkov, Qian, Nevo, Singer, and Shieber}]{VigGBQNSS20}
Jesse Vig, Sebastian Gehrmann, Yonatan Belinkov, Sharon Qian, Daniel Nevo, Yaron Singer, and Stuart~M. Shieber. 2020.
\newblock \href {https://proceedings.neurips.cc/paper/2020/hash/92650b2e92217715fe312e6fa7b90d82-Abstract.html} {Investigating gender bias in language models using causal mediation analysis}.
\newblock In \emph{Advances in Neural Information Processing Systems 33: Annual Conference on Neural Information Processing Systems 2020, NeurIPS 2020, December 6-12, 2020, virtual}.

\bibitem[{Voita et~al.(2021)Voita, Sennrich, and Titov}]{VoitaST20}
Elena Voita, Rico Sennrich, and Ivan Titov. 2021.
\newblock \href {https://doi.org/10.18653/V1/2021.ACL-LONG.91} {Analyzing the source and target contributions to predictions in neural machine translation}.
\newblock In \emph{Proceedings of the 59th Annual Meeting of the Association for Computational Linguistics and the 11th International Joint Conference on Natural Language Processing, {ACL/IJCNLP} 2021, (Volume 1: Long Papers), Virtual Event, August 1-6, 2021}, pages 1126--1140. Association for Computational Linguistics.

\bibitem[{von Garrel and Mayer(2023)}]{vonGarrel2023ArtificialII}
J{\"o}rg von Garrel and Jana Mayer. 2023.
\newblock \href {https://api.semanticscholar.org/CorpusID:265069203} {Artificial intelligence in studies—use of chatgpt and ai-based tools among students in germany}.
\newblock \emph{Humanities and Social Sciences Communications}, 10:1--9.

\bibitem[{Vulic et~al.(2020)Vulic, Baker, Ponti, Petti, Leviant, Wing, Majewska, Bar, Malone, Poibeau, Reichart, and Korhonen}]{VulicBPPLWMBMPR20}
Ivan Vulic, Simon Baker, Edoardo~Maria Ponti, Ulla Petti, Ira Leviant, Kelly Wing, Olga Majewska, Eden Bar, Matt Malone, Thierry Poibeau, Roi Reichart, and Anna Korhonen. 2020.
\newblock \href {https://doi.org/10.1162/COLI\_A\_00391} {Multi-simlex: {A} large-scale evaluation of multilingual and crosslingual lexical semantic similarity}.
\newblock \emph{Comput. Linguistics}, 46(4):847--897.

\bibitem[{Vulic et~al.(2023)Vulic, Glavas, Liu, Collier, Ponti, and Korhonen}]{VulicGLCPK23}
Ivan Vulic, Goran Glavas, Fangyu Liu, Nigel Collier, Edoardo~Maria Ponti, and Anna Korhonen. 2023.
\newblock \href {https://doi.org/10.18653/V1/2023.EACL-MAIN.153} {Probing cross-lingual lexical knowledge from multilingual sentence encoders}.
\newblock In \emph{Proceedings of the 17th Conference of the European Chapter of the Association for Computational Linguistics, {EACL} 2023, Dubrovnik, Croatia, May 2-6, 2023}, pages 2081--2097. Association for Computational Linguistics.

\bibitem[{Wallace et~al.(2018)Wallace, Feng, and Boyd{-}Graber}]{WallaceFB18}
Eric Wallace, Shi Feng, and Jordan~L. Boyd{-}Graber. 2018.
\newblock \href {https://doi.org/10.18653/V1/W18-5416} {Interpreting neural networks with nearest neighbors}.
\newblock In \emph{Proceedings of the Workshop: Analyzing and Interpreting Neural Networks for NLP, BlackboxNLP@EMNLP 2018, Brussels, Belgium, November 1, 2018}, pages 136--144. Association for Computational Linguistics.

\bibitem[{Wallace et~al.(2019)Wallace, Tuyls, Wang, Subramanian, Gardner, and Singh}]{WallaceTWSGS19}
Eric Wallace, Jens Tuyls, Junlin Wang, Sanjay Subramanian, Matt Gardner, and Sameer Singh. 2019.
\newblock \href {https://doi.org/10.18653/V1/D19-3002} {Allennlp interpret: {A} framework for explaining predictions of {NLP} models}.
\newblock In \emph{Proceedings of the 2019 Conference on Empirical Methods in Natural Language Processing and the 9th International Joint Conference on Natural Language Processing, {EMNLP-IJCNLP} 2019, Hong Kong, China, November 3-7, 2019 - System Demonstrations}, pages 7--12. Association for Computational Linguistics.

\bibitem[{Wang et~al.(2019{\natexlab{a}})Wang, Pruksachatkun, Nangia, Singh, Michael, Hill, Levy, and Bowman}]{WangPNSMHLB19}
Alex Wang, Yada Pruksachatkun, Nikita Nangia, Amanpreet Singh, Julian Michael, Felix Hill, Omer Levy, and Samuel~R. Bowman. 2019{\natexlab{a}}.
\newblock \href {https://proceedings.neurips.cc/paper/2019/hash/4496bf24afe7fab6f046bf4923da8de6-Abstract.html} {Superglue: {A} stickier benchmark for general-purpose language understanding systems}.
\newblock In \emph{Advances in Neural Information Processing Systems 32: Annual Conference on Neural Information Processing Systems 2019, NeurIPS 2019, December 8-14, 2019, Vancouver, BC, Canada}, pages 3261--3275.

\bibitem[{Wang et~al.(2019{\natexlab{b}})Wang, Singh, Michael, Hill, Levy, and Bowman}]{WangSMHLB19}
Alex Wang, Amanpreet Singh, Julian Michael, Felix Hill, Omer Levy, and Samuel~R. Bowman. 2019{\natexlab{b}}.
\newblock \href {https://openreview.net/forum?id=rJ4km2R5t7} {{GLUE:} {A} multi-task benchmark and analysis platform for natural language understanding}.
\newblock In \emph{7th International Conference on Learning Representations, {ICLR} 2019, New Orleans, LA, USA, May 6-9, 2019}. OpenReview.net.

\bibitem[{Wang et~al.(2023{\natexlab{a}})Wang, Mo, Wang, Zhou, and Chen}]{WangMWZC23}
Fei Wang, Wenjie Mo, Yiwei Wang, Wenxuan Zhou, and Muhao Chen. 2023{\natexlab{a}}.
\newblock \href {https://doi.org/10.18653/V1/2023.FINDINGS-EMNLP.1013} {A causal view of entity bias in (large) language models}.
\newblock In \emph{Findings of the Association for Computational Linguistics: {EMNLP} 2023, Singapore, December 6-10, 2023}, pages 15173--15184. Association for Computational Linguistics.

\bibitem[{Wang et~al.(2023{\natexlab{b}})Wang, Variengien, Conmy, Shlegeris, and Steinhardt}]{WangVCSS23}
Kevin~Ro Wang, Alexandre Variengien, Arthur Conmy, Buck Shlegeris, and Jacob Steinhardt. 2023{\natexlab{b}}.
\newblock \href {https://openreview.net/pdf?id=NpsVSN6o4ul} {Interpretability in the wild: a circuit for indirect object identification in {GPT-2} small}.
\newblock In \emph{The Eleventh International Conference on Learning Representations, {ICLR} 2023, Kigali, Rwanda, May 1-5, 2023}. OpenReview.net.

\bibitem[{Wang et~al.(2023{\natexlab{c}})Wang, Shang, and Zhong}]{WangSZ23}
Zihan Wang, Jingbo Shang, and Ruiqi Zhong. 2023{\natexlab{c}}.
\newblock \href {https://doi.org/10.18653/V1/2023.EMNLP-MAIN.657} {Goal-driven explainable clustering via language descriptions}.
\newblock In \emph{Proceedings of the 2023 Conference on Empirical Methods in Natural Language Processing, {EMNLP} 2023, Singapore, December 6-10, 2023}, pages 10626--10649. Association for Computational Linguistics.

\bibitem[{Waseem and Hovy(2016)}]{WaseemH16}
Zeerak Waseem and Dirk Hovy. 2016.
\newblock \href {https://doi.org/10.18653/V1/N16-2013} {Hateful symbols or hateful people? predictive features for hate speech detection on twitter}.
\newblock In \emph{Proceedings of the Student Research Workshop, SRW@HLT-NAACL 2016, The 2016 Conference of the North American Chapter of the Association for Computational Linguistics: Human Language Technologies, San Diego California, USA, June 12-17, 2016}, pages 88--93. The Association for Computational Linguistics.

\bibitem[{Webster et~al.(2018)Webster, Recasens, Axelrod, and Baldridge}]{WebsterRAB18}
Kellie Webster, Marta Recasens, Vera Axelrod, and Jason Baldridge. 2018.
\newblock \href {https://doi.org/10.1162/TACL\_A\_00240} {Mind the {GAP:} {A} balanced corpus of gendered ambiguous pronouns}.
\newblock \emph{Trans. Assoc. Comput. Linguistics}, 6:605--617.

\bibitem[{White et~al.(2017)White, Rastogi, Duh, and Durme}]{WhiteRDD17}
Aaron~Steven White, Pushpendre Rastogi, Kevin Duh, and Benjamin~Van Durme. 2017.
\newblock \href {https://aclanthology.org/I17-1100/} {Inference is everything: Recasting semantic resources into a unified evaluation framework}.
\newblock In \emph{Proceedings of the Eighth International Joint Conference on Natural Language Processing, {IJCNLP} 2017, Taipei, Taiwan, November 27 - December 1, 2017 - Volume 1: Long Papers}, pages 996--1005. Asian Federation of Natural Language Processing.

\bibitem[{Wiegreffe and Marasovic(2021)}]{WiegreffeM21}
Sarah Wiegreffe and Ana Marasovic. 2021.
\newblock \href {https://datasets-benchmarks-proceedings.neurips.cc/paper/2021/hash/698d51a19d8a121ce581499d7b701668-Abstract-round1.html} {Teach me to explain: {A} review of datasets for explainable natural language processing}.
\newblock In \emph{Proceedings of the Neural Information Processing Systems Track on Datasets and Benchmarks 1, NeurIPS Datasets and Benchmarks 2021, December 2021, virtual}.

\bibitem[{Wiegreffe and Pinter(2019)}]{WiegreffeP19}
Sarah Wiegreffe and Yuval Pinter. 2019.
\newblock \href {https://doi.org/10.18653/V1/D19-1002} {Attention is not not explanation}.
\newblock In \emph{Proceedings of the 2019 Conference on Empirical Methods in Natural Language Processing and the 9th International Joint Conference on Natural Language Processing, {EMNLP-IJCNLP} 2019, Hong Kong, China, November 3-7, 2019}, pages 11--20. Association for Computational Linguistics.

\bibitem[{Wilcox et~al.(2018)Wilcox, Levy, Morita, and Futrell}]{WilcoxLMF18}
Ethan Wilcox, Roger Levy, Takashi Morita, and Richard Futrell. 2018.
\newblock \href {https://doi.org/10.18653/V1/W18-5423} {What do {RNN} language models learn about filler-gap dependencies?}
\newblock In \emph{Proceedings of the Workshop: Analyzing and Interpreting Neural Networks for NLP, BlackboxNLP@EMNLP 2018, Brussels, Belgium, November 1, 2018}, pages 211--221. Association for Computational Linguistics.

\bibitem[{Wood{-}Doughty et~al.(2018)Wood{-}Doughty, Shpitser, and Dredze}]{Wood-DoughtySD18}
Zach Wood{-}Doughty, Ilya Shpitser, and Mark Dredze. 2018.
\newblock \href {https://aclanthology.org/D18-1488/} {Challenges of using text classifiers for causal inference}.
\newblock In \emph{Proceedings of the 2018 Conference on Empirical Methods in Natural Language Processing, Brussels, Belgium, October 31 - November 4, 2018}, pages 4586--4598. Association for Computational Linguistics.

\bibitem[{Wu et~al.(2021)Wu, Ribeiro, Heer, and Weld}]{WuRHW20}
Tongshuang Wu, Marco~T{\'{u}}lio Ribeiro, Jeffrey Heer, and Daniel~S. Weld. 2021.
\newblock \href {https://doi.org/10.18653/V1/2021.ACL-LONG.523} {Polyjuice: Generating counterfactuals for explaining, evaluating, and improving models}.
\newblock In \emph{Proceedings of the 59th Annual Meeting of the Association for Computational Linguistics and the 11th International Joint Conference on Natural Language Processing, {ACL/IJCNLP} 2021, (Volume 1: Long Papers), Virtual Event, August 1-6, 2021}, pages 6707--6723. Association for Computational Linguistics.

\bibitem[{Wu et~al.(2024)Wu, Zhao, Zhu, Shi, Yang, Liu, Zhai, Yao, Li, Du, and Liu}]{Wu2024}
Xuansheng Wu, Haiyan Zhao, Yaochen Zhu, Yucheng Shi, Fan Yang, Tianming Liu, Xiaoming Zhai, Wenlin Yao, Jundong Li, Mengnan Du, and Ninghao Liu. 2024.
\newblock \href {https://doi.org/10.48550/ARXIV.2403.08946} {Usable {XAI:} 10 strategies towards exploiting explainability in the {LLM} era}.
\newblock \emph{CoRR}, abs/2403.08946.

\bibitem[{Wu et~al.(2023{\natexlab{a}})Wu, D'Oosterlinck, Geiger, Zur, and Potts}]{WuDGZP23}
Zhengxuan Wu, Karel D'Oosterlinck, Atticus Geiger, Amir Zur, and Christopher Potts. 2023{\natexlab{a}}.
\newblock \href {https://proceedings.mlr.press/v202/wu23b.html} {Causal proxy models for concept-based model explanations}.
\newblock In \emph{International Conference on Machine Learning, {ICML} 2023, 23-29 July 2023, Honolulu, Hawaii, {USA}}, volume 202 of \emph{Proceedings of Machine Learning Research}, pages 37313--37334. {PMLR}.

\bibitem[{Wu et~al.(2023{\natexlab{b}})Wu, Geiger, Icard, Potts, and Goodman}]{WuGIPG23}
Zhengxuan Wu, Atticus Geiger, Thomas Icard, Christopher Potts, and Noah~D. Goodman. 2023{\natexlab{b}}.
\newblock \href {http://papers.nips.cc/paper\_files/paper/2023/hash/f6a8b109d4d4fd64c75e94aaf85d9697-Abstract-Conference.html} {Interpretability at scale: Identifying causal mechanisms in alpaca}.
\newblock In \emph{Advances in Neural Information Processing Systems 36: Annual Conference on Neural Information Processing Systems 2023, NeurIPS 2023, New Orleans, LA, USA, December 10 - 16, 2023}.

\bibitem[{Wu et~al.(2020)Wu, Chen, Kao, and Liu}]{WuCKL20}
Zhiyong Wu, Yun Chen, Ben Kao, and Qun Liu. 2020.
\newblock \href {https://doi.org/10.18653/V1/2020.ACL-MAIN.383} {Perturbed masking: Parameter-free probing for analyzing and interpreting {BERT}}.
\newblock In \emph{Proceedings of the 58th Annual Meeting of the Association for Computational Linguistics, {ACL} 2020, Online, July 5-10, 2020}, pages 4166--4176. Association for Computational Linguistics.

\bibitem[{Xiong et~al.(2023)Xiong, Hu, Lu, Li, Fu, He, and Hooi}]{Xiong2023}
Miao Xiong, Zhiyuan Hu, Xinyang Lu, Yifei Li, Jie Fu, Junxian He, and Bryan Hooi. 2023.
\newblock \href {https://doi.org/10.48550/ARXIV.2306.13063} {Can llms express their uncertainty? an empirical evaluation of confidence elicitation in llms}.
\newblock \emph{CoRR}, abs/2306.13063.

\bibitem[{Yang et~al.(2024)Yang, Jin, Tang, Han, Feng, Jiang, Zhong, Yin, and Hu}]{YangJTHFJZYH24}
Jingfeng Yang, Hongye Jin, Ruixiang Tang, Xiaotian Han, Qizhang Feng, Haoming Jiang, Shaochen Zhong, Bing Yin, and Xia~Ben Hu. 2024.
\newblock \href {https://doi.org/10.1145/3649506} {Harnessing the power of llms in practice: {A} survey on chatgpt and beyond}.
\newblock \emph{{ACM} Trans. Knowl. Discov. Data}, 18(6):160:1--160:32.

\bibitem[{Yang et~al.(2022)Yang, Yuan, and Lau}]{YangYL22}
Kai Yang, Hui Yuan, and Raymond Y.~K. Lau. 2022.
\newblock \href {https://doi.org/10.1016/J.ESWA.2022.116847} {Psycredit: An interpretable deep learning-based credit assessment approach facilitated by psychometric natural language processing}.
\newblock \emph{Expert Syst. Appl.}, 198:116847.

\bibitem[{Yang et~al.(2020)Yang, Chen, Hsieh, Wang, and Jordan}]{YangCHWJ20}
Puyudi Yang, Jianbo Chen, Cho{-}Jui Hsieh, Jane{-}Ling Wang, and Michael~I. Jordan. 2020.
\newblock \href {http://jmlr.org/papers/v21/19-569.html} {Greedy attack and gumbel attack: Generating adversarial examples for discrete data}.
\newblock \emph{J. Mach. Learn. Res.}, 21:43:1--43:36.

\bibitem[{Yao et~al.(2021)Yao, Chen, Ye, Jin, and Ren}]{YaoCYJR21}
Huihan Yao, Ying Chen, Qinyuan Ye, Xisen Jin, and Xiang Ren. 2021.
\newblock \href {https://proceedings.neurips.cc/paper/2021/hash/4b26dc4663ccf960c8538d595d0a1d3a-Abstract.html} {Refining language models with compositional explanations}.
\newblock In \emph{Advances in Neural Information Processing Systems 34: Annual Conference on Neural Information Processing Systems 2021, NeurIPS 2021, December 6-14, 2021, virtual}, pages 8954--8967.

\bibitem[{Yeh et~al.(2020)Yeh, Kim, Arik, Li, Pfister, and Ravikumar}]{YehKALPR20}
Chih{-}Kuan Yeh, Been Kim, Sercan~{\"{O}}mer Arik, Chun{-}Liang Li, Tomas Pfister, and Pradeep Ravikumar. 2020.
\newblock \href {https://proceedings.neurips.cc/paper/2020/hash/ecb287ff763c169694f682af52c1f309-Abstract.html} {On completeness-aware concept-based explanations in deep neural networks}.
\newblock In \emph{Advances in Neural Information Processing Systems 33: Annual Conference on Neural Information Processing Systems 2020, NeurIPS 2020, December 6-12, 2020, virtual}.

\bibitem[{Yu et~al.(2024)Yu, Healey, Leong, Kohane, and Manrai}]{Yu2024MedicalAI}
Kun-Hsing Yu, Elizabeth Healey, Tze-Yun Leong, Isaac~S. Kohane, and Arjun~Kumar Manrai. 2024.
\newblock \href {https://api.semanticscholar.org/CorpusID:270122572} {Medical artificial intelligence and human values.}
\newblock \emph{The New England journal of medicine}, 390 20:1895--1904.

\bibitem[{Yu et~al.(2023)Yu, Merullo, and Pavlick}]{YuMP23}
Qinan Yu, Jack Merullo, and Ellie Pavlick. 2023.
\newblock \href {https://doi.org/10.18653/V1/2023.EMNLP-MAIN.615} {Characterizing mechanisms for factual recall in language models}.
\newblock In \emph{Proceedings of the 2023 Conference on Empirical Methods in Natural Language Processing, {EMNLP} 2023, Singapore, December 6-10, 2023}, pages 9924--9959. Association for Computational Linguistics.

\bibitem[{Zaidan et~al.(2007)Zaidan, Eisner, and Piatko}]{ZaidanEP07}
Omar Zaidan, Jason Eisner, and Christine~D. Piatko. 2007.
\newblock \href {https://aclanthology.org/N07-1033/} {Using "annotator rationales" to improve machine learning for text categorization}.
\newblock In \emph{Human Language Technology Conference of the North American Chapter of the Association of Computational Linguistics, Proceedings, April 22-27, 2007, Rochester, New York, {USA}}, pages 260--267. The Association for Computational Linguistics.

\bibitem[{Zhang et~al.(2024{\natexlab{a}})Zhang, Zhang, Zhou, and Xu}]{ZhangFDA24}
Congzhi Zhang, Linhai Zhang, Deyu Zhou, and Guoqiang Xu. 2024{\natexlab{a}}.
\newblock \href {https://doi.org/10.48550/ARXIV.2403.02738} {Causal prompting: Debiasing large language model prompting based on front-door adjustment}.
\newblock \emph{CoRR}, abs/2403.02738.

\bibitem[{Zhang et~al.(2024{\natexlab{b}})Zhang, Ying, Zhuang, Weng, Ying, Zhang, Tong, and Liu}]{ZhangYZWY0TL24}
Huajie Zhang, Yuxin Ying, Fuzhen Zhuang, Haiqin Weng, Sun Ying, Zhao Zhang, Yiqi Tong, and Yan Liu. 2024{\natexlab{b}}.
\newblock \href {https://doi.org/10.1145/3589335.3651537} {Multi-round counterfactual generation: Interpreting and improving models of text classification}.
\newblock In \emph{Companion Proceedings of the {ACM} on Web Conference 2024, {WWW} 2024, Singapore, Singapore, May 13-17, 2024}, pages 774--777. {ACM}.

\bibitem[{Zhang et~al.(2023)Zhang, Kennard, Smith, McFarland, McCallum, and Keith}]{ZhangKSMMK23}
Raymond Zhang, Neha~Nayak Kennard, Daniel~Scott Smith, Daniel~A. McFarland, Andrew McCallum, and Katherine Keith. 2023.
\newblock \href {https://doi.org/10.18653/V1/2023.FINDINGS-ACL.83} {Causal matching with text embeddings: {A} case study in estimating the causal effects of peer review policies}.
\newblock In \emph{Findings of the Association for Computational Linguistics: {ACL} 2023, Toronto, Canada, July 9-14, 2023}, pages 1284--1297. Association for Computational Linguistics.

\bibitem[{Zhang et~al.(2020)Zhang, Wang, Zhang, and Wang}]{ZhangWZW20}
Weiguo Zhang, Chao Wang, Yue Zhang, and Junbo Wang. 2020.
\newblock \href {https://doi.org/10.1016/J.ELERAP.2020.100989} {Credit risk evaluation model with textual features from loan descriptions for {P2P} lending}.
\newblock \emph{Electron. Commer. Res. Appl.}, 42:100989.

\bibitem[{Zhang et~al.(2022)Zhang, Fang, Chen, and Namazi{-}Rad}]{ZhangFCN22}
Zihan Zhang, Meng Fang, Ling Chen, and Mohammad{-}Reza Namazi{-}Rad. 2022.
\newblock \href {https://doi.org/10.18653/V1/2022.NAACL-MAIN.285} {Is neural topic modelling better than clustering? an empirical study on clustering with contextual embeddings for topics}.
\newblock In \emph{Proceedings of the 2022 Conference of the North American Chapter of the Association for Computational Linguistics: Human Language Technologies, {NAACL} 2022, Seattle, WA, United States, July 10-15, 2022}, pages 3886--3893. Association for Computational Linguistics.

\bibitem[{Zhao et~al.(2024)Zhao, Chen, Yang, Liu, Deng, Cai, Wang, Yin, and Du}]{ZhaoCYLDCWYD24}
Haiyan Zhao, Hanjie Chen, Fan Yang, Ninghao Liu, Huiqi Deng, Hengyi Cai, Shuaiqiang Wang, Dawei Yin, and Mengnan Du. 2024.
\newblock \href {https://doi.org/10.1145/3639372} {Explainability for large language models: {A} survey}.
\newblock \emph{{ACM} Trans. Intell. Syst. Technol.}, 15(2):20:1--20:38.

\bibitem[{Zhao et~al.(2018)Zhao, Wang, Yatskar, Ordonez, and Chang}]{ZhaoWYOC18}
Jieyu Zhao, Tianlu Wang, Mark Yatskar, Vicente Ordonez, and Kai{-}Wei Chang. 2018.
\newblock \href {https://doi.org/10.18653/V1/N18-2003} {Gender bias in coreference resolution: Evaluation and debiasing methods}.
\newblock In \emph{Proceedings of the 2018 Conference of the North American Chapter of the Association for Computational Linguistics: Human Language Technologies, NAACL-HLT, New Orleans, Louisiana, USA, June 1-6, 2018, Volume 2 (Short Papers)}, pages 15--20. Association for Computational Linguistics.

\bibitem[{Zhao et~al.(2020)Zhao, Lin, Mi, Jaggi, and Sch{\"{u}}tze}]{ZhaoLMJS20}
Mengjie Zhao, Tao Lin, Fei Mi, Martin Jaggi, and Hinrich Sch{\"{u}}tze. 2020.
\newblock \href {https://doi.org/10.18653/V1/2020.EMNLP-MAIN.174} {Masking as an efficient alternative to finetuning for pretrained language models}.
\newblock In \emph{Proceedings of the 2020 Conference on Empirical Methods in Natural Language Processing, {EMNLP} 2020, Online, November 16-20, 2020}, pages 2226--2241. Association for Computational Linguistics.

\bibitem[{Zhao et~al.(2023{\natexlab{a}})Zhao, Joty, Wang, and Wang}]{ZhaoSF24}
Ruochen Zhao, Shafiq~R. Joty, Yongjie Wang, and Tan Wang. 2023{\natexlab{a}}.
\newblock \href {https://doi.org/10.48550/ARXIV.2305.02160} {Explaining language models' predictions with high-impact concepts}.
\newblock \emph{CoRR}, abs/2305.02160.

\bibitem[{Zhao et~al.(2023{\natexlab{b}})Zhao, Pang, Du, Yang, Li, Cheung, and Lin}]{ZhaoPDYLCL23}
Yunqing Zhao, Tianyu Pang, Chao Du, Xiao Yang, Chongxuan Li, Ngai{-}Man Cheung, and Min Lin. 2023{\natexlab{b}}.
\newblock \href {http://papers.nips.cc/paper\_files/paper/2023/hash/a97b58c4f7551053b0512f92244b0810-Abstract-Conference.html} {On evaluating adversarial robustness of large vision-language models}.
\newblock In \emph{Advances in Neural Information Processing Systems 36: Annual Conference on Neural Information Processing Systems 2023, NeurIPS 2023, New Orleans, LA, USA, December 10 - 16, 2023}.

\bibitem[{Zheng et~al.(2023)Zheng, Shi, Vafa, Feder, and Blei}]{ZhengSVFB23}
Carolina Zheng, Claudia Shi, Keyon Vafa, Amir Feder, and David~M. Blei. 2023.
\newblock \href {https://doi.org/10.18653/V1/2023.ACL-LONG.179} {An invariant learning characterization of controlled text generation}.
\newblock In \emph{Proceedings of the 61st Annual Meeting of the Association for Computational Linguistics (Volume 1: Long Papers), {ACL} 2023, Toronto, Canada, July 9-14, 2023}, pages 3186--3206. Association for Computational Linguistics.

\bibitem[{Zhou et~al.(2024)Zhou, Hwang, Ren, and Sap}]{Zhou2024}
Kaitlyn Zhou, Jena~D. Hwang, Xiang Ren, and Maarten Sap. 2024.
\newblock \href {https://doi.org/10.48550/ARXIV.2401.06730} {Relying on the unreliable: The impact of language models' reluctance to express uncertainty}.
\newblock \emph{CoRR}, abs/2401.06730.

\bibitem[{Zhou and He(2023)}]{Zhou023}
Yuxiang Zhou and Yulan He. 2023.
\newblock \href {https://doi.org/10.18653/V1/2023.FINDINGS-EMNLP.709} {Causal inference from text: Unveiling interactions between variables}.
\newblock In \emph{Findings of the Association for Computational Linguistics: {EMNLP} 2023, Singapore, December 6-10, 2023}, pages 10559--10571. Association for Computational Linguistics.

\bibitem[{Zhu et~al.(2023)Zhu, Zhang, An, Wu, Barrow, Wang, Huang, Nenkova, and Sun}]{ZhuDAN}
Sicheng Zhu, Ruiyi Zhang, Bang An, Gang Wu, Joe Barrow, Zichao Wang, Furong Huang, Ani Nenkova, and Tong Sun. 2023.
\newblock \href {https://doi.org/10.48550/ARXIV.2310.15140} {Autodan: Automatic and interpretable adversarial attacks on large language models}.
\newblock \emph{CoRR}, abs/2310.15140.

\bibitem[{Ziems et~al.(2024)Ziems, Held, Shaikh, Chen, Zhang, and Yang}]{ZiemsHSCZY24}
Caleb Ziems, William Held, Omar Shaikh, Jiaao Chen, Zhehao Zhang, and Diyi Yang. 2024.
\newblock \href {https://doi.org/10.1162/COLI\_A\_00502} {Can large language models transform computational social science?}
\newblock \emph{Comput. Linguistics}, 50(1):237--291.

\bibitem[{Zytek et~al.(2022)Zytek, Liu, Vaithianathan, and Veeramachaneni}]{ZytekLVV22}
Alexandra Zytek, Dongyu Liu, Rhema Vaithianathan, and Kalyan Veeramachaneni. 2022.
\newblock \href {https://doi.org/10.1109/TVCG.2021.3114864} {Sibyl: Understanding and addressing the usability challenges of machine learning in high-stakes decision making}.
\newblock \emph{{IEEE} Trans. Vis. Comput. Graph.}, 28(1):1161--1171.

\end{thebibliography}

\appendix

\renewcommand \thepart{}
\renewcommand \partname{}
\mtcsettitle{parttoc}{}
\addcontentsline{toc}{section}{Appendix} 
\part{Appendix} 
\parttoc 


\section{Properties: Brief}
\label{sec:properties}

This section briefly describes the properties proposed in \S\ref{sec:properties_discussion}.

\medskip\noindent[\textit{what}] \textbf{Explained mechanism \S\ref{sub:mechanism}:} Interpretability methods can explain different mechanisms of the NLP system. While most methods explain the whole system (an \cmechinputoutput{input-output} mechanism), other methods explain input representations (an \cmechinputinternal{input-internal} mechanism) or internal components such as neurons, attention heads, circuits, and more (an \cmechinternalinternal{internal-internal} mechanism). In addition, this property covers any abstraction of the mechanism states (see \S\ref{sub:terms}), for example, explaining the impact of concepts conveyed in the text instead of explaining long and complex raw input. In this case, which is thoroughly discussed in \S\ref{sub:concepts}, the explained mechanism is \cmechconceptoutput{concept-output}.

\medskip\noindent[\textit{what}] \textbf{Scope \S\ref{sub:local}:} Determined by whether the explanation is \clocalexp{local} -- describes the mechanism for an individual input instance, or \cglobalexp{global} -- describes the mechanism for the entire data distribution.

\medskip\noindent[\textit{how}] \textbf{Time \S\ref{sub:posthoc}:} Determined by the time the explanation is formed. \cposthoc{Post-hoc} methods produce explanations after the prediction, while \cintrinsic{intrinsic} methods are built-in: the explanation is generated during the prediction, and the model relies on it.

\medskip\noindent[\textit{how}] \textbf{Access \S\ref{sub:specific_or_agnostic}:} Determined by accessibility requirement to the explained model. \cmodelagnostic{Model-agnostic} methods can only access its inputs and outputs, while \cmodelspecific{model-specific} methods require access to the explained model during the training time of the interpretability method and can access its internal components or representations.

\medskip\noindent[\textit{how}] \textbf{Presentation \S\ref{sub:presenting}:} Determined by how insights extracted by the interpretability method are presented to the stakeholder. This includes 
\cprescores{scores}, such as importance scores or metrics, and \cprevis{visualization}, such as heatmaps and graphs. Other explanations present similar or contrastive \cpreexamples{examples} to stakeholders or communicate insights through \cpretext{texts} written in natural language.

\medskip\noindent[\textit{how}] \textbf{Causal-based \S\ref{sub:causality}:} Providing faithful explanations might involve incorporating techniques from the causality literature. This property determines whether the method is \ccausalbased{causal-based} or \cnotcausal{not}.

\section{Mechanism and Understandable Terms}
\label{sec:mechanism_terms_app}


\subsection{What is the Explained Mechanism?} 
\label{sub:def_mechanism}

\begin{mechanismbox}
\textbf{\textcolor{OliveGreen!85!black}{Mechanism:}} \\
\textit{A process that constitutes a relation between two states of the NLP system.}
\end{mechanismbox}
To complete the definition, A \textit{state of an NLP system} refers to any form of data at any stage within the data analysis process of the system. This includes the initial state, encompassing the raw input received, all intermediate states comprising various levels of transformed data, and the final state, the system's output or decision. For example, the raw input, tokenized input, embeddings, hidden states (of a specific layer),  activations, attention scores, logits, output, decision.
Accordingly, the mechanism we explain is defined by two system states.
For instance, the mechanism between a sentence and the final output is the whole NLP model; the mechanism between the representations of the third layer and those of the fourth layer is the fourth layer; the mechanism between the raw input and the tokenized input is the tokenizer. 

Notably, the explained mechanism does not need to encompass the entire NLP system. It is acceptable for the mechanism to be only a subsystem or a component. Furthermore, it is acceptable for an explanation to be partial with respect to the mechanism. In other words, the explanation may provide specific insight into the mechanism without fully explaining every aspect and functionality. For example, a scientist who wishes to validate a hypothesis might only be interested in the impact of one concept (e.g., how tone impacts the popularity of social media content \citep{TanLP14}). The idea of not providing a complete explanation is also grounded in the philosophy, psychology, and cognitive science literature. For instance, \citet{Miller2017ExplanationIA} advocates that explanations can be selective (humans select a few salient causes instead of a complete causal chain when explaining) and contrastive (Explanations should answer \textit{Why P instead of Q?} rather than \textit{Why P?}).

\subsection{What are Understandable Terms?}
\label{sub:terms}

\begin{termsbox}
\textbf{\textcolor{Peach!75!black}{Understandable terms:}} \\
\textit{The level of abstraction of the states in the mechanism we explain.}
\end{termsbox}
Note that in our description states can be either fully specified or abstracted to some extent. For example, if the input state is the text, then the interpretability method may consider the entire text, but it may also consider abstractions of the text, such as its summary or a list of concepts conveyed in the text. This also holds for the output state. 
For example, in probing methods (see \S\ref{sub:probing}), a classifier is trained to predict a property (often a linguistic property) from the representations of a particular layer of the model to provide insights into the knowledge encoded in model representations \citep{Belinkov22}. Accordingly, the \textit{input-representations} mechanism we explain is the part of the model that transforms input data into the probed layer's representations, and the output state of the mechanism (the representations) is abstracted to a property. For our convenience, we henceforth use the terminology of a \textit{state} for describing a \textit{fully specified state} or an \textit {abstracted state}, remembering that a state may have several different possible abstractions.

The degree of \textit{``understandable terms''}, the level of abstraction, or the form of cognitive chunks (\citet{doshi2017towards} define them to be the basic unit of an explanation) depends on the stakeholder and their specific needs, as they are the ones who utilize the explanation. 
This involves considering their level of expertise and familiarity with NLP models. 
For example, mechanistic interpretability methods (see \S\ref{sub:mechanistic}) aim to explain states of internal components like neurons, targeting developers \citep{Bereska2024}. While these terms are unsuitable for end-users, they can meet the \textit{``understandable''} criterion for developers, even without abstractions.

\section{Additional Analysis Details}
\label{sec:additional}

\begin{table}[!t]
\centering
\begin{adjustbox}{width=0.475\textwidth}
\begin{tabular}{m{0.4\linewidth}|cccc}
\toprule
 & \textbf{Para.} & \textbf{Mech.}  & \textbf{Scope}  & \textbf{Access.} \\
\midrule
Agreements & 92\% & 93\% & 81\% & 92\% \\
\midrule
Disagreements with unknowns & 12\% & 29\% & 69\% & 62\% \\
\midrule
Agreements without unknowns & 93\% & 95\% & 92\% & 97\% \\
\bottomrule
\end{tabular}
\end{adjustbox}
\caption{Agreement statistics between human and LLM annotations of different characteristics: \textit{Paradigm}, \textit{Mechanism}, \textit{Scope} and \textit{Accessibility}.
The first row presents the portion (in percentages) of agreements. The second row presents the portion of disagreements that involve an \textit{`unknown'} annotation (e.g., the LLM annotated the method scope as unknown, but sufficient domain knowledge could infer it.) within the disagreements. The third row presents the portion of agreements, excluding disagreements involving unknowns.
\textbf{Additional statistics:} 96\% of the papers annotated as relevant by the LLM were indeed relevant.
98\% of the \textit{Field} annotations were correct. 100\% of the \textit{Causal-based} property and of the \textit{LLM} field (whether the paper employs an LLM, see Table~\ref{tab:llm_trends}) annotations were correct.}
\label{tab:iaa}
\end{table}

\noindent\textbf{Retrieval:}
We retrieved tens of thousands of NLP interpretability papers using the Semantic Scholar API and by searching queries such as \texttt{NLP interpretability} (a full list of queries is provided in Box~\ref{box:queries}). We kept only papers whose titles or abstracts contained at least one NLP keyword (e.g., \texttt{NLP, LLM, BERT}; see Box~\ref{box:nlp_kw}) and one interpretability keyword (e.g., \texttt{interpretability, XAI, explanation}; see Box~\ref{box:int_kw}). This search and selection process yielded 14,676 papers.

\medskip\noindent\textbf{Annotation and Filtering:}
For determining the relevancy of the papers and annotating them, we employed an LLM (\texttt{gemini-1.5- pro-preview-0514}) and used the zero-shot prompt provided in Box~\ref{box:prompt}. We asked the LLM to determine the relevance of the paper, its field, the paradigm of the interpretability method, the mechanism being explained, the scope and accessibility of the method, and whether it is causal-based. Additionally, we asked the LLM to write a one-sentence summary of the paper and explain its paradigm annotation. The LLM was also instructed to explicitly extract the names of the interpretability methods employed in the paper. We generated three responses (in a JSON format with LLM annotations) for each paper and determined the final annotation of each question by the majority vote. After relevancy filtering, 2,009 papers remained. 

\medskip\noindent\textbf{Correction:} We then sampled and examined a subset of 20 annotated papers. Following this, we decided to apply some automatic rules to fix the annotations: (1) We merged the `computer science' field with the `NLP' field; (2) For the mechanism annotation, we replaced internal components with `internal-internal', and representations with `input-internal'; (3) Many of the scope annotations were `unknown'. In these cases, we replaced `unknown'  with `local' for feature attributions and natural language explanation paradigms and with `global' for probing, diagnostic sets, and mechanistic interpretability paradigms; (4) We replaced `unknown' values of the accessibility annotations with `model-specific' for the SHAP/LIME, probing and mechanistic interpretability paradigms, and with `model-agnostic' for the diagnostic sets paradigm. (5) Initially, we instructed the LLM to determine whether an LLM was employed in the paper. However, it frequently misclassified models such as BERT as LLMs. To improve accuracy, we instead searched the abstracts for specific keywords such as \texttt{LLM, GPT4, ChatGPT, Gemini, Llama}; (6) Since 2024 is not over, we adjusted the publication year of the papers such that each year spans from June of the previous year to the following June.

\medskip\noindent\textbf{Verification:}
To verify the accuracy of the LLM annotations, we randomly sampled another 100 papers, which one of the authors manually annotated. The agreement statistics are presented in Table~\ref{tab:iaa}. Note that many disagreements between human and LLM annotations involved an `unknown' LLM annotation (the second row in Table~\ref{tab:iaa} shows the proportion of such disagreements among all disagreements). For example, the LLM annotated the method scope as unknown, but sufficient domain knowledge could infer it. When excluding unknown disagreements, over 92\% of the annotations for each question were correct. Excluding unknown disagreements when computing the agreement statistics is reasonable since we exclude `unknown' annotations in our analysis in \S\ref{sec:trends}.

\begin{table}[!t]
\centering
\begin{adjustbox}{width=0.48\textwidth}
\begin{tabular}{l|rrr|rrr}
\toprule
\multirow{2}*{\textbf{Paradigm}} & \multicolumn{3}{c}{\textbf{NLP}} & \multicolumn{3}{|c}{\textbf{Outside}} \\
 & \textit{\underline{\#}} & \textit{\underline{\%}} & \textit{\underline{C}} & \textit{\underline{\#}} & \textit{\underline{\%}} & \textit{\underline{C}} \\
\midrule
\cellcolor{SkyBlue!25!white} Attributions&  491 &  32.8 &  20.6 &   200 &    39.0 &     9.7 \\
\cellcolor{RoyalBlue!35!white} LIME/SHAP&   65 &   4.3 &   7.4 &    49 &     9.6 &     4.8 \\
\cellcolor{LimeGreen!35!white} Probing&  168 &  11.2 &  17.9 &    32 &     6.2 &    19.6 \\
\cellcolor{Green!45!white} Clustering&   35 &   2.3 &  10.5 &    50 &     9.7 &     6.0 \\
\cellcolor{Melon!35!white} Mechanistic&  167 &  11.2 &  27.3 &     9 &     1.8 &     8.6 \\
\cellcolor{Red!45!white} Diagnostic&   54 &   3.6 &  17.5 &    12 &     2.3 &     4.9 \\
\cellcolor{Orange!45!white} Adversarial&   76 &   5.1 &  53.1 &     4 &     0.8 &     6.8 \\
\cellcolor{Orchid!35!white} Counterfactuals&   47 &   3.1 &  24.1 &     4 &     0.8 &     0.5 \\
\cellcolor{Fuchsia!55!white} Lang. Expl.&  222 &  14.8 &  13.2 &    77 &    15.0 &     4.7 \\
\cellcolor{Goldenrod!45!white} Self-explain &   98 &   6.6 &  15.7 &    25 &     4.9 &     3.6 \\
\cellcolor{Brown!45!white} Classic &    9 &   0.6 &   0.6 &    13 &     2.5 &     1.5 \\
Unknown                       &   64 &   4.3 &  32.3 &    38 &     7.4 &     6.7 \\
\midrule
Total & 1495 & 100\% & 20.9 & 514 & 100\% & 7.8 \\
\bottomrule
\end{tabular}
\end{adjustbox}
\caption{Absolute numbers (\textit{\underline{\#}}), proportions (\textit{\underline{\%}}), and average number of citations (\textit{\underline{C}}) of interpretability paradigm papers by field (\textbf{NLP} and fields \textbf{Outside} NLP) including all papers from 2015 to 2024.}
\label{tab:paradigms_all}
\end{table}

\onecolumn

\begin{prompt}[label={box:queries}]{Periwinkle}{Queries for semanticscholar search}
NLP interpretability, NLP model interpretability, LLM interpretability,
LLMs interpretability, language models interpretability, interpretability for NLP models, 
interpretability for NLP, interpretability for LLMs, interpretability for language models,
NLP explainability, NLP model explainability, LLM explainability,
language models explainability, explainability for NLP models, explainability for NLP,
explainability for LLMs, explainability for language models, explaining NLP models,
explaining LLMs, explaining language models, interpreting NLP models, interpreting LLMs, 
interpreting language models, NLP explanation, NLP model explanation, LLM explanation,
LLMs explanation, language models explanation, explanation for NLP models,
explanation for NLP, explanation for LLMs, explanation for language models,
explanations for NLP models, explanations for NLP, explanations for LLMs,
explanations for language models, NLP interpretation, NLP model interpretation,
LLM interpretation, LLMs interpretation, language models interpretation, 
interpretation of NLP models, interpretation of LLMs, interpretation fo language models,
black box NLP, black box NLP model, black box NLP models, 
black box LLM, black box LLMs, black box language models, black-box NLP, 
black-box NLP model, black-box NLP models, black-box LLM, black-box LLMs,
black-box language models, white box NLP, white box NLP model, white box NLP models,
white box LLM, white box LLMs, white box language models, white-box NLP,
white-box NLP model, white-box NLP models, white-box LLM, white-box LLMs, 
white-box language models, NLP XAI, NLP model XAI, NLP models XAI, LLM XAI,
LLMs XAI, language models XAI, XAI for NLP models, XAI for LLM, XAI for NLP,
XAI for LLMs, XAI for language models, NLP explainable AI,
LLM explainable AI, LLMs explainable AI, language models explainable AI,
explainable AI for NLP models, explainable AI for LLM, explainable AI for NLP,
explainable AI for LLMs, explainable AI for language models, explainable NLP models,
explainable LLM, explainable NLP, explainable LLMs, explainable language models,
interpretable AI for NLP models, interpretable AI for LLM, interpretable AI for NLP,
interpretable AI for LLMs, interpretable AI for language models, interpretable NLP models,
interpretable LLM, interpretable NLP, interpretable LLMs, interpretable language models,
NLP user trust, user trust in NLP, user trust in NLP models, user trust in LLM,
user trust in LLMs, user trust in language models, NLP transparency, 
NLP model transparency, LLM transparency, LLMs transparency,
language models transparency, transparency of NLP models, transparency of LLMs,
transparency of language models, transparent NLP, transparent NLP models, 
transparent LLMs, transparent LLM, transparent language models, trustworthy NLP models,
trustworthy LLM, trustworthy NLP, trustworthy LLMs, trustworthy language models,
NLP understanding, NLP model understanding, LLM understanding, LLMs understanding,
language models understanding, accountability for NLP models, accountability for LLM,
accountability for NLP, accountability for LLMs, accountability for language models, 
responsible AI for NLP models, responsible AI for LLM, responsible AI for NLP, 
responsible AI for LLMs, responsible AI for language models, responsible NLP models,
responsible LLM, responsible NLP, responsible LLMs, responsible language models
\end{prompt}

\begin{prompt}[label={box:nlp_kw}]{Orchid}{NLP Keywords}
nlp, language model, computatinal linguistics, language processing, llm, gpt, bert, llama
\end{prompt}

\begin{prompt}[label={box:int_kw}]{RedViolet}{Interpretability Keywords}
interpretability, explainability, explanation, interpretation, black box, blackbox,
black-box, white box, whitebox, white-box, xai, explainable, user trust,
interpretable, transparency, trustworthy, transparent, understanding, accountability
\end{prompt}

\begin{prompt}[label={box:prompt}]{Violet}{LLM prompt for annotating abstracts}
{\small
You will be provided with the title and abstract of a paper focused on NLP model interpretability.\\
Carefully read both the title and the abstract. Your task is to extract key information regarding  *only* the interpretability methods discussed in the paper.\\
Respond *only* in the JSON format below.\\
Please address the following questions and extract the specified information:\\
\\
* "relevant" * - (bool) Determine if the paper is relevant if and only if an interpretability method is used, presented or proposed in the paper. If the paper does not discuss interpretability methods or uses one to explain results, the paper is not relevant. Answer true or false.\\
\\
* "NLP research" * - (bool) Determine if the paper is related to NLP research, it can be that the paper is about domains other than NLP (e.g., medicine, social science, natural science, etc...), but uses NLP models with text input. Answer true or false.\\
\\
* "LLM" * - (bool) Determine if an LLM is employed in the paper.\\
\\
* "TL;DR interpretability method" * - (str) One sentence summarizing only the interpretability method used in the paper.\\
\\
* "field" * - (str) Identify the research field of the paper, select from these options:\\
\null\quad- "general NLP", "computer science", "medicine", "psychology", "neuroscience", "education", "engineering", "economics", "natural science", "humanities", "social science"
\\
* "paradigm explanation" * - (str) One sentence explaining the interpretability paradigm used in the paper and justify your answer to the next question.\\
\\
* "paradigm" * - (str) Select the paradigm of the interpretability method from the options below:\\
\null\quad- "feature attributions": Measuring relevance or importance of each input feature (e.g., tokens or words), including methods like perturbations, gradients, propagations, attention scores and attention visualizations.\\
\null\quad- "LIME/SHAP": Training and applying a surrogate model such as LIME or SHAP.\\
\null\quad- "probing": Training a classifier from model representations that predict properties or concepts, or aligning model representations with signals (like brain activity).\\
\null\quad- "clustering": Clustering the data with model representations or other clustering techniques such as Topic Modeling.\\
\null\quad- "mechanistic": Explaining the functionality of internal components like weights, neurons, layers, attention heads, and circuits, using stimuli, activations, patching, scrubbing, logit lens, projections, etc.\\
\null\quad- "diagnostic sets": Analyzing and evaluating the model using diagnostic sets, challenge sets, test suites, or subsets of examples with a common property (e.g., gender, culture).\\
\null\quad- "adversarial attacks": Generating adversarial attacks or writing adversarial prompts that break alignment.\\
\null\quad- "counterfactuals": Generating counterfactuals, contrastive examples, concept counterfactuals, causal matching and other causal-based methods.\\
\null\quad- "natural language explanations": Providing natural language explanations, extractive or abstractive, including rationales and chain-of-thoughts.\\
\null\quad- "classic": Classic and traditional ML models like Logistic Regression, Linear Regression, Decision Trees, Random Forest, XGBoost, SVM, HMM, KNN.\\
\null\quad- "whitebox": Special model architectures, inherently explainable, that provide intrinsic explanations, such as Concept Bottleneck, Neural Module Networks, Knowledge Graphs, KNN-based.\\
\null\quad- "unknown": If it cannot be inferred from the title and abstract.\\
\\
* "methods" * - (list) List the interpretability methods mentioned in the paper. Note that there might be more than one method.\\
\\
* "explaining what" * - (str) Specify what the interpretability method explains in the model, does it epxlain the whole model (input-output), input concepts (concept-output), representations, or internal components. Select from the following options:\\
\null\quad- "input-output", "concept-output", "representations", "word embeddings", "neurons", "layers", "attention heads", "MLPs", "unknown"\\
\\
* "causal" * - (bool) Determine if the abstract mentions the interpretability method is causal-based. Answer true or false.\\
\\
* "local or global" - (str) Determine if the explanation is global (general insights about the model or the whole data) or local (explaining an individual example). Select from the following options:\\
\null\quad- "global", "local", "both", "unknown"\\
\\
* "specific or agnostic" - (str) Determine if the explanation is model-specific (requires access to the model internals, or the interpretability method is trained using the explained model) or model-agnostic (does not require access to the model internals). Select from the following options:\\
\null\quad- "model-specific", "model-agnostic", "both", "unknown"\\
\\
Answer format:\\
```json\\
\{\\
\null\quad"relevant": bool,\\
\null\quad"NLP research": bool,\\
\null\quad"LLM": bool,\\
\null\quad"TL;DR interpretability method": str,\\
\null\quad"field": str,\\
\null\quad"paradigm explanation": str,\\
\null\quad"paradigm": str,\\
\null\quad"methods": list,\\
\null\quad"explaining what": str,\\
\null\quad"causal": bool,\\
\null\quad"local or global": str,\\
\null\quad"specific or agnostic": str\\
\}\\
```\\
\\
Title: [PAPER\_TITLE]\\
Abstract: [PAPER\_ABSTRACT]\\
\\
Answer:}
\end{prompt}



\end{document}